\pdfoutput=1
\documentclass[11pt]{article}

\usepackage[preprint]{acl}

\usepackage{times}
\usepackage{latexsym}

\usepackage[T1]{fontenc}
\usepackage[utf8]{inputenc}

\usepackage{microtype}

\usepackage{inconsolata}

\usepackage{graphicx}

\usepackage{amsmath,amsfonts,bm}

\def\figref#1{Fig.~\ref{#1}}
\def\secref#1{Sec.~\ref{#1}}
\def\eqref#1{equation~\ref{#1}}
\def\1{\bm{1}}

\DeclareMathAlphabet{\mathsfit}{\encodingdefault}{\sfdefault}{m}{sl}
\SetMathAlphabet{\mathsfit}{bold}{\encodingdefault}{\sfdefault}{bx}{n}

\usepackage{bbm}
\usepackage{tikz}
\usepackage{soul}
\usepackage{ulem}
\usepackage{lipsum}
\usepackage{amssymb}
\usepackage{amsmath}
\usepackage{xspace}
\usepackage{float}
\usepackage{multirow}
\usepackage{booktabs}
\usepackage{colortbl}
\usepackage{enumitem}
\usepackage{subcaption}
\usepackage{xcolor}
\usepackage{caption}
\usepackage{array}
\usepackage{makecell}
\usepackage{inconsolata}
\usepackage[framemethod=TikZ]{mdframed}
\newcommand{\SmallHeading}[1]{\noindent\textbf{#1}.}

\newcommand{\Figure}{Fig.\xspace}
\newcommand{\Table}{Tab.\xspace}
\newcommand{\Section}{Sec.\xspace}
\newcommand{\Appendix}{App.\xspace}

\definecolor{mypositive}{RGB}{0, 128, 0}
\definecolor{mynegative}{RGB}{220, 20, 60}
\definecolor{mypositive}{RGB}{24, 103, 173}
\definecolor{mynegative}{RGB}{255, 128, 0}

\newcommand{\grayColor}[0]{gray!10}

\newcommand{\midgrayline}[0]{\arrayrulecolor{gray!50}\midrule\arrayrulecolor{black}}

\newcommand{\taskalias}{\textsc{Track}\xspace}
\newcommand{\CODE}{CODE\xspace}
\newcommand{\WIKI}{WIKI\xspace}
\newcommand{\GROW}{WIKI\xspace}
\newcommand{\MATH}{MATH\xspace}

\newcommand\tabref[1]{Tab.~\ref{#1}}

\definecolor{customblue}{HTML}{1F77B4} 
\definecolor{customorange}{HTML}{FF7F0E}
\definecolor{customgreen}{HTML}{2CA02C} 
\definecolor{customred}{HTML}{D62728}

\newcommand{\icon}[2][0.4cm]{\begin{minipage}{#1}\includegraphics[width=#1]{#2}\end{minipage}}
\newcommand{\bfull}[0]{\icon{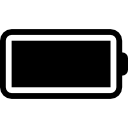}\xspace}
\newcommand{\bempty}[0]{\icon{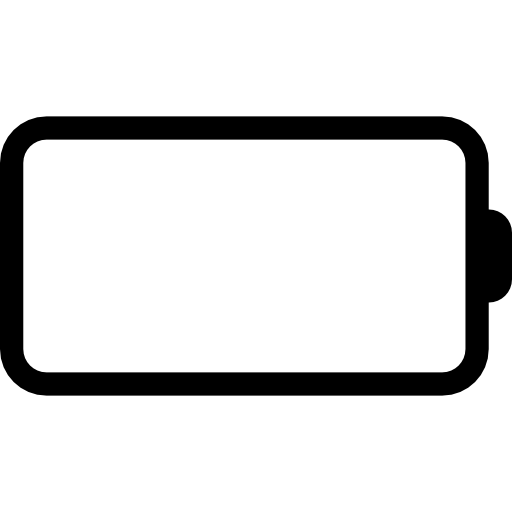}\xspace}
\newcommand{\bhalf}[0]{\icon{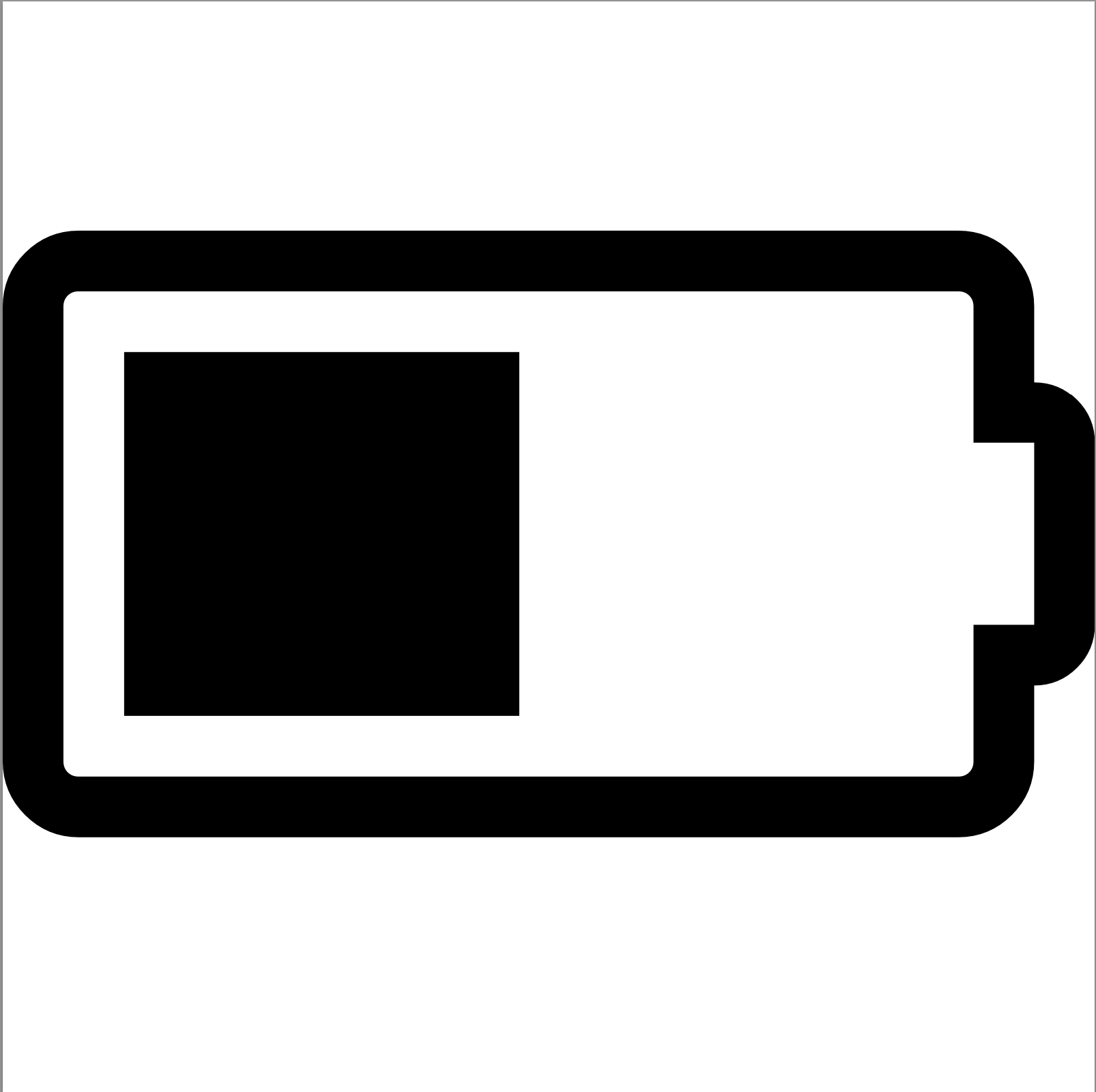}\xspace}
\newcommand{\pythonicon}{\icon{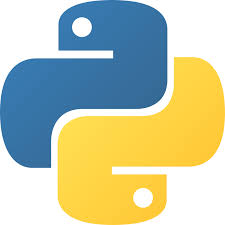}\xspace}
\newcommand{\openaiicon}{\icon{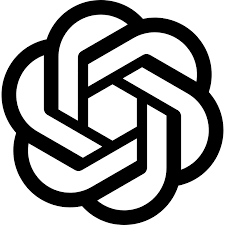}\xspace}

\newcolumntype{C}[1]{>{\centering\let\newline\\\arraybackslash\hspace{0pt}}m{#1}}

\usepackage[skins,breakable]{tcolorbox}

\newtcolorbox{boxblue}{enhanced,colback=blue!5!white,colframe=blue!75!black,breakable=true}

\newtcolorbox{boxsystem}{
  enhanced,
  colback=blue!5!white,
  colframe=blue!75!black,
  breakable=true,
  fontupper=\ttfamily
}

\newtcolorbox{boxuser}{
  enhanced,
  colback=yellow!5!white,
  colframe=yellow!75!black,
  breakable=true,
  fontupper=\ttfamily,
}

\newcommand{\systemp}[1]{
  \begin{boxsystem}
    {\rmfamily\textbf{System:}}\\[1ex]
    #1
  \end{boxsystem}
}

\newcommand{\userp}[1]{
  \begin{boxuser}
    {\rmfamily\textbf{User:}}\\[1ex]
    #1
  \end{boxuser}
}

\title{Tracking the Limits of Knowledge Propagation: How LLMs Fail at Multi-Step Reasoning with Conflicting Knowledge}
\vspace{4mm}
\newcommand{\affilEPFL}{\ensuremath{^\clubsuit}}
\newcommand{\affilSBU}{\ensuremath{^\blacklozenge}}
\author{\parbox[t]{1.0cm}{\raggedright
  \text{Yiyang Feng}\affilEPFL\affilSBU} \\
  \parbox[t]{4.2cm}{\raggedright
  \texttt{yiyang.feng@stonybrook.edu}} \\\And
  \parbox[t]{1.0cm}{\raggedright
  \text{Zeming Chen}~\affilEPFL} \\
  \parbox[t]{2.5cm}{\raggedright
  \texttt{zeming.chen@epfl.ch}} \\\And
  Haotian Wu~\affilEPFL \\
  \texttt{haotian.wu@epfl.ch} \\\AND
  Jiawei Zhou~\affilSBU \\
  \texttt{jiawei.zhou.1@stonybrook.edu} \\\And
  Antoine Bosselut~\affilEPFL \\
  \texttt{antoine.bosselut@epfl.ch} \\\AND
  {\normalfont \affilEPFL~EPFL} \quad
  {\normalfont \affilSBU~Stony Brook University} \quad \\
}

\begin{document}
\maketitle
\begin{abstract}
% One or two sentences of background, problem/gap. Knowledge propagation in complex reasoning is challenging because of knowledge conflicts, but limited benchmarks study it.

% Large Language Models (LLMs) fail at reasoning when their parametric knowledge is outdated or incorrect.
A common solution for mitigating outdated or incorrect information in Large Language Models (LLMs) is to provide updated facts in-context or through knowledge editing. However, these methods introduce knowledge conflicts when the knowledge update fails to overwrite the model's parametric knowledge, which propagate to faulty reasoning.
% but such updates can conflict with LLMs' parametric knowledge and limit their reasoning. 
Current benchmarks for this problem, however, largely focus only on single knowledge updates and fact recall without evaluating how these updates affect downstream reasoning. %, and relying on counterfactual data.
% Our contribution: what do we study and how do we do it.
In this work, we introduce \taskalias (\textit{Testing Reasoning Amid Conflicting Knowledge}), a new benchmark for studying how LLMs propagate new knowledge through multi-step reasoning when it conflicts with the model's initial parametric knowledge.
% More details
Spanning three reasoning-intensive scenarios (\WIKI, \CODE, and \MATH), \taskalias introduces multiple, realistic conflicts to mirror real-world complexity.
% Key results/Observations
Our results on \taskalias reveal that providing updated facts to models for reasoning can worsen performance compared to providing no updated facts to a model, and that this performance degradation exacerbates as more updated facts are provided. We show this failure stems from both inability to faithfully integrate updated facts, but also flawed reasoning even when knowledge is integrated. 
% Meaningfulness
\taskalias provides a rigorous new benchmark to measure and guide future progress on propagating conflicting knowledge in multi-step reasoning.\footnote{Code and data are available at \url{https://github.com/Wind-2375-like/crack}.}
\end{abstract}

% Introduction
\section{Introduction}
\label{sec:main:introduction}

\begin{figure}[t!]
    \begin{center}
        \includegraphics[width=0.45\textwidth]{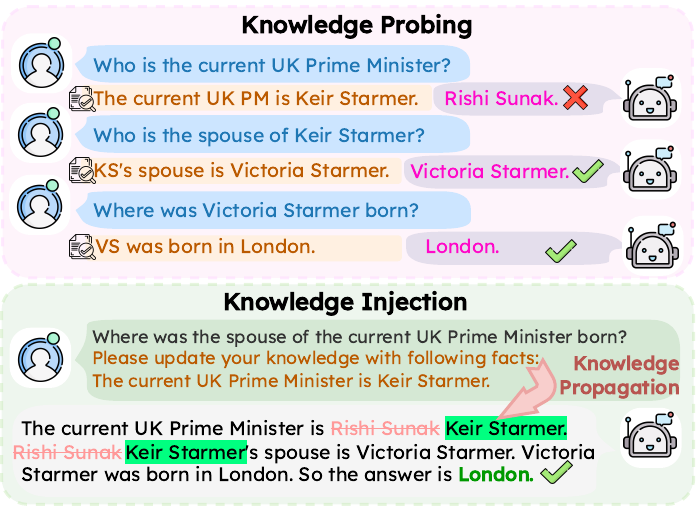}
    \end{center}
    \caption{An illustration of \taskalias's two-stage evaluation framework using a multi-hop QA example. (i) Knowledge Probing: We first decompose a complex question into atomic facts to identify a model's specific knowledge gaps. (ii) Knowledge Injection: We then provide correct facts as conflicting knowledge to test LLM's ability to propagate it within its reasoning.}
\label{fig:small_intro}
\end{figure}

% Set the stakes. Why is the problem important?
Large Language Models (LLMs) demonstrate remarkable reasoning capabilities~\citep{sprague2025to}, leading to their widespread use in scenarios such as multi-hop question answering~\citep{zhu2024fanoutqamultihopmultidocumentquestion}, coding~\citep{jiang2024surveylargelanguagemodels}, and mathematics~\citep{ahn-etal-2024-large,poiroux-etal-2025-reliable}. However, much of these abilities come from their parametric knowledge learned during pretraining, which can become outdated and inaccurate. For example, an LLM might hold outdated information about a current head of state~\citep{zhong-etal-2023-mquake}, or may generate non-executable codes using deprecated function signatures from older version libraries~\citep{liu2025codeupdatearenabenchmarkingknowledgeediting}, or misremember a specific mathematical theorem~\citep{singh2024exposingachillesheelevaluating}. %, thus failing to solve a problem that requires its application. 

To mitigate this issue, a common practice is to provide updated factual knowledge, either through in-context learning \citep{longpre-etal-2021-entity} or by directly editing model parameters \citep{de-cao-etal-2021-editing, mitchell2022fastmodeleditingscale}. However, because the provided new knowledge introduces knowledge conflicts that contradict the model's internal, parametric beliefs~\citep{xu-etal-2024-knowledge-conflicts}, models fail to overwrite their parametric knowledge under such conflicts, which challenges the reliability of multi-step reasoning. Investigating \textit{how effectively LLMs propagate new knowledge through multi-step reasoning amidst such conflicts} offers fundamental insights into how LLMs integrate knowledge and perform reasoning.

% Set the challenge. Why have other papers failed to reach the goal?
% To evaluate the propagation of updated knowledge into complex reasoning in a real-world setting, we need a benchmark that:
% (i) assesses whether a model can faithfully leverage new knowledge in multi-step reasoning, instead of simple fact recall;
% (ii) contains various scenarios with multiple updates, as would be expected in a real world setting, e.g., multiple API and single-hop updates in programming and multi-hop QA tasks, respectively.
% (iii) contains realistic rather than counterfactual world updates, e.g., a new president in a country rather than a city being in a different country than it really is.
% Unfortunately, no existing benchmarks for knowledge updates meet all of these criteria, limiting our ability to research such propagation.
Unfortunately, existing benchmarks are insufficient for evaluating knowledge propagation in multi-step reasoning.
First, they largely focus on simple fact recall~\cite{pmlr-v162-mitchell22a}, rather than assessing whether a model can faithfully leverage new knowledge in multi-step reasoning. Second, they typically evaluate single knowledge updates in isolated scenarios~\cite{xie2024adaptive}, which oversimplifies the real-world challenge of managing multiple, interacting conflicts. Finally, many of these benchmarks rely on counterfactual data (e.g., Eiffel Tower being in Rome), which fails to reflect the complexity of real-world knowledge evolution~\citep{liu2025codeupdatearenabenchmarkingknowledgeediting}, and can also contain unintentional factual errors~\citep{zhong2025mquakeremastered}. %A comprehensive comparison is in \Table~\ref{tab:main:related}. To our best of knowledge, there is no existing benchmark that addresses all these limitations simultaneously.

% Benchmark and Task
To address these limitations, we introduce \taskalias (\textit{Testing Reasoning Amid Conflicting Knowledge}), a new benchmark to evaluate how LLMs propagate conflicting knowledge through multi-step reasoning, illustrated in \Figure~\ref{fig:small_intro}. The benchmark is defined by a novel evaluation framework consisting of two stages. The first \textit{knowledge probing} stage identifies a model's specific knowledge gaps for a given problem. Then, in the \textit{knowledge injection} stage, we provide the updated facts as conflicting knowledge and evaluate if the model can successfully solve a downstream reasoning problem requiring this information. We compare performance between two settings: a closed-book setting where the model relies on its internal knowledge, and an open-book setting where the new facts are provided. Our framework moves beyond simple fact recall and explicitly tests knowledge propagation because each injected fact is a necessary piece of intermediate knowledge, requiring the model to perform further reasoning steps to arrive at the final answer.
% Down
% Our evaluation moves beyond simple fact recall to explicitly test for knowledge propagation. 

% Datasets and Evaluation
We instantiate the \taskalias benchmark with three datasets constructed from realistic conflicts in challenging scenarios: (i) multi-hop QA on recent Wikidata (\WIKI); (ii) code generation with external APIs (\CODE); and (iii) multi-step mathematical reasoning (\MATH).
To evaluate reasoning quality on \taskalias, we introduce novel metrics: \textit{Full Knowledge Entailment} (FKE), which measures if the reasoning is faithful to all facts required for the solution, and \textit{Holistic Pass} (HP), a strict metric requiring both a correct answer and a faithful reasoning chain. %These metrics allow us to measure LLM's capacity of both knowledge integration and logical reasoning.

% Evidence
Our experiments across a wide range of LLMs, including open-source and closed-source, and thinking and non-thinking, models reveal that current models struggle significantly on \taskalias. Specifically, we find that: (i) providing models with correct, conflicting facts (the open-book setting) yields surprisingly limited gains and can even backfire, with performance sometimes falling below the closed-book baseline; (ii) performance degrades as more conflicting facts are provided; and (iii) this failure stems from both an inability to faithfully integrate the new facts, but also flawed reasoning even when knowledge integration is successful.

Our contributions are threefold. First, we formalize a novel evaluation framework for rigorously measuring reasoning under knowledge conflicts, centered on a two-stage methodology of knowledge probing and injection. Second, we introduce~\taskalias, a new benchmark that instantiates this framework with three diverse datasets (\WIKI, \CODE, \MATH) featuring realistic, multi-fact conflicts, and novel evaluation metrics (FKE, HP) for a more fine-grained analysis. Finally, we conduct a comprehensive experimental analysis on a wide range of LLMs, revealing a critical failure where providing correct conflicting facts shows limited gains and even degrades performance, rooted in both unfaithful knowledge integration and flawed reasoning.

\begin{table*}[t!]
    \centering
    \resizebox{0.9\textwidth}{!}{
    \begin{tabular}{p{3.5cm}C{2.5cm}C{2.5cm}C{2.5cm}C{2.5cm}C{2.5cm}C{2.5cm}C{2.5cm}}
    \toprule
    \multirow{2}{*}{\textbf{Benchmark}} & 
    \multirow{2}{*}{\makecell[tc]{\textbf{Knowledge} \\ \textbf{Propagation}}} & 
    \multirow{2}{*}{\makecell[tc]{\textbf{Multiple} \\ \textbf{Conflicts}}} & 
    \multicolumn{3}{c}{\textbf{Reasoning Scenarios}} & 
    \multirow{2}{*}{\makecell[tc]{\textbf{Real-World} \\ \textbf{Knowledge}}} & 
    \multirow{2}{*}{\textbf{Size}} \\ 
    \cline{4-6}
    & & & 
    \makecell[c]{\textbf{Multi-Hop QA}} & 
    \makecell[c]{\textbf{Coding}} & 
    \makecell[c]{\textbf{Math}} & & \\
    \midrule
    \multicolumn{8}{c}{\textit{Knowledge Conflict Benchmarks}} \\
    \midgrayline
    \cite{longpre-etal-2021-entity} & \bempty & \bempty & \bfull & \bempty & \bempty & \bempty & 315,203 \\
    \cite{xie2024adaptive} & \bempty & \bempty & \bfull & \bempty & \bempty & \bempty & 16,557 \\
    \cite{kortukov2024studying} & \bfull & \bfull & \bfull & \bempty & \bempty & \bfull & 58,281 \\
    \cite{wang2024resolving} & \bempty & \bempty & \bfull & \bempty & \bempty & \bempty & 9,083 \\
    \cite{xu-etal-2024-earth} & \bempty & \bempty & \bfull & \bempty & \bempty & \bempty & 1,500 \\
    \cite{wu2024clasheval} & \bempty & \bempty & \bfull & \bempty & \bempty & \bhalf & 1,278 \\
    \cite{ying-etal-2024-intuitive} & \bempty & \bfull & \bempty & \bempty & \bfull & \bhalf & 11,684 \\
    \cite{feng2025unravelingmisinformationpropagationllm} & \bfull & \bfull & \bempty & \bempty & \bfull & \bempty & 400 \\ 
    \midrule
    \multicolumn{8}{c}{\textit{Knowledge Editing Benchmarks}} \\
    \midgrayline
    \cite{pmlr-v162-mitchell22a} & \bempty & \bfull & \bfull & \bempty & \bempty & \bfull & 15,000 \\
    \cite{zhong-etal-2023-mquake} & \bfull & \bfull & \bfull & \bempty & \bempty & \bhalf & 11,086 \\
    \cite{onoe-etal-2023-lms} & \bfull & \bempty & \bfull & \bempty & \bempty & \bfull & 1,000 \\
    \cite{10.1162/tacl_a_00644} & \bfull & \bempty & \bfull & \bempty & \bempty & \bempty & 5,000 \\
    \cite{hua-etal-2024-propagation} & \bfull & \bempty & \bfull & \bempty & \bempty & \bempty & 5,010 \\
    \cite{liu2025codeupdatearenabenchmarkingknowledgeediting} & \bfull & \bempty & \bempty & \bfull & \bempty & \bempty & 670 \\
    \cite{huang2025can} & \bfull & \bempty & \bfull & \bempty & \bempty & \bfull & 40,195 \\
    \cite{rosati-etal-2024-long} & \bfull & \bempty & \bfull & \bempty & \bempty & \bempty & 7,389 \\
    \cite{thede2025wikibigedit} & \bfull & \bempty & \bfull & \bempty & \bempty & \bfull & 502,382 \\
    \midrule
    \multicolumn{8}{c}{\textit{Our Benchmark: Testing Reasoning Amid Conflicting Knowledge}~(\taskalias)} \\
    \midgrayline
    \taskalias~(\textbf{ours}) & \bfull & \bfull & \bfull & \bfull & \bfull & \bfull & \textbf{1,500} \\
    \bottomrule
\end{tabular}
    }
    \caption{\textbf{Comparison between \protect\taskalias and prior knowledge conflict and editing benchmarks.} Our benchmark, \protect\taskalias, (i) requires true knowledge propagation\protect\footnotemark to achieve multi-step reasoning; (ii) presents multiple conflicts and spans multiple scenarios (multi-hop QA, coding, and math); and (iii) uses conflicts derived from real-world knowledge rather than counterfactuals. Symbols denote full (\bfull), partial (\bhalf), or no (\bempty) support for a feature.}
    \label{tab:main:related}
\end{table*}
% \footnotetext{Note that a task requiring reasoning does not necessarily test knowledge propagation. We define knowledge propagation as requiring inference where the new fact is an intermediate step in the reasoning, distinct from cases where a provided fact simply dictates the final answer. For example, \citep{ying-etal-2024-intuitive} do not require knowledge propagation though their task is machine reasoning comprehension, because the conflicting fact is the final answer itself (their Figure 2), which only tests a model's ability to recall, not to reason with new information.}

% Related Work
\section{Related Work}
\label{sec:main:related}

% Our work is the intersection of two related fields: knowledge conflicts and knowledge editing. As detailed in \Table~\ref{tab:main:related}, we compare \taskalias with benchmarks from both areas.

\SmallHeading{Knowledge Conflict}
We define knowledge conflicts as discrepancies between internal knowledge and external updates (context or edits), distinct from inter-context or intra-memory conflicts \cite{xu-etal-2024-knowledge-conflicts}. Since \citet{longpre-etal-2021-entity} introduced this domain, benchmarks have analyzed preferences between internal and external knowledge \citep{xie2024adaptive, wu2024clasheval}, misinformation susceptibility \citep{xu-etal-2024-earth}, and conflict resolution \citep{wang2024resolving}, yet generally neglect knowledge propagation, the use of new facts in multi-step reasoning. While recent studies explore propagation \citep{feng2025unravelingmisinformationpropagationllm} and real-world conflicts \cite{kortukov2024studying}, they remain limited in counterfactual data and reasoning variability. \taskalias addresses these gaps by employing multiple realistic conflicts from various scenarios to evaluate complex knowledge propagation.

% We define knowledge conflicts as discrepancies between internal knowledge and external updates (via context or parameter edits), an extension of context-memory conflicts, which differ from inter-context (conflicts within external updates) and intra-memory conflicts (conflicts within internal knowledge) \cite{xu-etal-2024-knowledge-conflicts}.
% \citet{longpre-etal-2021-entity} first introduced knowledge conflicts in the context of question answering, where contextual information contradicts a model's parametric beliefs. 
% Subsequent benchmarks investigate such contradiction between a model's internal prior and external evidence and analyze model preferences~\citep{xie2024adaptive, wu2024clasheval}, their susceptibility to misinformation~\citep{xu-etal-2024-earth}, and propose methods for conflict resolution~\citep{wang2024resolving}. However, these works typically do not evaluate knowledge propagation: a model's ability to integrate a new fact and leverage it in subsequent, multi-step reasoning. While some recent work has started to explore how misinformation propagates through reasoning~\citep{feng2025unravelingmisinformationpropagationllm} and how LLMs handle real-world knowledge conflicts in multi-hop QA~\cite{kortukov2024studying}, these studies are limited in counterfactual data or reasoning scenarios. \taskalias addresses these gaps by introducing multiple, realistic conflicts and shifting the focus to the more complex knowledge propagation challenge.

\SmallHeading{Knowledge Editing}
Knowledge editing aims to efficiently update a model's parametric knowledge~\citep{yao-etal-2023-editing}. A wide range of methods have been proposed, from early hypernetwork-based approaches~\citep{de-cao-etal-2021-editing, hase2021languagemodelsbeliefsmethods} and influential locate-then-edit techniques like ROME/MEMIT~\citep{meng2023locatingeditingfactualassociations, meng2023masseditingmemorytransformer}, to demonstrating that standard fine-tuning can be a strong baseline~\citep{gangadhar-stratos-2024-model}. However, the conceptual foundations of this field are heavily debated. Studies have questioned the core assumptions of localization~\citep{hase2023does}, pointed out fundamental problems with the belief revision paradigm itself~\citep{hase2024fundamental}, and highlighted unintended side effects, such as the amplification of social biases~\citep{halevy-etal-2024-flex}. Consequently, the evaluation of editing has also matured, with new benchmarks assessing multi-hop reasoning~\cite{zhong-etal-2023-mquake, onoe-etal-2023-lms, hua-etal-2024-propagation}, coding~\cite{liu2025codeupdatearenabenchmarkingknowledgeediting}, performance on long-form generation~\citep{rosati-etal-2024-long}, the correction of verified real-world hallucinations~\citep{huang2025can}, and large-scale lifelong updates~\citep{thede2025wikibigedit}. However, existing work is often limited to single reasoning scenarios or conflict types. In contrast, \taskalias provides the first unified benchmark to test propagation over multiple, realistic conflicts across three diverse domains. \Table~\ref{tab:main:related} presents a comprehensive comparison of these existing datasets.
% Knowledge editing aims to update a model's parametric knowledge to reflect new information~\citep{meng2023locatingeditingfactualassociations, pmlr-v162-mitchell22a}. While traditionally focused on simple fact recall, a more challenging frontier is now knowledge propagation: the ability to reason with newly acquired facts~\citep{zhong-etal-2023-mquake, onoe-etal-2023-lms, 10.1162/tacl_a_00644, liu2025codeupdatearenabenchmarkingknowledgeediting}. Although recent benchmarks have begun to evaluate this capability~\citep{zhong-etal-2023-mquake, onoe-etal-2023-lms, 10.1162/tacl_a_00644, liu2025codeupdatearenabenchmarkingknowledgeediting}, they remain limited because they are often constrained to a single reasoning scenario, rely on counterfactual data, or test only single conflicts. In contrast, \taskalias is the first to simultaneously require reasoning over multiple, realistic conflicts across three diverse scenarios, and it employs a fine-grained evaluation framework to evaluate both final answer correctness and faithfulness of knowledge to reasoning process.

% Problem
\section{Testing Reasoning Amid Conflicting Knowledge (\taskalias)}
\label{sec:main:task}

\begin{figure*}[t!]
    \begin{center}
        \includegraphics[width=0.94\textwidth]{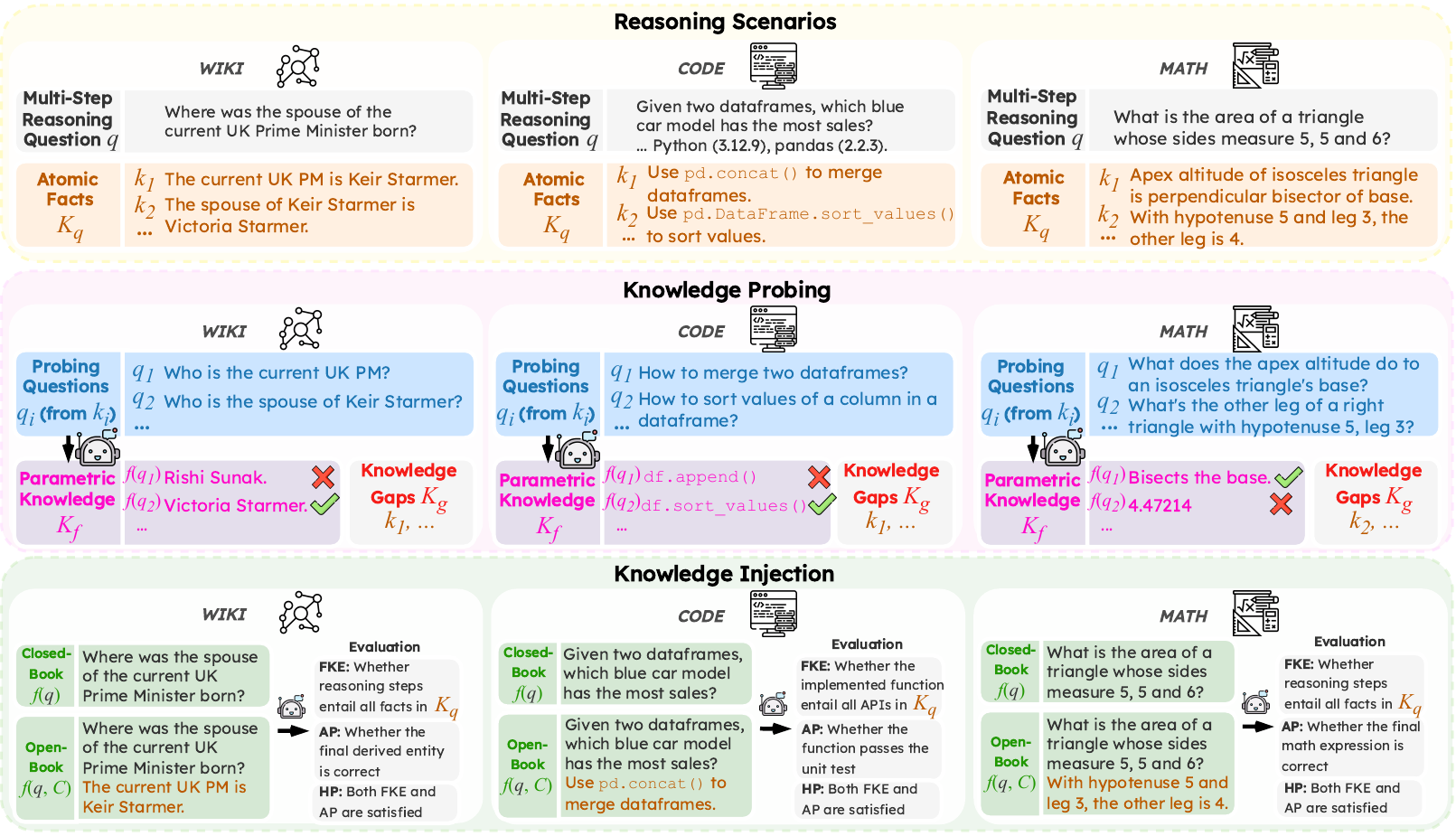}
    \end{center}
    \caption{\textbf{Full illustration of the \taskalias benchmark across our three reasoning scenarios}: Multi-Hop QA (\WIKI), Code Generation (\CODE), and Mathematical Reasoning (\MATH). The benchmark follows a two-stage process. (i) Knowledge Probing: we identify the model's knowledge gaps by testing it on the required atomic facts, the individual pieces of knowledge needed to solve the complex question. (ii) Knowledge Injection: we evaluate reasoning by comparing a closed-book setting (using only the model's internal knowledge) with an open-book setting (where the identified knowledge gaps are provided). Performance is assessed using our metrics: Answer Pass (AP), Full Knowledge Entailment (FKE), and Holistic Pass (HP).}
\label{fig:intro}
\end{figure*}

In this section, we introduce our benchmark, \textit{Testing Reasoning Amid Conflicting Knowledge} (\taskalias), to evaluate how LLMs perform multi-step reasoning when faced with knowledge conflicts. Specifically, we outline our problem formulation (\secref{sec:main:task:definition}), three instantiated scenarios (\secref{sec:main:task:scenarios}), and evaluation framework (\secref{sec:main:task:metrics}). The overall pipeline of our benchmark is illustrated in \Figure~\ref{fig:intro}.

\subsection{Problem Setup}
\label{sec:main:task:definition}

We investigate how an LLM $f$ reasons when newly introduced facts contradict its parametric knowledge. The benchmark comprises two stages: (i) knowledge probing to identify the model's knowledge gaps, and (ii) knowledge injection to integrate new relevant facts into reasoning.

\SmallHeading{Knowledge Probing}
Given a question $q$, we first decompose it into a set of atomic facts, $K_q = \{k_1, \cdots, k_n\}$, required to answer the question. To identify the model's knowledge gap ($K_g$), we convert each fact $k_i \in K_q$ into a probing question-answer pair $(q_i, a_i)$ and query the model. The knowledge gap is then defined as the subset of facts unknown by a model (detailed in \secref{sec:main:task:metrics}). $K_g$ pinpoints exact locations where providing correct external information will induce a knowledge conflict, making it necessary for the subsequent injection stage.

\footnotetext{We define knowledge propagation as using a new fact as a premise in a reasoning chain, not simply recalling a fact that is the final answer. For instance, the conflicting facts in \citet{ying-etal-2024-intuitive} are final answers (see their Fig. 2), thus testing recall rather than propagation.}

\SmallHeading{Knowledge Injection}
This stage evaluates how the model $f$ integrates facts from the knowledge gap $K_g$, generating a final answer $\hat{a}$ and reasoning $\hat{R}$. We compare a \textit{closed-book} baseline where the model only receives the question $q$, denoted $(\hat{R}_p, \hat{a}_p)=f(q)$; and an \textit{open-book} setting where the question is accompanied by a set of new information $C$ derived from $K_g$, denoted $(\hat{R}_c, \hat{a}_c)=f(q, C)$. To simulate shared update environment where models must discern relevant facts from concurrent inputs, the \textit{knowledge aggregation scope} (KAS) parameter controls how this set of information $C$ is constructed. We aggregate knowledge gaps from a batch $\mathcal{B}$ of $\text{KAS}$ questions to construct the set: $C_\mathcal{B} = \bigcup_{q \in \mathcal{B}} K_g$. For example, A $\text{KAS}$ of 1 is instance-specific (each question receives only its own missing facts), while a $\text{KAS}$ of 10 requires the model to reason using the combined knowledge from 10 different questions. We retain samples without knowledge gaps to ensure a fixed test set and avoid inductive bias \cite{si-etal-2023-measuring}. As shown in \Appendix~\ref{sec:appendix:results}, excluding them has negligible impact on scores and leaves conclusions unchanged.

\subsection{\taskalias Scenarios}
\label{sec:main:task:scenarios}

We instantiate the \taskalias~benchmark across three diverse reasoning scenarios (\WIKI, \CODE, \MATH).

\SmallHeading{\WIKI: Wikipedia Multi-Hop QA} A complex question $q$ like ``Where was the spouse of the current UK Prime Minister born?'' requires reasoning over a chain of relational facts from Wikidata~\cite{vrandevcic2014wikidata}. Each relation (e.g., ``The current UK Prime Minister is Keir Starmer.'') serves as an atomic fact $k_i$. A conflict arises from the model's outdated world knowledge (e.g., believing Rishi Sunak is the PM).

\SmallHeading{\CODE: Code Generation with External APIs} Given a coding problem $q$ like ``Given two dataframes, find which blue car model has the most sales? ... Python (3.12.9), pandas (2.2.3).'', the model must generate a correct function. Each atomic fact $k_i$ is the proper usage of an external API. Outdated API knowledge can introduce conflicts, such as using a deprecated function (e.g., \texttt{df.append}) instead of its modern equivalent (\texttt{pd.concat}).

\SmallHeading{\MATH: Multi-Step Mathematical Reasoning} For a standard math problem $q$ like ``What is the area of a triangle whose sides measure 5, 5, and 6?'', each atomic fact $k_i$ is a procedural step (``What's the other leg of a right triangle with hypotenuse 5 and leg 3?''). A conflict arises when the model's internal parametric knowledge contains an incorrect or hallucinated procedural step.

\subsection{Evaluation Metrics}
\label{sec:main:task:metrics}

We evaluate each of the two stages of the \taskalias benchmark. Details are in \Appendix~\ref{sec:appendix:evaluation}.

\SmallHeading{Evaluate Knowledge Probing}
For each atomic fact, we sample 10 responses to its probing question. Using an LLM judge (GPT-5-mini),\footnote{We use \texttt{gpt-5-mini-0807} for all following GPT-5-mini models.} we count the number of responses equivalent to the ground-truth answer. A fact is considered \textit{known} if its correct answer is the most frequent response; otherwise, it is \textit{unknown}. We define \textit{knowledge confidence} (KConf) as the proportion of correct responses.

\SmallHeading{Evaluate Knowledge Injection}
We assess the model's reasoning output using three metrics: 
(i) \textit{Answer Pass} (AP) is a binary metric for the final answer's correctness. % $\text{AP} = \text{Is\_Correct}(\hat{a}, a)$. 
Answer correctness check is scenario-specific: an equivalence check for \WIKI and \MATH using GPT-5-mini, and Pass@1~\citep{chen2021evaluatinglargelanguagemodels} for \CODE,
(ii) \textit{Full Knowledge Entailment} (FKE) is a binary metric checking if the reasoning is faithful to all required atomic facts. FKE holds if the reasoning chain entails every fact in the required set. (iii) \textit{Holistic Pass} (HP) is a stricter metric requiring both correctness (Answer Pass) and faithfulness (Full Knowledge Entailment).

% Benchmark
\section{\taskalias Benchmark Datasets}
\label{sec:main:dataset}

The \taskalias benchmark comprises 1,500 examples, with 500 for each of our three reasoning scenarios (\WIKI, \CODE, \MATH). We constructed these scenarios by sourcing and adapting data from Wikidata~\citep{vrandevcic2014wikidata}, BigCodeBench~\cite{zhuo2024bigcodebench}, and PRM800K~\cite{lightman2023lets}, respectively. Each example contains five key components (\figref{fig:data}): a reasoning question ($q$), its required atomic facts ($K_q$), corresponding probing question-answer pairs ($q_i$, $a_i$), and the final answer ($a$). The detailed data generation pipeline, which combines heuristic rules and the use of GPT-5-mini, is detailed in \Appendix~\ref{sec:appendix:dataset}.
Note that our pipeline supports regenerating the benchmark with future knowledge to ensure its continued relevance for evaluating new models. For example, researchers can run the pipeline on new Wikipedia dumps to generate recent \WIKI~questions, use it to parse up-to-date code pieces for new \CODE~challenges, and source \MATH~problems from recent competitions.

We install rigorous quality controls to ensure the factuality of probing question-answer pairs and the necessity of atomic facts. To enforce factuality, we filter out all non-factual probing pairs and corresponding atomic facts using GPT-5-mini. Additionally, \WIKI questions are time-stamped (e.g., ``By Sep. 2025''), and \CODE problems specify library versions (e.g., ``Python (3.12.9), pandas (2.2.3)''). For necessity, we compose \WIKI questions directly from the underlying fact chains, while \CODE and \MATH problems explicitly instruct the model to use the functions or procedures derived from the provided atomic facts. We further validated our data by checking whether each probing question-answer pair ($q_i, a_i$) is factual and each fact in atomic facts $K_q$ is necessary in the multi-step reasoning question $q$. Using both \texttt{gpt-5-mini} and \texttt{gemini-2.5-pro}\footnote{Released on June 17, 2025} as judges, two models annotate 94.3\% as factual and 88.0\% as necessary on average, and their annotations have high average F1 scores with one human annotator (98.0\% for factuality and 93.5\% for necessity). Details of the quality check are in \Appendix~\ref{sec:appendix:checklist:ai}.
% LLM-as-a-judge and human evaluation, achieving high annotation quality ($\text{F1}=$\todo{}) and inter-annotator agreement (Cohen's $\kappa=$\todo{}).

While automated pipelines guarantee factuality, we ensure high reasoning variability both across and within domains. Cross-domain structures span graph traversal (WIKI), procedural composition (CODE), and symbolic derivation (MATH), while within-domain diversity arises from adapting sources like BigCodeBench and PRM800K to create varied reasoning chains. As highlighted by the statistics in \tabref{tab:dataset_stats}, this diversity in reasoning complexity and question length ensures \taskalias provides a comprehensive evaluation of model capabilities across various scenarios.

\begin{table}[t!]
\centering
\resizebox{0.49\textwidth}{!}{
\begin{tabular}{p{5cm}C{1.2cm}C{1.2cm}C{1.2cm}}
\toprule
\textbf{Statistic} & \textbf{\GROW} & \textbf{\CODE} & \textbf{\MATH} \\
\midrule
\# Multi-Step Questions ($q$) & 500 & 500 & 500 \\
Avg. \# Atomic Facts ($k_i$) & 4.00 & 3.82 & 5.16 \\
Avg. Tokens in Question ($q$) & 38.02 & 198.68 & 97.02 \\
Avg. Tokens in Probing Q ($q_i$) & 19.80 & 46.88 & 28.12 \\
Avg. Tokens in Atomic Fact ($k_i$) & 12.32 & 26.17 & 39.62 \\
\bottomrule
\end{tabular}
}
\caption{Key statistics for the constructed datasets.}
\label{tab:dataset_stats}
\end{table}

% Method
% \section{Meta-COT}
% \label{sec:main:method}

% Experiments
\begin{figure*}[t!]
    \centering
    \begin{subfigure}[b]{0.27\textwidth}
        \centering
        \includegraphics[width=\textwidth]{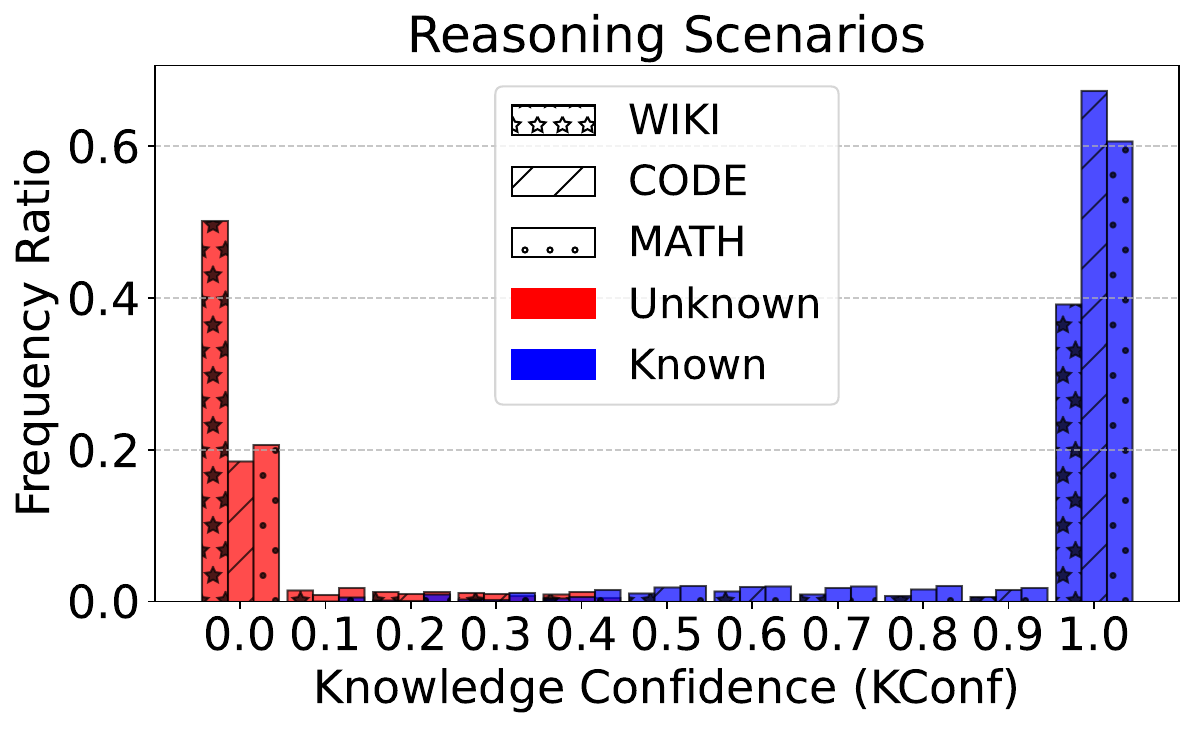}
        \caption{Reasoning scenarios.}
        \label{fig:kconf_by_scenario}
    \end{subfigure}
    \hfill
    \begin{subfigure}[b]{0.7\textwidth}
        \centering
        \includegraphics[width=\textwidth]{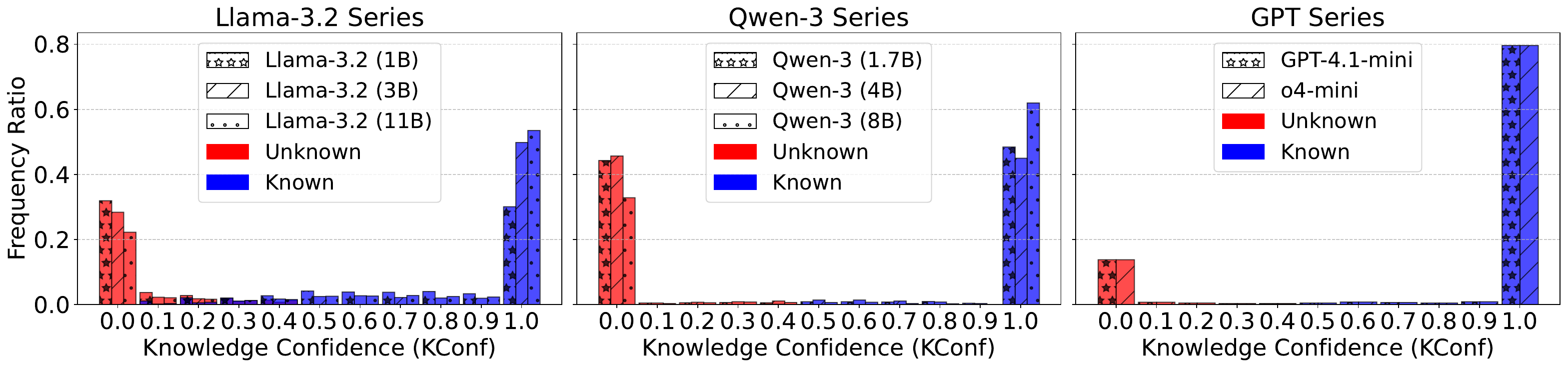}
        \caption{Model series.}
        \label{fig:kconf_by_model}
    \end{subfigure}
    
    \caption{\textbf{Knowledge Confidence (KConf) distributions} of Known and Unknown facts across (\subref{fig:kconf_by_scenario}) reasoning scenarios, and (\subref{fig:kconf_by_model}) model series. A consistent color scheme is used across both panels (\textcolor{blue}{Blue}: Known facts, \textcolor{red}{Red}: Unknown facts).}% We show that (\textit{i}) LLMs exhibit polarized confidence distributions, either recalling facts with full confidence or not at all. (\textit{ii}) The distribution of Known and Unknown facts varies across different scenarios and models.}
    \label{fig:kconf_distributions}
\end{figure*}

\section{Knowledge Probing}
\label{sec:main:probing}

In the knowledge probing stage, we identify each model's internal knowledge gaps. These identified gaps, which we release as a new diagnostic dataset, are necessary for preparing model-specific knowledge conflicts evaluated in the knowledge injection stage. This section outlines our probing setup (\secref{sec:main:probing:setup}) and analyzes the results (\secref{sec:main:probing:results}).

\subsection{Experimental Setup}
\label{sec:main:probing:setup}

\SmallHeading{Backbone LLMs}
We select a wide range of LLMs, including both open-source and closed-source, thinking and non-thinking models, across various scales. Specifically, we test Llama-3.2 (1B, 3B, 11B)\footnote{The Llama-3.2 11B is a multimodal model.}~\citep{grattafiori2024llama}, Qwen-3 (1.7B, 4B, 8B)~\citep{yang2025qwen3technicalreport}, GPT-4.1-mini\footnote{We use \texttt{gpt-4.1-mini-2025-04-14}.}~\citep{openai2025gpt41}, and o4-mini\footnote{We use \texttt{o4-mini-2025-04-16}.}~\citep{openai2025o4}.

\SmallHeading{Knowledge Probing}
To identify each model's knowledge gaps, we probe it with questions based on atomic facts. For each question, we generate $M=10$ responses. Detailed prompts and model-specific hyperparameters are in Appendix~\ref{sec:appendix:setup:probing}.

\subsection{Knowledge Probing Results}
\label{sec:main:probing:results}

% \takeaways{
% (\textit{i}) LLMs exhibit polarized confidence distributions, either recalling facts with full confidence or not at all, with little uncertainty.
% (\textit{ii}) The distribution of Known versus Unknown facts varies across different scenarios and models.
% }

\SmallHeading{Models exhibit polarized KConf}
As shown in \Figure~\ref{fig:kconf_distributions}, models exhibit a highly polarized knowledge confidence (KConf). The KConf for known facts is sharply concentrated at 1, while the confidence for unknown facts is skewed towards 0. This polarization indicates that models are rarely uncertain: they either know a fact with high confidence or are clearly aware they do not know it at all.

\SmallHeading{KConf distributions vary across scenarios and models}
The distribution of known versus unknown facts also varies significantly by both scenario (\Figure~\ref{fig:kconf_by_scenario}) and model (\Figure~\ref{fig:kconf_by_model}). The \WIKI~scenario, built from up-to-date Wikidata (Sep. 2025), contains the most unknown facts likely due to many of its facts post-dating the models' knowledge cutoffs.\footnote{Llama-3.2: Dec. 2023; Qwen-3: Sep. 2024; GPT-4.1-mini and o4-mini: Jun. 2024} In contrast, \CODE~and \MATH~contain more known facts, as they are reconstructed from established benchmarks likely included in the models' pre-training data. Knowledge also differs by model family and scale, with closed-source models (GPT series) and larger scales (e.g., Llama-3.2 11B v.s. 1B) generally possessing more known facts.
% For instance, Llama-3.2-3b knows more facts than its 1B counterpart, but the trend is not uniform across families, as \todo{the Qwen-3 series} does not show a proportional increase in Known facts over \todo{its smaller versions}. This variance underscores that a model's performance on a task is deeply tied to the specific knowledge required, which is not always correlated with scale.

\begin{table*}[t!]
    \centering
    \resizebox{0.93\textwidth}{!}{
    \begin{tabular}{p{2.5cm}p{2cm}p{1.5cm}p{1.5cm}p{1.5cm}p{1.5cm}p{1.5cm}p{1.5cm}p{1.5cm}p{1.5cm}p{1.5cm}}
\toprule
\multirow{2}{*}{\textbf{Backbone Model}} & \multirow{2}{*}{\textbf{Method}} & \multicolumn{3}{c}{\textbf{WIKI}} & \multicolumn{3}{c}{\textbf{CODE}} & \multicolumn{3}{c}{\textbf{MATH}} \\ 
\cmidrule(r){3-5} \cmidrule(l){6-8} \cmidrule(l){9-11}
 &  & \makecell[c]{\textbf{HP}} & \makecell[c]{\textbf{AP}} & 
 \makecell[c]{\textbf{FKE}} & \makecell[c]{\textbf{HP}} & 
 \makecell[c]{\textbf{AP}} & \makecell[c]{\textbf{FKE}} & \makecell[c]{\textbf{HP}} & \makecell[c]{\textbf{AP}} & \makecell[c]{\textbf{FKE}} \\ \midrule
\multirow{4}{*}{Llama-3.2 \textsubscript{(1B)}} & Base Model & 0.9 \textsubscript{$\pm$ 0.7} & 6.1 \textsubscript{$\pm$ 2.1} & 0.9 \textsubscript{$\pm$ 0.7} & 5.7 \textsubscript{$\pm$ 2.1} & 9.2 \textsubscript{$\pm$ 2.6} & 36.4 \textsubscript{$\pm$ 4.2} & 20.0 \textsubscript{$\pm$ 3.4} & 26.0 \textsubscript{$\pm$ 4.0} & 26.3 \textsubscript{$\pm$ 3.7} \\
\cmidrule(l){2-11}
& Append & 1.1 \textsubscript{$\pm$ 0.9} & 7.6 \textsubscript{$\pm$ 2.4} & 1.0 \textsubscript{$\pm$ 0.8} & 4.1 \textsubscript{$\pm$ 1.7} & 7.0 \textsubscript{$\pm$ 2.4} & 38.5 \textsubscript{$\pm$ 4.3} & 22.5 \textsubscript{$\pm$ 3.7} & 27.8 \textsubscript{$\pm$ 3.8} & 28.6 \textsubscript{$\pm$ 4.0} \\
& FT-CK & 0.9 \textsubscript{$\pm$ 0.7} & 6.8 \textsubscript{$\pm$ 2.2} & 0.9 \textsubscript{$\pm$ 0.7} & 5.1 \textsubscript{$\pm$ 1.9} & 8.5 \textsubscript{$\pm$ 2.3} & 36.5 \textsubscript{$\pm$ 4.3} & 21.8 \textsubscript{$\pm$ 3.6} & 28.2 \textsubscript{$\pm$ 3.8} & 28.1 \textsubscript{$\pm$ 3.9} \\
& MeLLo & 2.4 \textsubscript{$\pm$ 1.2} & 11.5 \textsubscript{$\pm$ 2.7} & 2.4 \textsubscript{$\pm$ 1.4} & 3.9 \textsubscript{$\pm$ 1.7} & 5.4 \textsubscript{$\pm$ 2.0} & 33.3 \textsubscript{$\pm$ 4.3} & 18.0 \textsubscript{$\pm$ 3.2} & 24.5 \textsubscript{$\pm$ 3.7} & 24.6 \textsubscript{$\pm$ 3.8} \\
\midrule
\multirow{4}{*}{Llama-3.2 \textsubscript{(3B)}} & Base Model & 5.3 \textsubscript{$\pm$ 1.9} & 23.1 \textsubscript{$\pm$ 3.5} & 5.6 \textsubscript{$\pm$ 2.0} & 15.3 \textsubscript{$\pm$ 3.1} & 22.0 \textsubscript{$\pm$ 3.4} & 54.7 \textsubscript{$\pm$ 4.5} & 40.5 \textsubscript{$\pm$ 4.1} & 52.9 \textsubscript{$\pm$ 4.1} & 44.6 \textsubscript{$\pm$ 4.2} \\
\cmidrule(l){2-11}
& Append & 7.3 \textsubscript{$\pm$ 2.1} & 23.3 \textsubscript{$\pm$ 3.7} & 7.8 \textsubscript{$\pm$ 2.4} & 16.8 \textsubscript{$\pm$ 3.2} & 23.3 \textsubscript{$\pm$ 3.7} & 52.9 \textsubscript{$\pm$ 4.5} & 40.4 \textsubscript{$\pm$ 4.2} & 52.7 \textsubscript{$\pm$ 4.3} & 43.9 \textsubscript{$\pm$ 4.3} \\
& FT-CK & 7.4 \textsubscript{$\pm$ 2.2} & 23.0 \textsubscript{$\pm$ 3.6} & 8.2 \textsubscript{$\pm$ 2.4} & 12.8 \textsubscript{$\pm$ 2.8} & 20.1 \textsubscript{$\pm$ 3.5} & 54.8 \textsubscript{$\pm$ 4.4} & 41.6 \textsubscript{$\pm$ 4.2} & 53.7 \textsubscript{$\pm$ 4.3} & 46.1 \textsubscript{$\pm$ 4.3} \\
& MeLLo & 7.3 \textsubscript{$\pm$ 2.3} & 19.3 \textsubscript{$\pm$ 3.3} & 8.2 \textsubscript{$\pm$ 2.2} & 12.2 \textsubscript{$\pm$ 2.8} & 17.4 \textsubscript{$\pm$ 3.2} & 45.1 \textsubscript{$\pm$ 4.7} & 37.9 \textsubscript{$\pm$ 4.3} & 47.5 \textsubscript{$\pm$ 4.7} & 40.5 \textsubscript{$\pm$ 4.3} \\
\midrule
\multirow{4}{*}{Llama-3.2 \textsubscript{(11B)}} & Base Model & 12.5 \textsubscript{$\pm$ 2.9} & 32.6 \textsubscript{$\pm$ 4.0} & 12.9 \textsubscript{$\pm$ 2.9} & 18.7 \textsubscript{$\pm$ 3.3} & 29.3 \textsubscript{$\pm$ 3.9} & 50.5 \textsubscript{$\pm$ 4.5} & 34.8 \textsubscript{$\pm$ 4.0} & 40.4 \textsubscript{$\pm$ 4.2} & 48.8 \textsubscript{$\pm$ 4.4} \\
\cmidrule(l){2-11}
& Append & 13.1 \textsubscript{$\pm$ 2.9} & 34.0 \textsubscript{$\pm$ 4.2} & 13.2 \textsubscript{$\pm$ 3.0} & 18.1 \textsubscript{$\pm$ 3.3} & 27.1 \textsubscript{$\pm$ 3.7} & 51.9 \textsubscript{$\pm$ 4.5} & 33.4 \textsubscript{$\pm$ 4.0} & 40.2 \textsubscript{$\pm$ 4.4} & 46.5 \textsubscript{$\pm$ 4.5} \\
& FT-CK & 12.7 \textsubscript{$\pm$ 2.9} & 35.2 \textsubscript{$\pm$ 4.4} & 12.7 \textsubscript{$\pm$ 2.9} & 19.1 \textsubscript{$\pm$ 3.5} & 28.1 \textsubscript{$\pm$ 3.9} & 51.9 \textsubscript{$\pm$ 4.5} & 32.6 \textsubscript{$\pm$ 4.2} & 38.7 \textsubscript{$\pm$ 4.3} & 49.6 \textsubscript{$\pm$ 4.4} \\
& MeLLo & 11.0 \textsubscript{$\pm$ 2.6} & 22.0 \textsubscript{$\pm$ 3.6} & 12.4 \textsubscript{$\pm$ 2.8} & 9.4 \textsubscript{$\pm$ 2.6} & 17.1 \textsubscript{$\pm$ 3.3} & 25.9 \textsubscript{$\pm$ 3.9} & 20.3 \textsubscript{$\pm$ 3.3} & 23.4 \textsubscript{$\pm$ 4.0} & 36.3 \textsubscript{$\pm$ 4.1} \\
\midrule
\multirow{5}{*}{Qwen-3 \textsubscript{(1.7B)}} & Base Model & 4.5 \textsubscript{$\pm$ 1.7} & 17.1 \textsubscript{$\pm$ 3.1} & 4.5 \textsubscript{$\pm$ 1.7} & 17.4 \textsubscript{$\pm$ 3.2} & 24.3 \textsubscript{$\pm$ 3.9} & 54.4 \textsubscript{$\pm$ 4.2} & 48.3 \textsubscript{$\pm$ 4.3} & 58.6 \textsubscript{$\pm$ 4.2} & 64.0 \textsubscript{$\pm$ 4.4} \\
\cmidrule(l){2-11}
& Append & \textbf{83.6} \textsubscript{$\pm$ 3.2} & 90.9 \textsubscript{$\pm$ 2.5} & \textbf{87.2} \textsubscript{$\pm$ 3.0} & 18.3 \textsubscript{$\pm$ 3.3} & 26.3 \textsubscript{$\pm$ 3.7} & 59.6 \textsubscript{$\pm$ 4.2} & 49.2 \textsubscript{$\pm$ 4.0} & 57.6 \textsubscript{$\pm$ 4.6} & 66.8 \textsubscript{$\pm$ 4.0} \\
& Append-T & 6.4 \textsubscript{$\pm$ 2.2} & 51.2 \textsubscript{$\pm$ 4.2} & 6.2 \textsubscript{$\pm$ 2.2} & 15.5 \textsubscript{$\pm$ 3.1} & 23.1 \textsubscript{$\pm$ 3.7} & 35.1 \textsubscript{$\pm$ 4.1} & 7.1 \textsubscript{$\pm$ 2.3} & 75.6 \textsubscript{$\pm$ 3.8} & 19.3 \textsubscript{$\pm$ 3.5} \\
& FT-CK & 4.8 \textsubscript{$\pm$ 1.8} & 16.8 \textsubscript{$\pm$ 3.2} & 4.7 \textsubscript{$\pm$ 1.7} & 16.7 \textsubscript{$\pm$ 3.3} & 22.4 \textsubscript{$\pm$ 3.6} & 51.2 \textsubscript{$\pm$ 4.2} & 45.6 \textsubscript{$\pm$ 4.4} & 56.3 \textsubscript{$\pm$ 4.3} & 60.9 \textsubscript{$\pm$ 4.3} \\
& MeLLo & 5.5 \textsubscript{$\pm$ 1.9} & 16.5 \textsubscript{$\pm$ 3.3} & 5.8 \textsubscript{$\pm$ 2.0} & 12.5 \textsubscript{$\pm$ 2.9} & 18.6 \textsubscript{$\pm$ 3.4} & 49.4 \textsubscript{$\pm$ 4.4} & 27.8 \textsubscript{$\pm$ 3.8} & 40.2 \textsubscript{$\pm$ 4.4} & 31.8 \textsubscript{$\pm$ 4.0} \\
\midrule
\multirow{5}{*}{Qwen-3 \textsubscript{(4B)}} & Base Model & 7.7 \textsubscript{$\pm$ 2.3} & 26.2 \textsubscript{$\pm$ 3.8} & 8.1 \textsubscript{$\pm$ 2.3} & 24.7 \textsubscript{$\pm$ 3.7} & 35.8 \textsubscript{$\pm$ 4.2} & 59.9 \textsubscript{$\pm$ 4.3} & 63.0 \textsubscript{$\pm$ 4.2} & 74.1 \textsubscript{$\pm$ 3.9} & 73.5 \textsubscript{$\pm$ 3.9} \\
\cmidrule(l){2-11}
& Append & 80.5 \textsubscript{$\pm$ 3.5} & \textbf{94.1} \textsubscript{$\pm$ 2.1} & 82.6 \textsubscript{$\pm$ 3.2} & 28.2 \textsubscript{$\pm$ 3.8} & 34.7 \textsubscript{$\pm$ 4.1} & 69.7 \textsubscript{$\pm$ 3.9} & 62.3 \textsubscript{$\pm$ 4.1} & 74.2 \textsubscript{$\pm$ 3.8} & 72.9 \textsubscript{$\pm$ 4.1} \\
& Append-T & 46.9 \textsubscript{$\pm$ 4.3} & 63.4 \textsubscript{$\pm$ 4.2} & 47.6 \textsubscript{$\pm$ 4.4} & 22.8 \textsubscript{$\pm$ 3.6} & 30.2 \textsubscript{$\pm$ 3.8} & 42.0 \textsubscript{$\pm$ 4.2} & 36.5 \textsubscript{$\pm$ 4.1} & 87.5 \textsubscript{$\pm$ 2.9} & 43.8 \textsubscript{$\pm$ 4.2} \\
& FT-CK & 9.0 \textsubscript{$\pm$ 2.4} & 30.3 \textsubscript{$\pm$ 3.9} & 9.0 \textsubscript{$\pm$ 2.4} & 25.6 \textsubscript{$\pm$ 3.6} & 36.9 \textsubscript{$\pm$ 4.3} & 60.6 \textsubscript{$\pm$ 4.4} & 62.2 \textsubscript{$\pm$ 4.4} & 74.7 \textsubscript{$\pm$ 3.9} & 72.7 \textsubscript{$\pm$ 3.9} \\
& MeLLo & 7.9 \textsubscript{$\pm$ 2.3} & 29.4 \textsubscript{$\pm$ 4.0} & 7.8 \textsubscript{$\pm$ 2.2} & 21.4 \textsubscript{$\pm$ 3.8} & 32.2 \textsubscript{$\pm$ 4.0} & 55.2 \textsubscript{$\pm$ 4.4} & 57.0 \textsubscript{$\pm$ 4.4} & 74.4 \textsubscript{$\pm$ 4.0} & 59.2 \textsubscript{$\pm$ 4.4} \\
\midrule
\multirow{5}{*}{Qwen-3 \textsubscript{(8B)}} & Base Model & 10.3 \textsubscript{$\pm$ 2.5} & 26.8 \textsubscript{$\pm$ 3.8} & 10.5 \textsubscript{$\pm$ 2.5} & 27.2 \textsubscript{$\pm$ 3.8} & 39.7 \textsubscript{$\pm$ 4.3} & 63.1 \textsubscript{$\pm$ 4.3} & 68.8 \textsubscript{$\pm$ 4.0} & 85.0 \textsubscript{$\pm$ 3.0} & 70.3 \textsubscript{$\pm$ 3.9} \\
\cmidrule(l){2-11}
& Append & 77.7 \textsubscript{$\pm$ 3.7} & 91.6 \textsubscript{$\pm$ 2.4} & 78.9 \textsubscript{$\pm$ 3.5} & 27.8 \textsubscript{$\pm$ 3.8} & 40.6 \textsubscript{$\pm$ 4.4} & 63.4 \textsubscript{$\pm$ 4.2} & 70.7 \textsubscript{$\pm$ 3.9} & 87.0 \textsubscript{$\pm$ 3.0} & 71.8 \textsubscript{$\pm$ 4.0} \\
& Append-T & 73.6 \textsubscript{$\pm$ 3.8} & 93.7 \textsubscript{$\pm$ 2.1} & 74.3 \textsubscript{$\pm$ 3.7} & 26.5 \textsubscript{$\pm$ 3.9} & 32.1 \textsubscript{$\pm$ 4.1} & 43.8 \textsubscript{$\pm$ 4.4} & 70.7 \textsubscript{$\pm$ 3.9} & 82.2 \textsubscript{$\pm$ 3.2} & 81.7 \textsubscript{$\pm$ 3.5} \\
& FT-CK & 10.4 \textsubscript{$\pm$ 2.6} & 28.4 \textsubscript{$\pm$ 4.0} & 10.3 \textsubscript{$\pm$ 2.5} & 28.5 \textsubscript{$\pm$ 3.9} & 38.6 \textsubscript{$\pm$ 4.4} & 65.5 \textsubscript{$\pm$ 4.1} & 68.7 \textsubscript{$\pm$ 3.9} & 84.6 \textsubscript{$\pm$ 3.0} & 70.4 \textsubscript{$\pm$ 4.2} \\
& MeLLo & 10.6 \textsubscript{$\pm$ 2.6} & 27.8 \textsubscript{$\pm$ 4.0} & 11.5 \textsubscript{$\pm$ 2.9} & 24.8 \textsubscript{$\pm$ 3.8} & 34.2 \textsubscript{$\pm$ 4.2} & 59.7 \textsubscript{$\pm$ 4.3} & 49.5 \textsubscript{$\pm$ 4.3} & 65.3 \textsubscript{$\pm$ 4.3} & 51.6 \textsubscript{$\pm$ 4.2} \\
\midrule
\multirow{2}{*}{GPT-4.1-mini \&} & Base Model & 22.9 \textsubscript{$\pm$ 3.7} & 49.8 \textsubscript{$\pm$ 4.6} & 22.6 \textsubscript{$\pm$ 3.6} & 35.9 \textsubscript{$\pm$ 4.1} & 48.0 \textsubscript{$\pm$ 4.4} & 71.9 \textsubscript{$\pm$ 4.1} & 79.3 \textsubscript{$\pm$ 3.5} & 94.0 \textsubscript{$\pm$ 2.0} & 80.9 \textsubscript{$\pm$ 3.5} \\
\cmidrule(l){2-11}
& Append & 70.8 \textsubscript{$\pm$ 4.0} & 93.7 \textsubscript{$\pm$ 2.1} & 71.5 \textsubscript{$\pm$ 3.9} & 36.8 \textsubscript{$\pm$ 4.2} & \textbf{49.2} \textsubscript{$\pm$ 4.2} & 72.7 \textsubscript{$\pm$ 3.9} & 86.4 \textsubscript{$\pm$ 3.0} & \textbf{98.7} \textsubscript{$\pm$ 0.9} & 87.2 \textsubscript{$\pm$ 2.8} \\
o4-mini & Append-T & 79.6 \textsubscript{$\pm$ 3.4} & 93.8 \textsubscript{$\pm$ 2.2} & 79.9 \textsubscript{$\pm$ 3.5} & \textbf{40.6} \textsubscript{$\pm$ 4.2} & 48.1 \textsubscript{$\pm$ 4.1} & \textbf{77.0} \textsubscript{$\pm$ 3.6} & \textbf{87.2} \textsubscript{$\pm$ 2.8} & \textbf{98.7} \textsubscript{$\pm$ 0.9} & \textbf{87.3} \textsubscript{$\pm$ 2.9} \\
\bottomrule
\end{tabular}
    }
    \caption{\textbf{Main results of knowledge injection methods} on the \GROW, \CODE, and \MATH scenarios. We report Holistic Pass (HP), Answer Pass (AP), and Full Knowledge Entailment (FKE) results. We compare the Base Model (closed-book) against several open-book knowledge injection methods (Append, FT-CK, MeLLo) across Llama-3.2, Qwen-3, and GPT series. We set KAS to 1 in the open-book setting so each question receives only its missing facts. We report 95\% confidence intervals (CIs) in the $\pm$ sign and \textbf{bold} the best scores per column.}
    \label{tab:main:main}
\end{table*}

\section{Knowledge Injection}
\label{sec:main:injection}

In the knowledge injection stage, we evaluate how various models and methods perform when presented with knowledge conflicts. These knowledge conflicts are constructed from identified knowledge gaps in the knowledge probing stage. This section details the experimental setup (\secref{sec:main:injection:setup}), presents the main results (\secref{sec:main:injection:results}), and provides an in-depth failure analysis (\secref{sec:main:injection:analysis}).

% \takeaways{
% (\textit{i}) LLMs struggle to integrate provided atomic facts into a correct reasoning process, leading to limited performance gains or even worse performance.
% (\textit{ii}) Performance degrades further as more knowledge is injected.
% (\textit{iii}) The knowledge injection failure stems from both the inability to faithfully leverage the provided facts and flawed reasoning even when knowledge is correctly integrated.
% }
% This section details our empirical study. We first outline the experimental setup (\secref{sec:main:experiments:setup}), then analyze results from the knowledge probing (\secref{sec:main:experiments:probing_results}) and injection (\secref{sec:main:experiments:reasoning_results}) stages, and conclude by diagnosing the primary reasoning failures (\secref{sec:main:experiments:injection_analysis}).
% \todo{I will add more experiment results this week, and feel free to skip the detailed analysis.}

\subsection{Experimental Setup}
\label{sec:main:injection:setup}

\SmallHeading{Knowledge Injection}
Building on the same set of backbone LLMs (\Section~\ref{sec:main:probing:setup}), we evaluate how they perform multi-step reasoning when injected with conflicting knowledge. We compare a closed-book \ul{Base Model} baseline against a diverse set of open-book injection methods representing paradigms like in-context learning, test-time inference, fine-tuning, and retrieval-augmentation. These methods include \ul{Append},\footnote{Early experiments showed that appending outperforms prepending facts in-context, potentially because the question provides context for the model to attend to relevant knowledge.} which provides new facts in-context; Append-Thinking (\ul{Append-T}), a variant that enables longer thinking for Qwen-3 and o4-mini; Fine-tuning on Conflicting Knowledge (\ul{FT-CK}); and \ul{MeLLo}~\citep{zhong-etal-2023-mquake}, which uses an external memory to store injected facts and a question-decomposition approach for retrieval and reasoning. %We do not use other popular knowledge editing methods like ROME~\cite{meng2023locatingeditingfactualassociations}, MEMIT~\cite{pmlr-v162-mitchell22a}, or MEND~\cite{mitchell2022fastmodeleditingscale}, as they require strict data formats or are optimized for short-form responses, making them unsuitable for our task. 
For all open-book methods, we vary the number of KAS (\Section~\ref{sec:main:task:definition}) to 1, 10, 100, and 500. Full implementation specifics are in Appendix~\ref{sec:appendix:setup:injection}.

\subsection{Knowledge Injection Effectiveness}
\label{sec:main:injection:results}

\begin{table*}[t!]
    \centering
    \resizebox{0.99\textwidth}{!}{
    \begin{tabular}{p{2.6cm}p{1.6cm}p{1.5cm}p{1.5cm}p{1.5cm}p{1.5cm}p{1.5cm}p{1.5cm}p{1.5cm}p{1.5cm}p{1.5cm}p{1.5cm}p{1.5cm}p{1.5cm}}
\toprule

\multirow{2}{*}{\textbf{Backbone Model}} & \multirow{2}{*}{\textbf{Method}} & \multicolumn{4}{c}{\textbf{WIKI}} & \multicolumn{4}{c}{\textbf{CODE}} & \multicolumn{4}{c}{\textbf{MATH}}\\
\cmidrule(r){3-6} \cmidrule(l){7-10} \cmidrule(l){11-14}

& & $\text{KAS}$=1 & $\text{KAS}$=10 & $\text{KAS}$=100 & $\text{KAS}$=500 & $\text{KAS}$=1 & $\text{KAS}$=10 & $\text{KAS}$=100 & $\text{KAS}$=500 & $\text{KAS}$=1 & $\text{KAS}$=10 & $\text{KAS}$=100 & $\text{KAS}$=500 \\
\midrule
\multirow{3}{*}{Llama-3.2 \textsubscript{(1B)}} & Append & 1.1 \textsubscript{$\pm$ 0.9} & 1.1 \textsubscript{$\pm$ 0.9} & 0.7 \textsubscript{$\pm$ 0.7} & 0.9 \textsubscript{$\pm$ 0.7} & 4.1 \textsubscript{$\pm$ 1.7} & 4.4 \textsubscript{$\pm$ 1.8} & 5.4 \textsubscript{$\pm$ 2.0} & 4.3 \textsubscript{$\pm$ 1.9} & 22.5 \textsubscript{$\pm$ 3.7} & 18.6 \textsubscript{$\pm$ 3.4} & 18.1 \textsubscript{$\pm$ 3.3} & 20.6 \textsubscript{$\pm$ 3.4} \\
& FT-CK & 0.9 \textsubscript{$\pm$ 0.7} & 1.1 \textsubscript{$\pm$ 0.9} & 1.3 \textsubscript{$\pm$ 0.9} & 1.3 \textsubscript{$\pm$ 0.9} & 5.1 \textsubscript{$\pm$ 1.9} & 4.1 \textsubscript{$\pm$ 1.7} & 4.9 \textsubscript{$\pm$ 1.9} & 5.7 \textsubscript{$\pm$ 1.9} & 21.8 \textsubscript{$\pm$ 3.6} & 19.3 \textsubscript{$\pm$ 3.5} & 18.4 \textsubscript{$\pm$ 3.4} & 19.8 \textsubscript{$\pm$ 3.4} \\
& MeLLo & 2.4 \textsubscript{$\pm$ 1.2} & 2.1 \textsubscript{$\pm$ 1.3} & 1.9 \textsubscript{$\pm$ 1.1} & 2.3 \textsubscript{$\pm$ 1.3} & 3.9 \textsubscript{$\pm$ 1.7} & 5.8 \textsubscript{$\pm$ 2.0} & 5.3 \textsubscript{$\pm$ 2.1} & 3.4 \textsubscript{$\pm$ 1.6} & 18.0 \textsubscript{$\pm$ 3.2} & 17.5 \textsubscript{$\pm$ 3.1} & 20.9 \textsubscript{$\pm$ 3.3} & 18.6 \textsubscript{$\pm$ 3.4} \\
\midrule
\multirow{3}{*}{Llama-3.2 \textsubscript{(3B)}} & Append & 7.3 \textsubscript{$\pm$ 2.1} & 6.9 \textsubscript{$\pm$ 2.3} & 5.6 \textsubscript{$\pm$ 2.0} & 6.7 \textsubscript{$\pm$ 2.3} & 16.8 \textsubscript{$\pm$ 3.2} & 16.2 \textsubscript{$\pm$ 3.4} & 14.1 \textsubscript{$\pm$ 3.1} & 14.3 \textsubscript{$\pm$ 3.1} & 40.4 \textsubscript{$\pm$ 4.2} & 40.1 \textsubscript{$\pm$ 4.5} & 41.1 \textsubscript{$\pm$ 4.3} & 40.4 \textsubscript{$\pm$ 4.2} \\
& FT-CK & 7.4 \textsubscript{$\pm$ 2.2} & 7.7 \textsubscript{$\pm$ 2.3} & 6.1 \textsubscript{$\pm$ 2.1} & 7.3 \textsubscript{$\pm$ 2.3} & 12.8 \textsubscript{$\pm$ 2.8} & 16.4 \textsubscript{$\pm$ 3.4} & 16.1 \textsubscript{$\pm$ 3.1} & 14.7 \textsubscript{$\pm$ 3.1} & 41.6 \textsubscript{$\pm$ 4.2} & 40.9 \textsubscript{$\pm$ 4.3} & 41.0 \textsubscript{$\pm$ 4.2} & 41.6 \textsubscript{$\pm$ 4.4} \\
& MeLLo & 7.3 \textsubscript{$\pm$ 2.3} & 6.9 \textsubscript{$\pm$ 2.3} & 6.7 \textsubscript{$\pm$ 2.1} & 7.4 \textsubscript{$\pm$ 2.2} & 12.2 \textsubscript{$\pm$ 2.8} & 11.4 \textsubscript{$\pm$ 2.8} & 11.4 \textsubscript{$\pm$ 2.8} & 12.9 \textsubscript{$\pm$ 2.9} & 37.9 \textsubscript{$\pm$ 4.3} & 41.2 \textsubscript{$\pm$ 4.4} & 37.5 \textsubscript{$\pm$ 4.3} & 35.3 \textsubscript{$\pm$ 4.1} \\
\midrule
\multirow{3}{*}{Llama-3.2 \textsubscript{(11B)}} & Append & 13.1 \textsubscript{$\pm$ 2.9} & 12.2 \textsubscript{$\pm$ 2.8} & 12.3 \textsubscript{$\pm$ 2.7} & 14.2 \textsubscript{$\pm$ 3.0} & 18.1 \textsubscript{$\pm$ 3.3} & 18.2 \textsubscript{$\pm$ 3.6} & 19.5 \textsubscript{$\pm$ 3.7} & 19.2 \textsubscript{$\pm$ 3.4} & 33.4 \textsubscript{$\pm$ 4.0} & 32.2 \textsubscript{$\pm$ 4.2} & 33.0 \textsubscript{$\pm$ 4.2} & 30.6 \textsubscript{$\pm$ 4.2} \\
& FT-CK & 12.7 \textsubscript{$\pm$ 2.9} & 13.6 \textsubscript{$\pm$ 3.0} & 13.2 \textsubscript{$\pm$ 3.0} & 12.5 \textsubscript{$\pm$ 2.9} & 19.1 \textsubscript{$\pm$ 3.5} & 19.4 \textsubscript{$\pm$ 3.6} & 18.7 \textsubscript{$\pm$ 3.5} & 20.3 \textsubscript{$\pm$ 3.5} & 32.6 \textsubscript{$\pm$ 4.2} & 33.9 \textsubscript{$\pm$ 4.3} & 33.8 \textsubscript{$\pm$ 4.2} & 34.6 \textsubscript{$\pm$ 4.2} \\
& MeLLo & 11.0 \textsubscript{$\pm$ 2.6} & 10.5 \textsubscript{$\pm$ 2.5} & 10.3 \textsubscript{$\pm$ 2.5} & 11.4 \textsubscript{$\pm$ 2.6} & 9.4 \textsubscript{$\pm$ 2.6} & 9.2 \textsubscript{$\pm$ 2.6} & 8.7 \textsubscript{$\pm$ 2.5} & 8.6 \textsubscript{$\pm$ 2.4} & 20.3 \textsubscript{$\pm$ 3.3} & 20.3 \textsubscript{$\pm$ 3.5} & 18.7 \textsubscript{$\pm$ 3.3} & 18.6 \textsubscript{$\pm$ 3.2} \\
\midrule
\multirow{4}{*}{Qwen-3 \textsubscript{(1.7B)}} & Append & \textbf{83.6} \textsubscript{$\pm$ 3.2} & 59.3 \textsubscript{$\pm$ 4.1} & 46.7 \textsubscript{$\pm$ 4.3} & 22.2 \textsubscript{$\pm$ 3.6} & 18.3 \textsubscript{$\pm$ 3.3} & 16.7 \textsubscript{$\pm$ 3.3} & 15.0 \textsubscript{$\pm$ 3.2} & 17.8 \textsubscript{$\pm$ 3.4} & 49.2 \textsubscript{$\pm$ 4.0} & 50.8 \textsubscript{$\pm$ 4.2} & 47.6 \textsubscript{$\pm$ 4.2} & 48.6 \textsubscript{$\pm$ 4.2} \\
& Append-T & 6.4 \textsubscript{$\pm$ 2.2} & 3.6 \textsubscript{$\pm$ 1.6} & 1.3 \textsubscript{$\pm$ 0.9} & 1.4 \textsubscript{$\pm$ 1.0} & 15.5 \textsubscript{$\pm$ 3.1} & 15.3 \textsubscript{$\pm$ 3.1} & 14.8 \textsubscript{$\pm$ 3.2} & 15.6 \textsubscript{$\pm$ 3.2} & 7.1 \textsubscript{$\pm$ 2.3} & 7.3 \textsubscript{$\pm$ 2.3} & 7.1 \textsubscript{$\pm$ 2.1} & 7.1 \textsubscript{$\pm$ 2.1} \\
& FT-CK & 4.8 \textsubscript{$\pm$ 1.8} & 0.7 \textsubscript{$\pm$ 0.7} & 9.3 \textsubscript{$\pm$ 2.5} & 1.0 \textsubscript{$\pm$ 0.8} & 16.7 \textsubscript{$\pm$ 3.3} & 16.8 \textsubscript{$\pm$ 3.4} & 17.3 \textsubscript{$\pm$ 3.3} & 1.7 \textsubscript{$\pm$ 1.1} & 45.6 \textsubscript{$\pm$ 4.4} & 47.9 \textsubscript{$\pm$ 4.5} & 43.3 \textsubscript{$\pm$ 4.1} & 48.1 \textsubscript{$\pm$ 4.1} \\
& MeLLo & 5.5 \textsubscript{$\pm$ 1.9} & 6.3 \textsubscript{$\pm$ 2.1} & 9.4 \textsubscript{$\pm$ 2.4} & 9.3 \textsubscript{$\pm$ 2.5} & 12.5 \textsubscript{$\pm$ 2.9} & 14.0 \textsubscript{$\pm$ 3.0} & 13.3 \textsubscript{$\pm$ 3.1} & 12.2 \textsubscript{$\pm$ 2.8} & 27.8 \textsubscript{$\pm$ 3.8} & 27.8 \textsubscript{$\pm$ 4.0} & 26.7 \textsubscript{$\pm$ 3.7} & 25.3 \textsubscript{$\pm$ 3.5} \\
\midrule
\multirow{4}{*}{Qwen-3 \textsubscript{(4B)}} & Append & 80.5 \textsubscript{$\pm$ 3.5} & 75.5 \textsubscript{$\pm$ 3.7} & 67.0 \textsubscript{$\pm$ 4.0} & 58.8 \textsubscript{$\pm$ 4.4} & 28.2 \textsubscript{$\pm$ 3.8} & 25.2 \textsubscript{$\pm$ 3.8} & 25.3 \textsubscript{$\pm$ 3.7} & 26.0 \textsubscript{$\pm$ 3.8} & 62.3 \textsubscript{$\pm$ 4.1} & 78.0 \textsubscript{$\pm$ 3.6} & 64.7 \textsubscript{$\pm$ 4.1} & 62.8 \textsubscript{$\pm$ 4.2} \\
& Append-T & 46.9 \textsubscript{$\pm$ 4.3} & 11.4 \textsubscript{$\pm$ 2.8} & 10.6 \textsubscript{$\pm$ 2.6} & 3.5 \textsubscript{$\pm$ 1.7} & 22.8 \textsubscript{$\pm$ 3.6} & 23.0 \textsubscript{$\pm$ 3.8} & 23.1 \textsubscript{$\pm$ 3.7} & 22.8 \textsubscript{$\pm$ 3.6} & 36.5 \textsubscript{$\pm$ 4.1} & 32.4 \textsubscript{$\pm$ 4.2} & 32.6 \textsubscript{$\pm$ 3.8} & 33.4 \textsubscript{$\pm$ 4.2} \\
& FT-CK & 9.0 \textsubscript{$\pm$ 2.4} & 0.7 \textsubscript{$\pm$ 0.7} & 0.0 \textsubscript{$\pm$ 0.0} & 0.0 \textsubscript{$\pm$ 0.0} & 25.6 \textsubscript{$\pm$ 3.6} & 26.0 \textsubscript{$\pm$ 3.6} & 8.4 \textsubscript{$\pm$ 2.4} & 0.0 \textsubscript{$\pm$ 0.0} & 62.2 \textsubscript{$\pm$ 4.4} & 65.8 \textsubscript{$\pm$ 4.2} & 62.7 \textsubscript{$\pm$ 4.3} & 66.2 \textsubscript{$\pm$ 4.0} \\
& MeLLo & 7.9 \textsubscript{$\pm$ 2.3} & 9.6 \textsubscript{$\pm$ 2.6} & 17.4 \textsubscript{$\pm$ 3.2} & 26.6 \textsubscript{$\pm$ 3.8} & 21.4 \textsubscript{$\pm$ 3.8} & 21.1 \textsubscript{$\pm$ 3.5} & 22.5 \textsubscript{$\pm$ 3.7} & 18.7 \textsubscript{$\pm$ 3.5} & 57.0 \textsubscript{$\pm$ 4.4} & 57.0 \textsubscript{$\pm$ 4.4} & 55.8 \textsubscript{$\pm$ 4.4} & 55.0 \textsubscript{$\pm$ 4.4} \\
\midrule
\multirow{4}{*}{Qwen-3 \textsubscript{(8B)}} & Append & 77.7 \textsubscript{$\pm$ 3.7} & 62.3 \textsubscript{$\pm$ 4.1} & 57.0 \textsubscript{$\pm$ 4.4} & 51.5 \textsubscript{$\pm$ 4.1} & 27.8 \textsubscript{$\pm$ 3.8} & 29.4 \textsubscript{$\pm$ 4.0} & 28.6 \textsubscript{$\pm$ 4.0} & 26.0 \textsubscript{$\pm$ 3.6} & 70.7 \textsubscript{$\pm$ 3.9} & 67.7 \textsubscript{$\pm$ 4.1} & 69.7 \textsubscript{$\pm$ 3.9} & 70.2 \textsubscript{$\pm$ 3.8} \\
& Append-T & 73.6 \textsubscript{$\pm$ 3.8} & 45.6 \textsubscript{$\pm$ 4.2} & 24.9 \textsubscript{$\pm$ 3.7} & 11.2 \textsubscript{$\pm$ 2.6} & 26.5 \textsubscript{$\pm$ 3.9} & 25.9 \textsubscript{$\pm$ 3.7} & 26.1 \textsubscript{$\pm$ 3.9} & 27.6 \textsubscript{$\pm$ 3.8} & 70.7 \textsubscript{$\pm$ 3.9} & 69.0 \textsubscript{$\pm$ 4.0} & 71.4 \textsubscript{$\pm$ 4.0} & 69.4 \textsubscript{$\pm$ 4.0} \\
& FT-CK & 10.4 \textsubscript{$\pm$ 2.6} & 10.8 \textsubscript{$\pm$ 2.6} & 19.0 \textsubscript{$\pm$ 3.4} & 4.1 \textsubscript{$\pm$ 1.7} & 28.5 \textsubscript{$\pm$ 3.9} & 29.1 \textsubscript{$\pm$ 3.7} & 28.0 \textsubscript{$\pm$ 3.8} & 28.3 \textsubscript{$\pm$ 3.9} & 68.7 \textsubscript{$\pm$ 3.9} & 69.9 \textsubscript{$\pm$ 3.9} & 70.2 \textsubscript{$\pm$ 3.8} & 72.6 \textsubscript{$\pm$ 3.8} \\
& MeLLo & 10.6 \textsubscript{$\pm$ 2.6} & 11.5 \textsubscript{$\pm$ 2.7} & 24.7 \textsubscript{$\pm$ 3.7} & 24.7 \textsubscript{$\pm$ 3.7} & 24.8 \textsubscript{$\pm$ 3.8} & 23.3 \textsubscript{$\pm$ 3.7} & 22.8 \textsubscript{$\pm$ 3.6} & 23.8 \textsubscript{$\pm$ 3.6} & 49.5 \textsubscript{$\pm$ 4.3} & 50.5 \textsubscript{$\pm$ 4.5} & 51.1 \textsubscript{$\pm$ 4.3} & 48.8 \textsubscript{$\pm$ 4.4} \\
\midrule
GPT-4.1-mini \& & Append & 70.8 \textsubscript{$\pm$ 4.0} & 69.8 \textsubscript{$\pm$ 4.0} & 71.4 \textsubscript{$\pm$ 4.0} & 70.8 \textsubscript{$\pm$ 4.0} & 36.8 \textsubscript{$\pm$ 4.2} & 33.7 \textsubscript{$\pm$ 4.1} & 37.6 \textsubscript{$\pm$ 4.2} & 34.8 \textsubscript{$\pm$ 4.2} & 86.4 \textsubscript{$\pm$ 3.0} & 82.3 \textsubscript{$\pm$ 3.3} & 82.6 \textsubscript{$\pm$ 3.2} & \textbf{81.7} \textsubscript{$\pm$ 3.3} \\
o4-mini & Append-T & 79.6 \textsubscript{$\pm$ 3.4} & \textbf{80.3} \textsubscript{$\pm$ 3.5} & \textbf{77.5} \textsubscript{$\pm$ 3.5} & \textbf{78.2} \textsubscript{$\pm$ 3.6} & \textbf{40.6} \textsubscript{$\pm$ 4.2} & \textbf{38.3} \textsubscript{$\pm$ 4.3} & \textbf{38.6} \textsubscript{$\pm$ 4.4} & \textbf{39.4} \textsubscript{$\pm$ 4.2} & \textbf{87.2} \textsubscript{$\pm$ 2.8} & \textbf{85.5} \textsubscript{$\pm$ 3.1} & \textbf{83.3} \textsubscript{$\pm$ 3.1} & 80.7 \textsubscript{$\pm$ 3.5} \\
\bottomrule
\end{tabular}
    }
    \caption{\textbf{Impact of Knowledge Aggregation Scope (KAS) on Holistic Pass (HP).} We inject a batch of conflicted knowledge associated with $\text{KAS}=1,10,100,500$, comparing open-book knowledge injection methods (Append, Append-T, FT-CK, MeLLo). We report 95\% CIs in the $\pm$ sign and \textbf{bold} the best scores per column.}
    \label{tab:main:kas}
\end{table*}

\begin{figure*}[t!]
    \centering
    \begin{subfigure}[t]{0.3\textwidth}
        \centering
        \includegraphics[width=\textwidth]{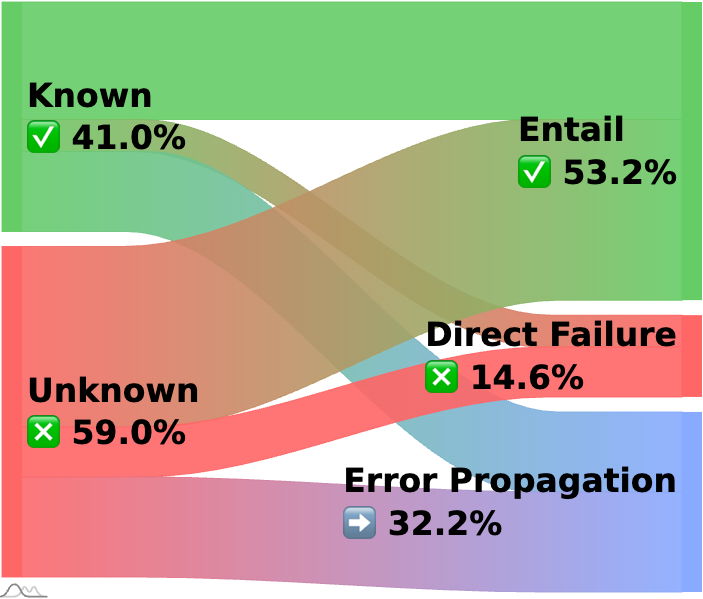}
        \caption{\GROW}
        \label{fig:sankey_grow}
    \end{subfigure}
    \hfill
    \begin{subfigure}[t]{0.3\textwidth}
        \centering
        \includegraphics[width=\textwidth]{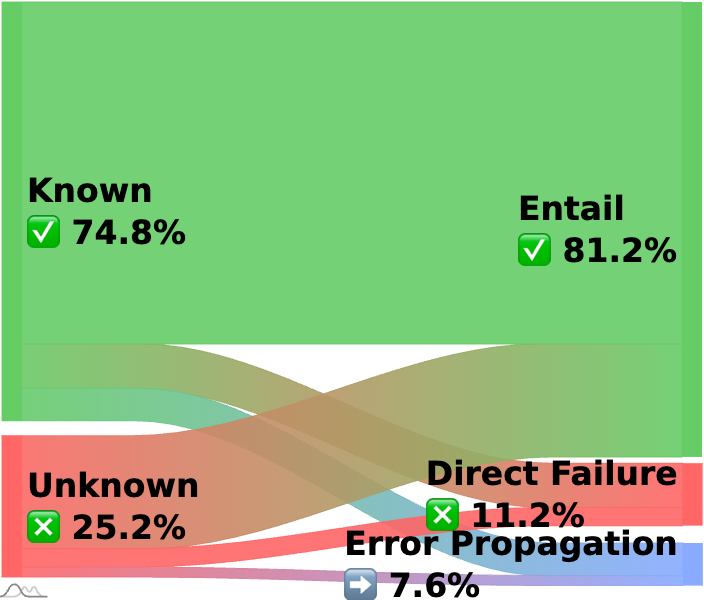}
        \caption{\CODE}
        \label{fig:sankey_code}
    \end{subfigure}
    \hfill
    \begin{subfigure}[t]{0.3\textwidth}
        \centering
        \includegraphics[width=\textwidth]{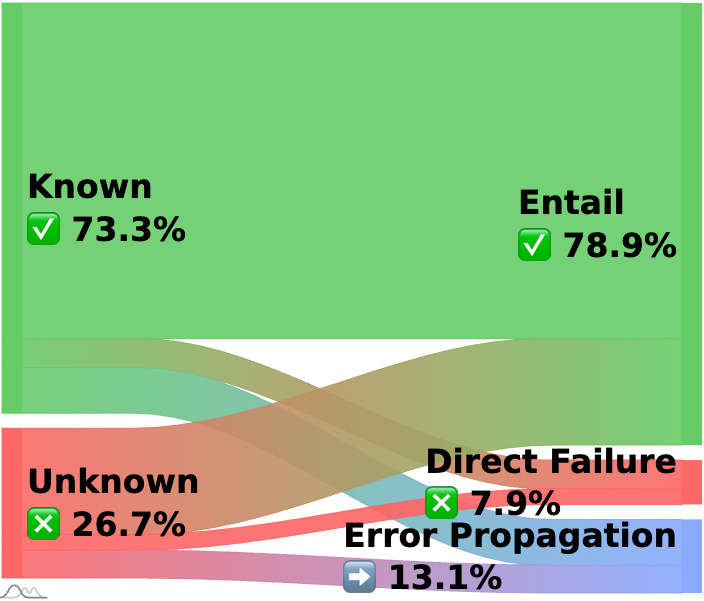}
        \caption{\MATH}
        \label{fig:sankey_math}
    \end{subfigure}
    \caption{\textbf{Failure analysis at the atomic fact level for the Append method}, averaged across all models. The flow shows the percentage of all facts that were initially known or unknown (left) and their outcome in the reasoning (right): successfully integrated (Entail), being the first point of failure (Direct Failure), or failing due to a previous error (Error Propagation).}
    \label{fig:sankey_failure_analysis}
\end{figure*}

\SmallHeading{Most methods show limited reasoning improvements or degraded performance}
We evaluate knowledge injection performance at KAS=1. As shown in \Table~\ref{tab:main:main}, the poor performance of the Base Model in \GROW and \CODE shows that reasoning struggles without the necessary atomic facts. In the open-book setting, the simple Append method yields the most substantial gains for Qwen-3 and GPT series on the \WIKI scenario. For example, Qwen-3 1.7B's HP score on \WIKI jumps from 4.5\% to 83.6\%. However, for other scenarios (\CODE and \MATH), models (Llama-3.2 series, GPT series), and methods (FT-CK, MeLLo), the improvement is limited or even worse than the Base Model without injection. Notably, longer thinking has a limited improvement on o4-mini and even decreases Qwen-3 models' performance.

\SmallHeading{Some models and methods show performance degradation under increased knowledge scopes}
We increase the KAS (\Section~\ref{sec:main:task:metrics}), which aggregates unknown facts from multiple questions. As shown in \Table~\ref{tab:main:kas}, performance with the Append and Append-T methods degrade significantly for some models as KAS increases. For instance, on \WIKI, the HP score for Qwen-3 1.7B Append method drops from 83.6\% (KAS=1) to 22.2\% (KAS=500), and for Qwen-3 8B Append-T method drops from 73.6\% (KAS=1) to 11.2\% (KAS=500). This suggests that as the provided information becomes saturated with more irrelevant facts, the model's attention mechanism struggles to identify and use the pertinent knowledge. In contrast, some models or methods exhibit greater resilience: GPT-4.1-mini remains highly stable with slight performance drops for both methods, and MeLLo on Llama-3.2 and Qwen-3 can even improve with a larger KAS (e.g., on \WIKI scenario). This indicates that longer reasoning or explicit retrieval is better at handling more distractor information.

\subsection{Analysis of Knowledge Injection Failure}
\label{sec:main:injection:analysis}

To diagnose why performance remains low even with injected knowledge, we analyze the failure modes. We identify two primary causes: (i) the model's inability to faithfully use the provided atomic facts, and (ii) flawed reasoning, even when all facts are correctly used.

\SmallHeading{Models struggle to integrate atomic facts faithfully}
A primary failure mode is the model's lack of faithfulness to the provided facts, evidenced by a large gap between achieving a correct answer (AP) and using faithful reasoning (FKE) in Llama-3.2 models on \WIKI scenario (\Table~\ref{tab:main:main}). This disparity suggests that models often arrive at the correct answer by coincidence; for instance, a model might use an outdated entity (e.g., a former president of the US) yet reach the correct final answer because the incorrect entity happens to share the queried attribute (e.g., a spouse's birthplace) with the correct one. This lack of grounding is a critical failure, proving the model cannot be reliably steered by new information. 

\footnotetext{We treat the Append method for the o4-mini model as the Append-T method for the GPT-4.1-mini model, because evidence shows that GPT-4.1-mini and o4-mini share the same knowledge (according to \url{https://platform.openai.com/docs/models/gpt-4.1-mini}, \url{https://platform.openai.com/docs/models/o4-mini}, and results in \Figure~\ref{fig:kconf_by_model}). We hypothesize that o4-mini and GPT-4.1-mini use similar architectures, but o4-mini is optimized for longer thinking.}

We further dissect these faithfulness failures into \textit{direct failures} (the first required atomic fact that is not entailed in reasoning) and \textit{error propagation} (subsequent facts not entailed). As shown in \Figure~\ref{fig:sankey_failure_analysis}, direct failures are a significant issue, accounting for 7.9–14.6\% of outcomes across all scenarios. These failures occur on both known and unknown facts, revealing an inconsistency in how models apply their knowledge during multi-step reasoning. More critically, these initial errors trigger a massive cascade of error propagation failures (7.6\%–32.2\%), demonstrating a compounding effect where a single mistake derails the entire reasoning process.
% Direct failure is the entailment failure of the first atomic fact $k_i$ in the required facts ($K_q$), conditioned on all preceding facts ($k_{<i}$) being successfully entailed. Any subsequent facts ($k_{>i}$) not entailed in reasoning are then considered an error propagation failure.
% As illustrated in \Figure~\ref{fig:sankey_failure_analysis}, direct failures are a significant issue, accounting for 17.10\% (\GROW), 18.80\% (\CODE), and 10.00\% (\MATH) of outcomes, and such failures occur on both Known and Unknown facts, revealing an inconsistency between a model's standalone knowledge and its ability to apply that knowledge during reasoning.
% Besides, direct failures trigger more error propagation failures: 23.00\% (\GROW), 35.50\% (\CODE), and 19.40\% (\MATH). This demonstrates that a single, initial error has a compounding effect, derailing the entire reasoning process and preventing the model from using subsequent atomic facts even they are known.

\SmallHeading{Successful fact integration does not guarantee accurate reasoning}
Even when models successfully incorporate all atomic facts, they can still fail due to flawed reasoning. This is evident from the consistent gap between the FKE and HP scores in \Table~\ref{tab:main:main}. This is most evident in the \CODE scenario. For instance, o4-mini with the Append-T method achieves a 76.7\% FKE, but this collapses to just a 40.4\% HP. This means that in nearly half of the cases where the model correctly utilized all API information, it still fails to produce a functionally correct program. Performance drops are also observed in other scenarios. For \MATH, Qwen-3 4B's FKE of 73.0\% results in an HP of only 62.1\%, and on \GROW, its 82.5\% FKE leads to an 80.6\% HP. This demonstrates that incorporating facts into correct reasoning and deriving correct final answers remains challenging.

% Conclusion
\section{Conclusion}

% To address the limitations of knowledge propagation evaluation in multi-step reasoning, we introduce \taskalias, a new dedicated benchmark featuring realistic, multi-fact conflicts across diverse reasoning scenarios (\GROW, \CODE, and \MATH). We also propose novel evaluation metrics that assess both the faithfulness and correctness of the reasoning. Our extensive experiments reveal a critical failure where providing correct conflicting facts shows limited improvements and even degrades performance, rooted in both unfaithful knowledge integration and flawed reasoning. The \taskalias benchmark provides a tool for the community to measure and drive progress on this crucial capability for building more reliable and adaptable models.
We introduce \taskalias, a benchmark that instantiates a two-stage framework (probing and injection) with realistic, multi-fact conflicts across diverse scenarios and novel metrics for evaluating reasoning faithfulness. Our experiments reveal a critical failure: providing correct facts often yields limited gains and can even degrade performance, due to both poor knowledge integration and flawed reasoning. \taskalias provides a rigorous new tool for the community to measure and guide progress on this crucial challenge.

% \clearpage

% Limitations and Ethical Considerations
\section*{Limitations}

First, our study is restricted to knowledge propagation within English, text-based reasoning scenarios. Consequently, we do not evaluate performance in more complex environments, such as multi-modal reasoning, multi-lingual contexts, or agent settings involving tool use, which may exhibit different propagation dynamics.
Additionally, we conduct experiments on a specific set of open-source and proprietary models (Llama-3.2, Qwen-3, GPT-4.1-mini, and o4-mini). While we include diverse architectures and capabilities (e.g., thinking vs. non-thinking variants), this selection is not exhaustive, and our findings may not strictly generalize to significantly larger models or alternative architectures not tested here.
Finally, our benchmark represents a static snapshot of knowledge. Unlike mathematical domains, facts in multi-hop QA and code libraries are subject to temporal drift and may become outdated. Although our dynamic data construction pipeline allows for regeneration using updated sources (e.g., new Wikidata dumps), the current evaluation does not cover such dynamics. This built-in extensibility paves the way for future studies on continual and life-long knowledge propagation.

% First, our study focuses on knowledge propagation within English, text-based reasoning scenarios. This scope provides a clear foundation for future work to explore more complex environments, such as multi-modal reasoning, multi-lingual contexts, and agentic settings involving tool use.
% Additionally, our experiments are conducted on a carefully selected set of representative open-source and proprietary models (Llama-3.2, Qwen-3, GPT-4.1-mini, and o4-mini). While not exhaustive, this selection was designed to cover a diverse range of architectures, sizes, and capabilities (e.g., thinking v.s. non-thinking variants) to provide initial and generalizable insights.
% Finally, we acknowledge that our benchmark is a static snapshot of knowledge at a specific point in time. While knowledge in domains like mathematics is relatively stable, facts in areas like multi-hop QA and code libraries can become outdated. However, a key feature of our framework is its dynamic data construction pipeline, which allows researchers to easily regenerate the benchmark with updated knowledge from sources like new Wikidata dumps or current code repositories. This built-in extensibility paves the way for future studies on continual and life-long knowledge propagation.

\section*{Ethical Considerations}

Our work probes a specific dimension of model reliability: the ability to faithfully reason with new, correct information against internal knowledge, which is foundational for trustworthy AI, particularly in real-world scenarios where knowledge is constantly evolving.
However, we acknowledge the ethical risks of blindly following user inputs.
Models excelling on our benchmark could be vulnerable to malicious prompt injection attacks~\cite{liu2024promptinjectionattackllmintegrated, xu-etal-2024-earth} or amplify unintentional misinformation~\cite{kumar2023math, xu2024ai, feng2025unravelingmisinformationpropagationllm}, for instance in domains like mathematical reasoning.
Our paper focuses on the first step: ensuring models can reliably process trustworthy information. We argue this is a prerequisite for the more advanced challenge of discerning when to trust an input. Future work should build upon our benchmark by incorporating non-factual or malicious data, thereby creating a more comprehensive testbed for developing models that can both faithfully utilize correct information and critically reject harmful input.

\section*{Acknowledgments}
We are grateful to EPFL NLP lab and specifically Gail Weiss, Silin Gao, and Beatriz Borges for their invaluable feedback and insightful suggestions.
We gratefully acknowledge the financial and IT support provided by EPFL, which allows us to use the server and computing resources.
We also gratefully acknowledge the support of the Swiss National Science Foundation (No. 215390), Innosuisse (PFFS-21-29), the EPFL Center for Imaging, Sony Group Corporation, and a Meta LLM Evaluation Research Grant.

\bibliography{custom}

\begin{thebibliography}{58}
\providecommand{\natexlab}[1]{#1}

\bibitem[{Achiam et~al.(2023)Achiam, Adler, Agarwal, Ahmad, Akkaya, Aleman, Almeida, Altenschmidt, Altman, Anadkat et~al.}]{achiam2023gpt}
Josh Achiam, Steven Adler, Sandhini Agarwal, Lama Ahmad, Ilge Akkaya, Florencia~Leoni Aleman, Diogo Almeida, Janko Altenschmidt, Sam Altman, Shyamal Anadkat, et~al. 2023.
\newblock \href {https://arxiv.org/abs/2303.08774} {Gpt-4 technical report}.
\newblock \emph{arXiv preprint arXiv:2303.08774}.

\bibitem[{Ahn et~al.(2024)Ahn, Verma, Lou, Liu, Zhang, and Yin}]{ahn-etal-2024-large}
Janice Ahn, Rishu Verma, Renze Lou, Di~Liu, Rui Zhang, and Wenpeng Yin. 2024.
\newblock \href {https://aclanthology.org/2024.eacl-srw.17/} {Large language models for mathematical reasoning: Progresses and challenges}.
\newblock In \emph{Proceedings of the 18th Conference of the European Chapter of the Association for Computational Linguistics: Student Research Workshop}, pages 225--237, St. Julian{'}s, Malta. Association for Computational Linguistics.

\bibitem[{{amCharts}(2025)}]{amcharts5}
{amCharts}. 2025.
\newblock \href {https://www.amcharts.com/docs/v5/} {amcharts 5: Javascript charts}.
\newblock Accessed: 2025-02-15.

\bibitem[{Chen et~al.(2021)Chen, Tworek, Jun, Yuan, de~Oliveira~Pinto, Kaplan, Edwards, Burda, Joseph, Brockman, Ray, Puri, Krueger, Petrov, Khlaaf, Sastry, Mishkin, Chan, Gray, Ryder, Pavlov, Power, Kaiser, Bavarian, Winter, Tillet, Such, Cummings, Plappert, Chantzis, Barnes, Herbert-Voss, Guss, Nichol, Paino, Tezak, Tang, Babuschkin, Balaji, Jain, Saunders, Hesse, Carr, Leike, Achiam, Misra, Morikawa, Radford, Knight, Brundage, Murati, Mayer, Welinder, McGrew, Amodei, McCandlish, Sutskever, and Zaremba}]{chen2021evaluatinglargelanguagemodels}
Mark Chen, Jerry Tworek, Heewoo Jun, Qiming Yuan, Henrique~Ponde de~Oliveira~Pinto, Jared Kaplan, Harri Edwards, Yuri Burda, Nicholas Joseph, Greg Brockman, Alex Ray, Raul Puri, Gretchen Krueger, Michael Petrov, Heidy Khlaaf, Girish Sastry, Pamela Mishkin, Brooke Chan, Scott Gray, Nick Ryder, Mikhail Pavlov, Alethea Power, Lukasz Kaiser, Mohammad Bavarian, Clemens Winter, Philippe Tillet, Felipe~Petroski Such, Dave Cummings, Matthias Plappert, Fotios Chantzis, Elizabeth Barnes, Ariel Herbert-Voss, William~Hebgen Guss, Alex Nichol, Alex Paino, Nikolas Tezak, Jie Tang, Igor Babuschkin, Suchir Balaji, Shantanu Jain, William Saunders, Christopher Hesse, Andrew~N. Carr, Jan Leike, Josh Achiam, Vedant Misra, Evan Morikawa, Alec Radford, Matthew Knight, Miles Brundage, Mira Murati, Katie Mayer, Peter Welinder, Bob McGrew, Dario Amodei, Sam McCandlish, Ilya Sutskever, and Wojciech Zaremba. 2021.
\newblock \href {https://arxiv.org/abs/2107.03374} {Evaluating large language models trained on code}.
\newblock \emph{Preprint}, arXiv:2107.03374.

\bibitem[{Cohen et~al.(2024)Cohen, Biran, Yoran, Globerson, and Geva}]{10.1162/tacl_a_00644}
Roi Cohen, Eden Biran, Ori Yoran, Amir Globerson, and Mor Geva. 2024.
\newblock \href {https://doi.org/10.1162/tacl_a_00644} {Evaluating the ripple effects of knowledge editing in language models}.
\newblock \emph{Transactions of the Association for Computational Linguistics}, 12:283--298.

\bibitem[{Comanici et~al.(2025)Comanici, Bieber, Schaekermann, Pasupat, Sachdeva, Dhillon, Blistein, Ram, Zhang, Rosen et~al.}]{comanici2025gemini}
Gheorghe Comanici, Eric Bieber, Mike Schaekermann, Ice Pasupat, Noveen Sachdeva, Inderjit Dhillon, Marcel Blistein, Ori Ram, Dan Zhang, Evan Rosen, et~al. 2025.
\newblock \href {https://arxiv.org/abs/2507.06261} {Gemini 2.5: Pushing the frontier with advanced reasoning, multimodality, long context, and next generation agentic capabilities}.
\newblock \emph{arXiv preprint arXiv:2507.06261}.

\bibitem[{Dao(2024)}]{dao2023flashattention2}
Tri Dao. 2024.
\newblock \href {https://openreview.net/forum?id=mZn2Xyh9Ec} {Flashattention-2: Faster attention with better parallelism and work partitioning}.
\newblock In \emph{The Twelfth International Conference on Learning Representations}.

\bibitem[{De~Cao et~al.(2021)De~Cao, Aziz, and Titov}]{de-cao-etal-2021-editing}
Nicola De~Cao, Wilker Aziz, and Ivan Titov. 2021.
\newblock \href {https://doi.org/10.18653/v1/2021.emnlp-main.522} {Editing factual knowledge in language models}.
\newblock In \emph{Proceedings of the 2021 Conference on Empirical Methods in Natural Language Processing}, pages 6491--6506, Online and Punta Cana, Dominican Republic. Association for Computational Linguistics.

\bibitem[{Feng et~al.(2025)Feng, Wang, Cui, Faltings, Lee, and Zhou}]{feng2025unravelingmisinformationpropagationllm}
Yiyang Feng, Yichen Wang, Shaobo Cui, Boi Faltings, Mina Lee, and Jiawei Zhou. 2025.
\newblock \href {https://arxiv.org/abs/2505.18555} {Unraveling misinformation propagation in llm reasoning}.
\newblock \emph{Preprint}, arXiv:2505.18555.

\bibitem[{Gangadhar and Stratos(2024)}]{gangadhar-stratos-2024-model}
Govind~Krishnan Gangadhar and Karl Stratos. 2024.
\newblock \href {https://doi.org/10.18653/v1/2024.findings-acl.352} {Model editing by standard fine-tuning}.
\newblock In \emph{Findings of the Association for Computational Linguistics: ACL 2024}, pages 5907--5913, Bangkok, Thailand. Association for Computational Linguistics.

\bibitem[{Grattafiori et~al.(2024)Grattafiori, Dubey, Jauhri, Pandey, Kadian, Al-Dahle, Letman, Mathur, Schelten, Vaughan et~al.}]{grattafiori2024llama}
Aaron Grattafiori, Abhimanyu Dubey, Abhinav Jauhri, Abhinav Pandey, Abhishek Kadian, Ahmad Al-Dahle, Aiesha Letman, Akhil Mathur, Alan Schelten, Alex Vaughan, et~al. 2024.
\newblock \href {https://arxiv.org/abs/2407.21783} {The llama 3 herd of models}.
\newblock \emph{Preprint}, arXiv:2407.21783.

\bibitem[{Halevy et~al.(2024)Halevy, Sotnikova, AlKhamissi, Montariol, and Bosselut}]{halevy-etal-2024-flex}
Karina Halevy, Anna Sotnikova, Badr AlKhamissi, Syrielle Montariol, and Antoine Bosselut. 2024.
\newblock \href {https://doi.org/10.18653/v1/2024.emnlp-main.494} {``flex tape can{'}t fix that'': Bias and misinformation in edited language models}.
\newblock In \emph{Proceedings of the 2024 Conference on Empirical Methods in Natural Language Processing}, pages 8690--8707, Miami, Florida, USA. Association for Computational Linguistics.

\bibitem[{Harris et~al.(2020)Harris, Millman, Van Der~Walt, Gommers, Virtanen, Cournapeau, Wieser, Taylor, Berg, Smith et~al.}]{harris2020array}
Charles~R Harris, K~Jarrod Millman, St{\'e}fan~J Van Der~Walt, Ralf Gommers, Pauli Virtanen, David Cournapeau, Eric Wieser, Julian Taylor, Sebastian Berg, Nathaniel~J Smith, et~al. 2020.
\newblock \href {https://www.nature.com/articles/s41586-020-2649-2} {Array programming with numpy}.
\newblock \emph{Nature}, 585(7825):357--362.

\bibitem[{Hase et~al.(2023)Hase, Bansal, Kim, and Ghandeharioun}]{hase2023does}
Peter Hase, Mohit Bansal, Been Kim, and Asma Ghandeharioun. 2023.
\newblock \href {https://openreview.net/forum?id=EldbUlZtbd} {Does localization inform editing? surprising differences in causality-based localization vs. knowledge editing in language models}.
\newblock In \emph{Thirty-seventh Conference on Neural Information Processing Systems}.

\bibitem[{Hase et~al.(2021)Hase, Diab, Celikyilmaz, Li, Kozareva, Stoyanov, Bansal, and Iyer}]{hase2021languagemodelsbeliefsmethods}
Peter Hase, Mona Diab, Asli Celikyilmaz, Xian Li, Zornitsa Kozareva, Veselin Stoyanov, Mohit Bansal, and Srinivasan Iyer. 2021.
\newblock \href {https://arxiv.org/abs/2111.13654} {Do language models have beliefs? methods for detecting, updating, and visualizing model beliefs}.
\newblock \emph{Preprint}, arXiv:2111.13654.

\bibitem[{Hase et~al.(2024)Hase, Hofweber, Zhou, Stengel-Eskin, and Bansal}]{hase2024fundamental}
Peter Hase, Thomas Hofweber, Xiang Zhou, Elias Stengel-Eskin, and Mohit Bansal. 2024.
\newblock \href {https://openreview.net/forum?id=LRf19n5Ly3} {Fundamental problems with model editing: How should rational belief revision work in {LLM}s?}
\newblock \emph{Transactions on Machine Learning Research}.

\bibitem[{Hua et~al.(2024)Hua, Guo, Dong, Zhu, Ng, and Wang}]{hua-etal-2024-propagation}
Wenyue Hua, Jiang Guo, Mingwen Dong, Henghui Zhu, Patrick Ng, and Zhiguo Wang. 2024.
\newblock \href {https://doi.org/10.18653/v1/2024.findings-acl.743} {Propagation and pitfalls: Reasoning-based assessment of knowledge editing through counterfactual tasks}.
\newblock In \emph{Findings of the Association for Computational Linguistics: ACL 2024}, pages 12503--12525, Bangkok, Thailand. Association for Computational Linguistics.

\bibitem[{Huang et~al.(2025)Huang, Chen, Xu, Payani, and Shu}]{huang2025can}
Baixiang Huang, Canyu Chen, Xiongxiao Xu, Ali Payani, and Kai Shu. 2025.
\newblock \href {https://openreview.net/forum?id=hmDt068MoZ} {Can knowledge editing really correct hallucinations?}
\newblock In \emph{The Thirteenth International Conference on Learning Representations}.

\bibitem[{Hunter and Dale(2007)}]{hunter2007matplotlib}
John Hunter and Darren Dale. 2007.
\newblock The matplotlib user’s guide.
\newblock \emph{Matplotlib 0.90. 0 user’s guide}.

\bibitem[{Jiang et~al.(2024)Jiang, Wang, Shen, Kim, and Kim}]{jiang2024surveylargelanguagemodels}
Juyong Jiang, Fan Wang, Jiasi Shen, Sungju Kim, and Sunghun Kim. 2024.
\newblock \href {https://arxiv.org/abs/2406.00515} {A survey on large language models for code generation}.
\newblock \emph{Preprint}, arXiv:2406.00515.

\bibitem[{Kortukov et~al.(2024)Kortukov, Rubinstein, Nguyen, and Oh}]{kortukov2024studying}
Evgenii Kortukov, Alexander Rubinstein, Elisa Nguyen, and Seong~Joon Oh. 2024.
\newblock \href {https://openreview.net/forum?id=xm8zYRfrqE} {Studying large language model behaviors under context-memory conflicts with real documents}.
\newblock In \emph{First Conference on Language Modeling}.

\bibitem[{Kumar et~al.(2023)Kumar, Rothschild, Goldstein, and Hofman}]{kumar2023math}
Harsh Kumar, David~M Rothschild, Daniel~G Goldstein, and Jake~M Hofman. 2023.
\newblock \href {https://papers.ssrn.com/sol3/papers.cfm?abstract_id=4641653} {Math education with large language models: Peril or promise?}
\newblock \emph{Available at SSRN 4641653}.

\bibitem[{Lightman et~al.(2024)Lightman, Kosaraju, Burda, Edwards, Baker, Lee, Leike, Schulman, Sutskever, and Cobbe}]{lightman2023lets}
Hunter Lightman, Vineet Kosaraju, Yuri Burda, Harrison Edwards, Bowen Baker, Teddy Lee, Jan Leike, John Schulman, Ilya Sutskever, and Karl Cobbe. 2024.
\newblock \href {https://openreview.net/forum?id=v8L0pN6EOi} {Let's verify step by step}.
\newblock In \emph{The Twelfth International Conference on Learning Representations}.

\bibitem[{Liu et~al.(2024)Liu, Deng, Li, Wang, Wang, Wang, Zhang, Liu, Wang, Zheng, and Liu}]{liu2024promptinjectionattackllmintegrated}
Yi~Liu, Gelei Deng, Yuekang Li, Kailong Wang, Zihao Wang, Xiaofeng Wang, Tianwei Zhang, Yepang Liu, Haoyu Wang, Yan Zheng, and Yang Liu. 2024.
\newblock \href {https://arxiv.org/abs/2306.05499} {Prompt injection attack against llm-integrated applications}.
\newblock \emph{Preprint}, arXiv:2306.05499.

\bibitem[{Liu et~al.(2025)Liu, Pandit, Ye, Choi, and Durrett}]{liu2025codeupdatearenabenchmarkingknowledgeediting}
Zeyu~Leo Liu, Shrey Pandit, Xi~Ye, Eunsol Choi, and Greg Durrett. 2025.
\newblock \href {https://arxiv.org/abs/2407.06249} {Codeupdatearena: Benchmarking knowledge editing on api updates}.
\newblock \emph{Preprint}, arXiv:2407.06249.

\bibitem[{Longpre et~al.(2021)Longpre, Perisetla, Chen, Ramesh, DuBois, and Singh}]{longpre-etal-2021-entity}
Shayne Longpre, Kartik Perisetla, Anthony Chen, Nikhil Ramesh, Chris DuBois, and Sameer Singh. 2021.
\newblock \href {https://doi.org/10.18653/v1/2021.emnlp-main.565} {Entity-based knowledge conflicts in question answering}.
\newblock In \emph{Proceedings of the 2021 Conference on Empirical Methods in Natural Language Processing}, pages 7052--7063, Online and Punta Cana, Dominican Republic. Association for Computational Linguistics.

\bibitem[{McKinney et~al.(2011)}]{mckinney2011pandas}
Wes McKinney et~al. 2011.
\newblock \href {https://www.researchgate.net/publication/265194455_pandas_a_Foundational_Python_Library_for_Data_Analysis_and_Statistics} {pandas: a foundational python library for data analysis and statistics}.
\newblock \emph{Python for high performance and scientific computing}, 14(9):1--9.

\bibitem[{Meng et~al.(2023{\natexlab{a}})Meng, Bau, Andonian, and Belinkov}]{meng2023locatingeditingfactualassociations}
Kevin Meng, David Bau, Alex Andonian, and Yonatan Belinkov. 2023{\natexlab{a}}.
\newblock \href {https://arxiv.org/abs/2202.05262} {Locating and editing factual associations in gpt}.
\newblock \emph{Preprint}, arXiv:2202.05262.

\bibitem[{Meng et~al.(2023{\natexlab{b}})Meng, Sharma, Andonian, Belinkov, and Bau}]{meng2023masseditingmemorytransformer}
Kevin Meng, Arnab~Sen Sharma, Alex Andonian, Yonatan Belinkov, and David Bau. 2023{\natexlab{b}}.
\newblock \href {https://arxiv.org/abs/2210.07229} {Mass-editing memory in a transformer}.
\newblock \emph{Preprint}, arXiv:2210.07229.

\bibitem[{Mitchell et~al.(2022{\natexlab{a}})Mitchell, Lin, Bosselut, Finn, and Manning}]{mitchell2022fastmodeleditingscale}
Eric Mitchell, Charles Lin, Antoine Bosselut, Chelsea Finn, and Christopher~D. Manning. 2022{\natexlab{a}}.
\newblock \href {https://arxiv.org/abs/2110.11309} {Fast model editing at scale}.
\newblock \emph{Preprint}, arXiv:2110.11309.

\bibitem[{Mitchell et~al.(2022{\natexlab{b}})Mitchell, Lin, Bosselut, Manning, and Finn}]{pmlr-v162-mitchell22a}
Eric Mitchell, Charles Lin, Antoine Bosselut, Christopher~D Manning, and Chelsea Finn. 2022{\natexlab{b}}.
\newblock \href {https://proceedings.mlr.press/v162/mitchell22a.html} {Memory-based model editing at scale}.
\newblock In \emph{Proceedings of the 39th International Conference on Machine Learning}, volume 162 of \emph{Proceedings of Machine Learning Research}, pages 15817--15831. PMLR.

\bibitem[{Onoe et~al.(2023)Onoe, Zhang, Padmanabhan, Durrett, and Choi}]{onoe-etal-2023-lms}
Yasumasa Onoe, Michael Zhang, Shankar Padmanabhan, Greg Durrett, and Eunsol Choi. 2023.
\newblock \href {https://doi.org/10.18653/v1/2023.acl-long.300} {Can {LM}s learn new entities from descriptions? challenges in propagating injected knowledge}.
\newblock In \emph{Proceedings of the 61st Annual Meeting of the Association for Computational Linguistics (Volume 1: Long Papers)}, pages 5469--5485, Toronto, Canada. Association for Computational Linguistics.

\bibitem[{{OpenAI}(2025{\natexlab{a}})}]{openai2025gpt41}
{OpenAI}. 2025{\natexlab{a}}.
\newblock \href {https://openai.com/index/gpt-4-1/} {Introducing gpt-4.1 in the api}.

\bibitem[{{OpenAI}(2025{\natexlab{b}})}]{openai2025o4}
{OpenAI}. 2025{\natexlab{b}}.
\newblock \href {https://openai.com/index/introducing-o3-and-o4-mini/} {Introducing openai o3 and o4-mini}.

\bibitem[{Paszke et~al.(2019)Paszke, Gross, Massa, Lerer, Bradbury, Chanan, Killeen, Lin, Gimelshein, Antiga et~al.}]{paszke2019pytorch}
Adam Paszke, Sam Gross, Francisco Massa, Adam Lerer, James Bradbury, Gregory Chanan, Trevor Killeen, Zeming Lin, Natalia Gimelshein, Luca Antiga, et~al. 2019.
\newblock \href {https://arxiv.org/abs/1912.01703} {Pytorch: An imperative style, high-performance deep learning library}.
\newblock \emph{Advances in neural information processing systems}, 32.

\bibitem[{Poiroux et~al.(2025)Poiroux, Weiss, Kun{\v{c}}ak, and Bosselut}]{poiroux-etal-2025-reliable}
Auguste Poiroux, Gail Weiss, Viktor Kun{\v{c}}ak, and Antoine Bosselut. 2025.
\newblock \href {https://doi.org/10.18653/v1/2025.emnlp-main.907} {Reliable evaluation and benchmarks for statement autoformalization}.
\newblock In \emph{Proceedings of the 2025 Conference on Empirical Methods in Natural Language Processing}, pages 17958--17980, Suzhou, China. Association for Computational Linguistics.

\bibitem[{Rosati et~al.(2024)Rosati, Gonzales, Chen, Yu, Kayani, Rudzicz, and Sajjad}]{rosati-etal-2024-long}
Domenic Rosati, Robie Gonzales, Jinkun Chen, Xuemin Yu, Yahya Kayani, Frank Rudzicz, and Hassan Sajjad. 2024.
\newblock \href {https://doi.org/10.18653/v1/2024.naacl-long.208} {Long-form evaluation of model editing}.
\newblock In \emph{Proceedings of the 2024 Conference of the North American Chapter of the Association for Computational Linguistics: Human Language Technologies (Volume 1: Long Papers)}, pages 3749--3780, Mexico City, Mexico. Association for Computational Linguistics.

\bibitem[{Si et~al.(2023)Si, Friedman, Joshi, Feng, Chen, and He}]{si-etal-2023-measuring}
Chenglei Si, Dan Friedman, Nitish Joshi, Shi Feng, Danqi Chen, and He~He. 2023.
\newblock \href {https://doi.org/10.18653/v1/2023.acl-long.632} {Measuring inductive biases of in-context learning with underspecified demonstrations}.
\newblock In \emph{Proceedings of the 61st Annual Meeting of the Association for Computational Linguistics (Volume 1: Long Papers)}, pages 11289--11310, Toronto, Canada. Association for Computational Linguistics.

\bibitem[{Singh et~al.(2024)Singh, Nambi, and Vineet}]{singh2024exposingachillesheelevaluating}
Joykirat Singh, Akshay Nambi, and Vibhav Vineet. 2024.
\newblock \href {https://arxiv.org/abs/2406.10834} {Exposing the achilles' heel: Evaluating llms ability to handle mistakes in mathematical reasoning}.
\newblock \emph{Preprint}, arXiv:2406.10834.

\bibitem[{Sprague et~al.(2025)Sprague, Yin, Rodriguez, Jiang, Wadhwa, Singhal, Zhao, Ye, Mahowald, and Durrett}]{sprague2025to}
Zayne~Rea Sprague, Fangcong Yin, Juan~Diego Rodriguez, Dongwei Jiang, Manya Wadhwa, Prasann Singhal, Xinyu Zhao, Xi~Ye, Kyle Mahowald, and Greg Durrett. 2025.
\newblock \href {https://openreview.net/forum?id=w6nlcS8Kkn} {To cot or not to cot? chain-of-thought helps mainly on math and symbolic reasoning}.
\newblock In \emph{The Thirteenth International Conference on Learning Representations}.

\bibitem[{Thede et~al.(2025)Thede, Roth, Bethge, Akata, and Hartvigsen}]{thede2025wikibigedit}
Lukas Thede, Karsten Roth, Matthias Bethge, Zeynep Akata, and Thomas Hartvigsen. 2025.
\newblock \href {https://openreview.net/forum?id=9NVm1Bf7CS} {Wikibigedit: Understanding the limits of lifelong knowledge editing in {LLM}s}.
\newblock In \emph{Forty-second International Conference on Machine Learning}.

\bibitem[{TogetherAI(2025)}]{togetherai}
TogetherAI. 2025.
\newblock \href {https://www.together.ai/} {Together ai: The ai acceleration cloud}.

\bibitem[{Vrande\v{c}i\'{c} and Kr\"{o}tzsch(2014)}]{vrandevcic2014wikidata}
Denny Vrande\v{c}i\'{c} and Markus Kr\"{o}tzsch. 2014.
\newblock \href {https://doi.org/10.1145/2629489} {Wikidata: a free collaborative knowledgebase}.
\newblock \emph{Commun. ACM}, 57(10):78–85.

\bibitem[{Wang et~al.(2024)Wang, Feng, Wang, Shi, Balachandran, He, and Tsvetkov}]{wang2024resolving}
Yike Wang, Shangbin Feng, Heng Wang, Weijia Shi, Vidhisha Balachandran, Tianxing He, and Yulia Tsvetkov. 2024.
\newblock \href {https://openreview.net/forum?id=ptvV5HGTNN} {Resolving knowledge conflicts in large language models}.
\newblock In \emph{First Conference on Language Modeling}.

\bibitem[{Waskom(2021)}]{waskom2021seaborn}
Michael~L Waskom. 2021.
\newblock \href {https://joss.theoj.org/papers/10.21105/joss.03021} {Seaborn: statistical data visualization}.
\newblock \emph{Journal of Open Source Software}, 6(60):3021.

\bibitem[{Wolf et~al.(2020)Wolf, Debut, Sanh, Chaumond, Delangue, Moi, Cistac, Rault, Louf, Funtowicz, Davison, Shleifer, von Platen, Ma, Jernite, Plu, Xu, Le~Scao, Gugger, Drame, Lhoest, and Rush}]{wolf2020transformers}
Thomas Wolf, Lysandre Debut, Victor Sanh, Julien Chaumond, Clement Delangue, Anthony Moi, Pierric Cistac, Tim Rault, Remi Louf, Morgan Funtowicz, Joe Davison, Sam Shleifer, Patrick von Platen, Clara Ma, Yacine Jernite, Julien Plu, Canwen Xu, Teven Le~Scao, Sylvain Gugger, Mariama Drame, Quentin Lhoest, and Alexander Rush. 2020.
\newblock \href {https://doi.org/10.18653/v1/2020.emnlp-demos.6} {Transformers: State-of-the-art natural language processing}.
\newblock In \emph{Proceedings of the 2020 Conference on Empirical Methods in Natural Language Processing: System Demonstrations}, pages 38--45, Online. Association for Computational Linguistics.

\bibitem[{Wu et~al.(2024)Wu, Wu, and Zou}]{wu2024clasheval}
Kevin Wu, Eric Wu, and James Zou. 2024.
\newblock \href {https://openreview.net/forum?id=WGoCZl2itU} {Clasheval: Quantifying the tug-of-war between an {LLM}{\textquoteright}s internal prior and external evidence}.
\newblock In \emph{The Thirty-eight Conference on Neural Information Processing Systems Datasets and Benchmarks Track}.

\bibitem[{Xie et~al.(2024)Xie, Zhang, Chen, Lou, and Su}]{xie2024adaptive}
Jian Xie, Kai Zhang, Jiangjie Chen, Renze Lou, and Yu~Su. 2024.
\newblock \href {https://openreview.net/forum?id=auKAUJZMO6} {Adaptive chameleon or stubborn sloth: Revealing the behavior of large language models in knowledge conflicts}.
\newblock In \emph{The Twelfth International Conference on Learning Representations}.

\bibitem[{Xu et~al.(2024{\natexlab{a}})Xu, Lin, Yang, Zhang, Shi, Zhang, Fang, Xu, and Qiu}]{xu-etal-2024-earth}
Rongwu Xu, Brian Lin, Shujian Yang, Tianqi Zhang, Weiyan Shi, Tianwei Zhang, Zhixuan Fang, Wei Xu, and Han Qiu. 2024{\natexlab{a}}.
\newblock \href {https://doi.org/10.18653/v1/2024.acl-long.858} {The earth is flat because...: Investigating {LLM}s' belief towards misinformation via persuasive conversation}.
\newblock In \emph{Proceedings of the 62nd Annual Meeting of the Association for Computational Linguistics (Volume 1: Long Papers)}, pages 16259--16303, Bangkok, Thailand. Association for Computational Linguistics.

\bibitem[{Xu et~al.(2024{\natexlab{b}})Xu, Qi, Guo, Wang, Wang, Zhang, and Xu}]{xu-etal-2024-knowledge-conflicts}
Rongwu Xu, Zehan Qi, Zhijiang Guo, Cunxiang Wang, Hongru Wang, Yue Zhang, and Wei Xu. 2024{\natexlab{b}}.
\newblock \href {https://doi.org/10.18653/v1/2024.emnlp-main.486} {Knowledge conflicts for {LLM}s: A survey}.
\newblock In \emph{Proceedings of the 2024 Conference on Empirical Methods in Natural Language Processing}, pages 8541--8565, Miami, Florida, USA. Association for Computational Linguistics.

\bibitem[{Xu et~al.(2025)Xu, Zhang, Chu, Wang, and Wen}]{xu2024ai}
Tianlong Xu, Yi-Fan Zhang, Zhendong Chu, Shen Wang, and Qingsong Wen. 2025.
\newblock \href {https://doi.org/10.1609/aaai.v39i28.35144} {Ai-driven virtual teacher for enhanced educational efficiency: leveraging large pretrained models for autonomous error analysis and correction}.
\newblock In \emph{Proceedings of the Thirty-Ninth AAAI Conference on Artificial Intelligence and Thirty-Seventh Conference on Innovative Applications of Artificial Intelligence and Fifteenth Symposium on Educational Advances in Artificial Intelligence}, AAAI'25/IAAI'25/EAAI'25. AAAI Press.

\bibitem[{Yang et~al.(2025)Yang, Li, Yang, Zhang, Hui, Zheng, Yu, Gao, Huang, Lv, Zheng, Liu, Zhou, Huang, Hu, Ge, Wei, Lin, Tang, Yang, Tu, Zhang, Yang, Yang, Zhou, Zhou, Lin, Dang, Bao, Yang, Yu, Deng, Li, Xue, Li, Zhang, Wang, Zhu, Men, Gao, Liu, Luo, Li, Tang, Yin, Ren, Wang, Zhang, Ren, Fan, Su, Zhang, Zhang, Wan, Liu, Wang, Cui, Zhang, Zhou, and Qiu}]{yang2025qwen3technicalreport}
An~Yang, Anfeng Li, Baosong Yang, Beichen Zhang, Binyuan Hui, Bo~Zheng, Bowen Yu, Chang Gao, Chengen Huang, Chenxu Lv, Chujie Zheng, Dayiheng Liu, Fan Zhou, Fei Huang, Feng Hu, Hao Ge, Haoran Wei, Huan Lin, Jialong Tang, Jian Yang, Jianhong Tu, Jianwei Zhang, Jianxin Yang, Jiaxi Yang, Jing Zhou, Jingren Zhou, Junyang Lin, Kai Dang, Keqin Bao, Kexin Yang, Le~Yu, Lianghao Deng, Mei Li, Mingfeng Xue, Mingze Li, Pei Zhang, Peng Wang, Qin Zhu, Rui Men, Ruize Gao, Shixuan Liu, Shuang Luo, Tianhao Li, Tianyi Tang, Wenbiao Yin, Xingzhang Ren, Xinyu Wang, Xinyu Zhang, Xuancheng Ren, Yang Fan, Yang Su, Yichang Zhang, Yinger Zhang, Yu~Wan, Yuqiong Liu, Zekun Wang, Zeyu Cui, Zhenru Zhang, Zhipeng Zhou, and Zihan Qiu. 2025.
\newblock \href {https://arxiv.org/abs/2505.09388} {Qwen3 technical report}.
\newblock \emph{Preprint}, arXiv:2505.09388.

\bibitem[{Yao et~al.(2023)Yao, Wang, Tian, Cheng, Li, Deng, Chen, and Zhang}]{yao-etal-2023-editing}
Yunzhi Yao, Peng Wang, Bozhong Tian, Siyuan Cheng, Zhoubo Li, Shumin Deng, Huajun Chen, and Ningyu Zhang. 2023.
\newblock \href {https://doi.org/10.18653/v1/2023.emnlp-main.632} {Editing large language models: Problems, methods, and opportunities}.
\newblock In \emph{Proceedings of the 2023 Conference on Empirical Methods in Natural Language Processing}, pages 10222--10240, Singapore. Association for Computational Linguistics.

\bibitem[{Ying et~al.(2024)Ying, Cao, Xiong, Cui, He, and Liu}]{ying-etal-2024-intuitive}
Jiahao Ying, Yixin Cao, Kai Xiong, Long Cui, Yidong He, and Yongbin Liu. 2024.
\newblock \href {https://doi.org/10.18653/v1/2024.acl-long.232} {Intuitive or dependent? investigating {LLM}s' behavior style to conflicting prompts}.
\newblock In \emph{Proceedings of the 62nd Annual Meeting of the Association for Computational Linguistics (Volume 1: Long Papers)}, pages 4221--4246, Bangkok, Thailand. Association for Computational Linguistics.

\bibitem[{Zhong et~al.(2025)Zhong, Lu, Shao, Bhushanam, Du, Wan, Shi, Zha, Wang, Liu, Zhou, Xu, Chang, Feng, Chaudhary, and Hu}]{zhong2025mquakeremastered}
Shaochen Zhong, Yifan Lu, Lize Shao, Bhargav Bhushanam, Xiaocong Du, Yixin Wan, Yucheng Shi, Daochen Zha, Yiwei Wang, Ninghao Liu, Kaixiong Zhou, Shuai Xu, Kai-Wei Chang, Louis Feng, Vipin Chaudhary, and Xia Hu. 2025.
\newblock \href {https://openreview.net/forum?id=m9wG6ai2Xk} {{MQ}u{AKE}-remastered: Multi-hop knowledge editing can only be advanced with reliable evaluations}.
\newblock In \emph{The Thirteenth International Conference on Learning Representations}.

\bibitem[{Zhong et~al.(2023)Zhong, Wu, Manning, Potts, and Chen}]{zhong-etal-2023-mquake}
Zexuan Zhong, Zhengxuan Wu, Christopher Manning, Christopher Potts, and Danqi Chen. 2023.
\newblock \href {https://doi.org/10.18653/v1/2023.emnlp-main.971} {{MQ}u{AKE}: Assessing knowledge editing in language models via multi-hop questions}.
\newblock In \emph{Proceedings of the 2023 Conference on Empirical Methods in Natural Language Processing}, pages 15686--15702, Singapore. Association for Computational Linguistics.

\bibitem[{Zhu et~al.(2024)Zhu, Hwang, Dugan, and Callison-Burch}]{zhu2024fanoutqamultihopmultidocumentquestion}
Andrew Zhu, Alyssa Hwang, Liam Dugan, and Chris Callison-Burch. 2024.
\newblock \href {https://arxiv.org/abs/2402.14116} {Fanoutqa: A multi-hop, multi-document question answering benchmark for large language models}.
\newblock \emph{Preprint}, arXiv:2402.14116.

\bibitem[{Zhuo et~al.(2025)Zhuo, Chien, Chim, Hu, Yu, Widyasari, Yusuf, Zhan, He, Paul, Brunner, GONG, Hoang, Zebaze, Hong, Li, Kaddour, Xu, Zhang, Yadav, Jain, Gu, Cheng, Liu, Liu, Wang, Lo, Hui, Muennighoff, Fried, Du, de~Vries, and Werra}]{zhuo2024bigcodebench}
Terry~Yue Zhuo, Vu~Minh Chien, Jenny Chim, Han Hu, Wenhao Yu, Ratnadira Widyasari, Imam Nur~Bani Yusuf, Haolan Zhan, Junda He, Indraneil Paul, Simon Brunner, Chen GONG, James Hoang, Armel~Randy Zebaze, Xiaoheng Hong, Wen-Ding Li, Jean Kaddour, Ming Xu, Zhihan Zhang, Prateek Yadav, Naman Jain, Alex Gu, Zhoujun Cheng, Jiawei Liu, Qian Liu, Zijian Wang, David Lo, Binyuan Hui, Niklas Muennighoff, Daniel Fried, Xiaoning Du, Harm de~Vries, and Leandro~Von Werra. 2025.
\newblock \href {https://openreview.net/forum?id=YrycTjllL0} {Bigcodebench: Benchmarking code generation with diverse function calls and complex instructions}.
\newblock In \emph{The Thirteenth International Conference on Learning Representations}.

\end{thebibliography}

\clearpage
\appendix

% Evaluation
\section{Evaluation Metrics}
\label{sec:appendix:evaluation}

\subsection{Knowledge Probing Evaluation}
\label{sec:appendix:evaluation:probing}

We classify an atomic fact $k_i$ as \textit{known} by verifying LLM's responses to the corresponding probing question $q_i$ against the probing answer $a_i$ (i.e., $\text{Verify}(f(q_i), a_i) = \text{True}$). The verification process is as follows. First, we sample $M=10$ responses for the probing question $q_i$ and group them by identical text to obtain unique answers $\{\hat{a}_{i,1}, \cdots, \hat{a}_{i,M}\}$ with corresponding frequencies $\{c_{i,1}, \cdots, c_{i,M}\}$. The equivalence ($\equiv$) between each unique response and the ground-truth probing answer $a_i$ is judged by \texttt{gpt-5-mini} with a tailored prompt for each scenario. Based on these judgments, we compute the total count for the merged group of correct answers, $c_\text{correct} = \sum_{j~s.t.~\hat{a}_{i,j} \equiv a_i} c_{i,j}$. A fact $k_i$ is Known if $c_\text{correct}=\max (c_\text{correct}, c_{i,1}, \cdots, c_{i,M})$. Facts that do not meet this criterion are classified as \textit{unknown}. We also define \textit{Knowledge Confidence (KConf)} as the total correct count normalized by the number of samples: $\text{KConf} = \frac{c_\text{correct}}{N} \times 100\%$.

Our prompt includes detailed instructions and demonstrations tailored to each complex reasoning scenario. The prompts are as follows:

\SmallHeading{Scenario \GROW}

\systemp{
You are given a question, a response, and a ground truth answer. The ground truth answer is an entity, and each response might contain an entity. Your task is to use commonsense knowledge to evaluate whether the response most probably refers the same entity as the ground truth.
\\[1ex]
If they are equivalent, answer 'Yes' and provide an explanation. Otherwise, answer 'No' and provide an explanation.
\\[1ex]
Note that if the response does not contain an entity, it should be treated as 'N/A' and not equivalent to the answer.
\\[1ex]
Examples:
\\[1ex]
Question:

Who is the current US president?

Response:

Therefore, the answer is Donald Trump.

Answer:

Donald J. Trump

Equivalence:

Yes, Donald Trump is the same person as Donald J. Trump.
\\[1ex]
Question:

Who is the current US president?

Response:

Therefore, the answer is Donald Trump.

Answer:

Joe Biden

Equivalence:

No, Donald Trump is a different person from Joe Biden. They belong to different political parties.
\\[1ex]
Question:

Who is Albuquerque’s head of government?

Response:

Based on my cutoff knowledge, Albuquerque’s head of government is Tim Keller.

Answer:

Timothy M. Keller

Equivalence:

Yes, 'Tim' is a common short form of 'Timothy'.
\\[1ex]
Question:

Who is Albuquerque’s head of government?

Response:

I cannot provide an exact answer based on my cutoff knowledge.

Answer:

Timothy M. Keller

Equivalence:

No, the response fails to provide an entity (N/A), while the answer provides the entity 'Timothy M. Keller'.
}

\userp{
Question: 

\{a probing question\}

Response: 

\{a model's response to the probing question\}

Answer: 

\{the ground-truth probing answer\}

Equivalence:
}

\SmallHeading{Scenario \CODE}

\systemp{
You are given a question, a canonical function from a library for the question, and the model's response. Each response might also contain a function call from a library. Your task is to use basic Python coding knowledge to evaluate whether the model's response is most probably correct.
\\[1ex]
If the answer is correct, answer 'Yes' and provide an explanation. Otherwise, answer 'No' and provide an explanation.
\\[1ex]
Note that if the response does not contain a function call, it should be treated as 'N/A' and not correct.
\\[1ex]
--- Examples 1 ---
\\[1ex]
Question:

Given the library pandas, how can we create a DataFrame by explicitly passing the input data (such as an ndarray, Iterable, dict, or DataFrame) using the `data` parameter?

Function:

pandas.DataFrame(data)

Response:

```python

pandas.DataFrame(arr)

```

Correct:

Yes, the response contains the same function call as the ground truth function call. The function call `pandas.DataFrame(arr)` is equivalent to the ground truth function call, which creates a DataFrame from the provided data.
\\[1ex]
--- Example 2 ---
\\[1ex]
Question: 

Given the library pandas, how can we create a DataFrame by explicitly passing the input data (such as an ndarray, Iterable, dict, or DataFrame) using the `data` parameter?

Function: 

pandas.DataFrame(data)

Response:

```python

pandas.DataFrame({"id": [0, 1, 2, 3, 4], "val": [100, 200, -2, 34, 45.2]}, dtype=None)

```

Correct:

Yes, the response contains the same function call as the ground truth function call. The function call `pandas.DataFrame({"id": [0, 1, 2, 3, 4], "val": [100, 200, -2, 34, 45.2]}, dtype=None)` is equivalent to the ground truth function call, which creates a DataFrame from the provided data. The `dtype` parameter is optional and defaults to None, so it does not change the equivalence.
\\[1ex]
--- Example 3 ---
\\[1ex]
Question: 

Given the library pandas, how can we create a DataFrame by explicitly passing the input data (such as an ndarray, Iterable, dict, or DataFrame) using the `data` parameter?

Function: 
pandas.DataFrame(data)

Response:

```python

pandas.DataFrame(data, dtype="float")

```

Correct:

No, the response contains a different function call than the ground truth function call. The function call `pandas.DataFrame(data, dtype="float")` specifies a dtype of "float", which is not equivalent to the ground truth function call that does not specify a dtype. The ground truth function call creates a DataFrame from the provided data without any specific dtype.
}

\userp{
Question: 

\{a probing question\}

Function: 

\{the ground-truth API calling function\}

Response: 

\{the model's response of API calling\}

Correct:
}

\SmallHeading{Scenario \MATH}

\systemp{
You are given a question, a response, and a ground truth answer. Your task is to use math knowledge and ground truth to evaluate whether the response most probably answers the question.
\\[1ex]
If the response answers the question, answer 'Yes' and provide an explanation. Otherwise, answer 'No' and provide an explanation.
\\[1ex]
Note that the ground truth is just a reference of correct answer. An answer is correct does not mean that it should be strictly same as the ground truth. You should perform mathematical knowledge to infer that.
\\[1ex]
Examples:
\\[1ex]
Question:

Given the equations $3p+e=1.24$ and $5p+e=1.82$, what specific operation will eliminate the variable 'e'?

Response:

Subtracting the first equation from the second one, then we will have $2p=0.58$ and eliminate 'e'.

Answer:

Given the equations $3p+e=1.24$ and $5p+e=1.82$, subtracting the first equation from the second will eliminate the variable 'e'.

Correctness:

Yes, it mentions what it is in the ground truth that subtracting one equation from another will eliminate the variable 'e'.
\\[1ex]
Question:

If half the value of 3x - 9 is equal to x + 37, how can this relationship be written as an equation?

Response:

3x - 9 = 2(x + 37)

Answer:

If half the value of 3x - 9 is equal to x + 37, it can be written as (1/2)(3x - 9) = x + 37.

Correctness:

Yes, it mentions 3x - 9 = 2(x + 37) while the ground truth mentions (1/2)(3x - 9) = x + 37. Though they are different, they are mathematically equivalent and the response is also correct.
\\[1ex]
Question:

How do you combine the like imaginary terms -3i and 12i in a sum?

Response:

To combine the like imaginary terms, add or subtract their coefficients, which are the numbers in front of the 'i'.

Answer:

Like terms with the same imaginary unit i are combined by adding their coefficients, so -3 + 12 = 9, giving 9i.

Correctness:

Yes, though the response does not explicitly give 9i, the answer "add or subtract their coefficients, which are the numbers in front of the 'i'" is also valid because it tells how to imaginary terms -3i and 12i.
\\[1ex]
Question:

According to the associative property of addition, how can we regroup the sum 6 - 3i + 12i?

Response:

The associative property of addition states that the grouping of terms does not affect their sum, so 6 - 3i + 12i can be written as (6 - 3i) + 12i.

Answer:

We can regroup the sum as 6 + (12i - 3i).

Correctness:

Yes, though the response does not match the specific regrouping given in the answer, both are valid regroupings due to the associative property.
\\[1ex]
Question:

Given the equations $3p+e=1.24$ and $5p+e=1.82$, what specific operation will eliminate the variable 'e'?

Response:

Adding the first equation to the second one will eliminate 'e'.

Answer:

Given the equations $3p+e=1.24$ and $5p+e=1.82$, subtracting the first equation from the second will eliminate the variable 'e'.

Correctness:

No, it mentions adding one equation to another, different from the ground truth answer which subtracts the first equation from the second one.
\\[1ex]
Question:

Given the equations $3p+e=1.24$ and $5p+e=1.82$, what specific operation will eliminate the variable 'e'?

Response:

After eliminating 'e', we have $p=0.29$.

Answer:

Given the equations $3p+e=1.24$ and $5p+e=1.82$, subtracting the first equation from the second will eliminate the variable 'e'.

Correctness:

No, it fails to mention that subtracting the first equation from the second one will eliminate 'e'.
}

\userp{
Question: 

\{a probing question\}

Response: 

\{a model's response to the probing question\}

Answer: 

\{the ground-truth probing answer\}

Correctness:
}

\subsection{Knowledge Injection Evaluation}
\label{sec:appendix:evaluation:injection}

For each complex reasoning question $q$, we evaluate a model $f$'s response in both closed-book $f(q)$ and open-book settings $f(q,C)$, which consists of reasoning steps $\hat{R}$ and a final answer $\hat{a}$, using three metrics: Answer Pass (AP) for final answer correctness, Full Knowledge Entailment (FKE) for reasoning faithfulness, and Holistic Pass (HP), which requires both AP and FKE to be true ($\text{HP} = \text{AP} \wedge \text{FKE}$).

The specific implementation of AP and FKE is tailored to each scenario, though the FKE check always involves using \texttt{gpt-5-mini} to perform Natural Language Inference (NLI).

\begin{itemize}[leftmargin=*]
    \item For the \GROW~and \MATH~scenarios: The AP is determined by using \texttt{gpt-4.1-mini} to assess the equivalence between the model's extracted final answer $\hat{a}$ and the ground-truth answer $a^*$. The FKE check verifies that the reasoning steps $\hat{R}$ entail each required atomic fact.

    \item For the \CODE~scenario: The evaluation is based on functional correctness. The model's generated function is treated as the reasoning steps $\hat{R}$. The AP is true if this function passes the predefined unit tests from BigCodeBench~\citep{zhuo2024bigcodebench}. The FKE check verifies that the implemented code in $\hat{R}$ is faithful to the functionality described in the provided API documentation (the atomic facts).
\end{itemize}

Below are detailed prompts for AP and FKE metrics together with \GROW, \CODE, and \MATH scenarios.

\subsubsection{Answer Pass (AP)}

AP is a binary metric of correctness for the final answer $\hat{a}$: $\text{AP} = \text{Is\_Correct}(\hat{a}, a)$. Its verification method is domain-specific. For \GROW~and \MATH, the \texttt{gpt-5-mini} model assesses the equivalence between the model's extracted final answer and the ground truth $a$, i.e., $\text{Is\_Correct}(\hat{a}, a)=\mathbb{I}(\hat{a} \equiv a)$. For \CODE, AP is determined by functional correctness, using standard Pass@1~\citep{chen2021evaluatinglargelanguagemodels}: AP is true if the executed function passes a predefined suite of unit tests once, i.e., $\text{Is\_Correct}(\hat{a}, a)=\text{Pass@1}(\hat{a})$. Given some implementation issues of building unit test environments in the original BigCodeBench, we use GPT-5-mini as a judge to detect whether the LLM-generated code can pass the unit test compared with the ground truth code.

\SmallHeading{Scenario \GROW}
We use the same prompt as Scenario \GROW in \S~\ref{sec:appendix:evaluation:probing}.

\SmallHeading{Scenario \CODE}
We pass the function to predefined unit tests from BigCodeBench~\citep{zhuo2024bigcodebench}. The unit test result is treated as the final answer $\hat{a}$, and is correct if it passes the unit test.

\SmallHeading{Scenario \MATH}

\systemp{
You are given a question, a final step of response, and a ground truth answer. The final step may contain a step number and "the answer is ...". PLEASE IGNORE THE STEP NUMBER. Your task is to use math knowledge to evaluate whether the response is mathematically equivalent to the ground truth answer.
\\[1ex]
If they are equivalent, answer 'Yes' and provide an explanation. Otherwise, answer 'No' and provide an explanation.
\\[1ex]
Examples:
\\[1ex]
Question:

Evaluate $(1+2i)6-3i$.

Response:

7. The answer is $9i+6$.

Ground Truth:

6+9i

Correctness:

Yes, the answer from response $9i+6$ is equivalent to the ground truth $6+9i$.
\\[1ex]
Question:

Evaluate $(1+2i)6-3i$.

Response:

7. The answer is $9+6i$.

Ground Truth:

6+9i

Correctness:

No, the answer from response is different from ground truth. The real part and the imaginary part are reversed.
\\[1ex]
Question:

The lengths of two opposite sides of a square are decreased by $40\%$ while the lengths of the other two sides are increased by $50\%$ to form a rectangle. By what percent does the square's area decrease?

Response:

Therefore, the square's area decreases by 10\%.

Ground Truth:

10

Correctness:

Yes, the response "10\%" is equivalent to the ground truth "10" because the question asks for a percentage.
}

\userp{
Question: 

\{a complex reasoning question\}

Response: 

\{a model's final step to the complex reasoning question\}

Answer: 

\{the ground-truth final answer\}

Correctness:
}

\subsubsection{Full Knowledge Entailment (FKE)}

FKE is a binary metric that evaluates whether the reasoning process is faithful to the atomic facts. The entailment check is performed by \texttt{gpt-5-mini}, guided by domain-specific NLI prompts. FKE is true if the reasoning steps $\hat{R}$ together with final answer $\hat{a}$ entail every required atomic fact $k_i$ from the full set $K_q$: $\text{FKE} = \bigwedge_{k_i \in K_q} \mathbb{I}(\text{NLI}(k_i, \hat{R}) = \text{Entailment})$.

\SmallHeading{Scenario \GROW}

\systemp{
You are an expert in natural language inference and commonsense reasoning. You will be given a "Context" (the model's reasoning response) and a "Statement" (a piece of knowledge). Your task is to determine if the Context finally entails, contradicts, or is neutral with respect to the Statement.
\\[1ex]
Answer "Entailment", "Contradiction", or "Neutral" and provide a brief explanation of your reasoning.
\\[1ex]
Note that if the Context mentions some knowledge is "unknown", it should be treated as "N/A" and contradictory to the Statement.
\\[1ex]
Examples:
\\[1ex]
Context:

1. As of my knowledge cutoff in December 2023, the current Prime Minister of the United Kingdom is Rishi Sunak, and Keir Starmer's spouse is Victoria Starmer.

2. The user provided that Keir Starmer is the current head of government. As of my knowledge, he is the leader of the Opposition and the leader of the Labour Party in the UK. So it is possible that he is the Prime Minister now.

3. Therefore, the current British Prime Minister is Keir Starmer.

4. Victoria Starmer is married to British Prime Minister.

5. The answer is Victoria Starmer.

Statement:

The name of the current head of the British government is Keir Starmer.

NLI:

Entailment. The context initially stated that Rishi Sunak is the current Prime Minister, but later it corrected that Keir Starmer is the current head of government.
\\[1ex]
Context:

... (the same context as above)

Statement:

The name of the current head of the British government is Rishi Sunak.

NLI:

Contradiction. The context initially stated that Rishi Sunak is the current Prime Minister, but later it corrected that Keir Starmer is the current head of government.
\\[1ex]
Context:

... (the same context as above)

Statement:

The name of the current head of the US government is Donald Trump.

NLI:

Neutral. The context does not provide any information about the current head of the US government.
\\[1ex]
Context:

1. I don't know the current head of the British government.

2. I don't have the information about the answer.

Statement:

The name of the current head of the British government is Keir Starmer.

NLI:

Contradiction. The context states that it does not know the current head of the British government, which contradicts the statement that Keir Starmer is the current head of the British government.
\\[1ex]
Context:

... (the same context as above)

Statement:

The name of the current head of the US government is Donald Trump.

NLI:

Neutral. The context does not provide any information about the current head of the US government.
}

\userp{
Context: 

\{a model's generated reasoning steps\}

Statement: 

\{one of the atomic facts required to answer the complex reasoning question\}

NLI:
}

\SmallHeading{Scenario \CODE}

\systemp{
You are an expert in Python programming and natural language inference. You will be given a 'Code' snippet and a 'Function'. Your task is to determine if the Code's usage of the function **Entails**, **Contradicts**, or is **Neutral** with respect to the correct usage of the function.
\\[1ex]
**Reasoning Framework:**
\\[1ex]
1.  **Check for Usage:** First, determine if the `Code` calls the specified function. If it does not, the answer is **Neutral**.
\\[1ex]
2.  **Validate the Call:** If the function is called, analyze the arguments used in the `Code` against the `Function Signature`.

\quad\quad* **Entailment:** The usage is valid. This means:

\quad\quad\quad\quad* All required arguments are provided.

\quad\quad\quad\quad* Any keyword arguments used are valid (i.e., they exist in the function signature).

\quad\quad\quad\quad* **Crucially, omitting optional arguments (e.g., those with default values like `r=None`) is a valid use case.**

\quad\quad* **Contradiction:** The usage is invalid. This means:

\quad\quad\quad\quad* A required argument is missing.

\quad\quad\quad\quad* An incorrect or non-existent keyword argument is used (e.g., `datatype=` when it should be `dtype=`).
\\[1ex]
--- Example 1 ---
\\[1ex]
Code:

```python

import pandas as pd

def task\_func(dealer\_sales\_data):

\quad\quad\# Step 1: Create DataFrame \& Step 2: Handle Empty Input (if dealer\_sales\_data is empty)

\quad\quad df = pd.DataFrame(dealer\_sales\_data, dtype=None)

\quad\quad if not dealer\_sales\_data:

\quad\quad\quad\quad return []

\quad\quad\# Ensure 'id' and 'num\_sold' columns exist, otherwise it's malformed input

\quad\quad if 'id' not in df.columns or 'num\_sold' not in df.columns:

\quad\quad\quad\quad return []

\quad\quad\# Step 3: Find Max Sales

\quad\quad max\_sold = df['num\_sold'].max()

\quad\quad\# Step 4: Identify Top Sellers

\quad\quad top\_selling\_cars = df[df['num\_sold'] == max\_sold]

\quad\quad\# Step 5: Extract and Sort IDs

\quad\quad top\_selling\_ids = top\_selling\_cars['id'].tolist()

\quad\quad sorted\_ids = sorted(top\_selling\_ids)

\quad\quad\# Step 6: Return Result

\quad\quad return sorted\_ids

```
\\[1ex]
Function: pandas.DataFrame(data)
\\[1ex]
NLI:

Entailment. The code contains the function call `pd.DataFrame(dealer\_sales\_data)`. The usage `pd.DataFrame(dealer\_sales\_data)` is entailed in the provided function. Only one positional parameter and the optional parameter `dtype` is set to `None`, which is by default.
\\[1ex]
--- Example 2 ---
\\[1ex]
Code:

```python

import pandas as pd

def task\_func(dealer\_sales\_data):

\quad\quad\# Step 1: Create DataFrame \& Step 2: Handle Empty Input (if dealer\_sales\_data is empty)

\quad\quad df = pd.DataFrame(dealer\_sales\_data, dtype="float")

\quad\quad if not dealer\_sales\_data:

\quad\quad\quad\quad return []

\quad\quad\# Ensure 'id' and 'num\_sold' columns exist, otherwise it's malformed input

\quad\quad if 'id' not in df.columns or 'num\_sold' not in df.columns:

\quad\quad\quad\quad return []

\quad\quad\# Step 3: Find Max Sales

\quad\quad max\_sold = df['num\_sold'].max()

\quad\quad\# Step 4: Identify Top Sellers

\quad\quad top\_selling\_cars = df[df['num\_sold'] == max\_sold]

\quad\quad\# Step 5: Extract and Sort IDs

\quad\quad top\_selling\_ids = top\_selling\_cars['id'].tolist()

\quad\quad sorted\_ids = sorted(top\_selling\_ids)

\quad\quad\# Step 6: Return Result

\quad\quad return sorted\_ids

```
\\[1ex]
Function: pandas.DataFrame(data)
\\[1ex]
NLI:

Contradiction. The code contains a related function call `pd.DataFrame(dealer\_sales\_data, datatype="float")`, but it uses a different keyword argument `datatype` which does not exist. The usage of the function contradicts with the correct usage of of the provided function.
\\[1ex]
--- Example 3 ---
\\[1ex]
Code:

... (the same code as above)
\\[1ex]
Function: sklearn.linear\_model.LinearRegress

ion()
\\[1ex]
NLI:

Neutral. The code does not contain any function call related to the provided function.
\\[1ex]
--- Example 4 ---
\\[1ex]
Code:

N/A
\\[1ex]
Function: ... (any function)
\\[1ex]
NLI:

Neutral. The code does not contain any function call related to the provided function.

}

\userp{
Code: 

\{a model's generated code\}

Function:  \{one of the required external function to solve the coding problem\}

NLI:
}

\SmallHeading{Scenario \MATH}

\systemp{
You are an expert in natural language inference and math reasoning. You will be given a "Context" (the model's reasoning response) and a "Statement" (a piece of knowledge). Your task is to determine if the Context finally entails, contradicts, or is neutral with respect to the Statement.
\\[1ex]
Answer "Entailment", "Contradiction", or "Neutral" and provide a brief explanation of your reasoning.
\\[1ex]
Note that if the Statement can be indirectly implied from the Context with math reasoning, it should also be treated as entailment.
\\[1ex]
Examples:
\\[1ex]
Context:

Three pencils and a jumbo eraser cost \$1.24. Five pencils and a jumbo eraser cost \$1.82. No prices include tax. In cents, what is the cost of a pencil?

1. Let's call the price of a pencil x and the price of a jumbo eraser y. Then we can write two equations.

2. We have $3x+y=1.24$ and $5x+y=1.82$.

3. To solve this system, let's subtract the first equation from the second equation. This will eliminate y.

4. This simplifies to $x=0.29$.

5. That means a pencil costs 29 cents.

6. The answer is 29 cents.

Statement:

After subtracting $3p+e=1.24$ from $5p+e=1.82$, we will have $2p = 0.58$, which solves to $p = 0.29$.

NLI:

Entailment. The statement is a direct summary of the mathematical reasoning presented in steps 2, 3, and 4 in the context. It describes the exact same process and reaches the identical conclusion. The use of different variables (p and e instead of x and y) is a superficial change that doesn't affect the logic, and missing $2p = 0.58$ does not matter because $x = 0.29$ already implies it.
\\[1ex]
Context:

Three pencils and a jumbo eraser cost \$1.24. Five pencils and a jumbo eraser cost \$1.82. No prices include tax. In cents, what is the cost of a pencil?

1. Let's call the price of a pencil x and the price of a jumbo eraser y. Then we can write two equations.

2. We have $3x+y=1.24$ and $5x+y=1.82$.

3. To solve this system, let's add the first equation to the second equation. This will eliminate y.

4. This simplifies to $2x=0.58$. So $x=0.29$.

5. That means a pencil costs 29 cents.

6. The answer is 29 cents.

Statement:

After subtracting $3p+e=1.24$ from $5p+e=1.82$, we will have $2p = 0.58$, which solves to $p = 0.29$.

NLI:
Contradiction. The context states in step 3 that the two equations should be added. The statement, however, describes the process using subtraction. Since adding and subtracting are opposite operations, the statement directly contradicts the method described in the context.
\\[1ex]
Context:

Three pencils and a jumbo eraser cost \$1.24. Five pencils and a jumbo eraser cost \$1.82. No prices include tax. In cents, what is the cost of a pencil?

1. Let's call the price of a pencil x and the price of a jumbo eraser y. Then we can write two equations.

2. We have $3x+y=1.24$ and $5x+y=1.82$.

3. To solve this system, let's add the first equation to the second equation. This will eliminate y.

4. This simplifies to $2x=0.58$. So $x=0.29$.

5. That means a pencil costs 29 cents.

6. The answer is 29 cents.

Statement:

The commutative rule, also known as the commutative property, states that the order of numbers in addition and multiplication doesn't change the result.

NLI:

Neutral. The context does not provide any information about the commutative rules. The statement is completely irrelevant to the context.
}

\userp{
Context: 

\{a model's generated reasoning steps\}

Statement: 

\{one of the atomic facts required to answer the complex reasoning question\}

NLI:
}

\subsubsection{Holistic Pass (HP)}

HP is a stricter binary metric requiring both a correct final answer and faithful reasoning: $\text{HP} = \text{AP} \wedge \text{FKE}$.

% Dataset
\section{Dataset Generation}
\label{sec:appendix:dataset}

\begin{figure*}[t!]
    \begin{center}
        \includegraphics[width=0.99\textwidth]{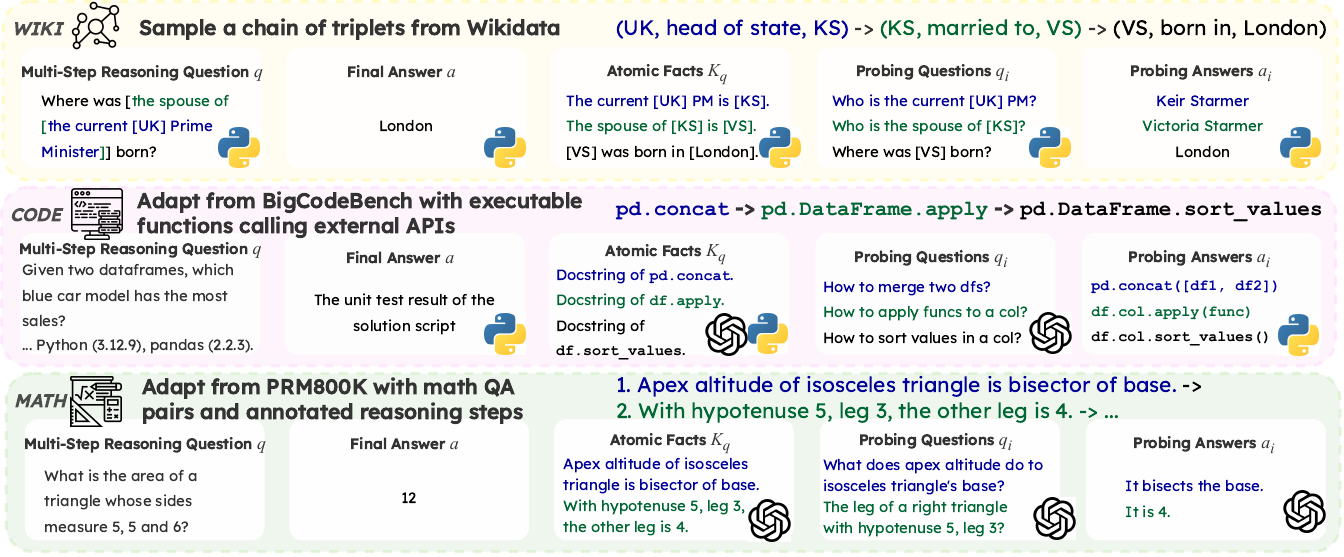}
    \end{center}
    \caption{\textbf{The data generation pipeline for our three \taskalias~scenarios.} Each pipeline sources from Wikidata~\citep{vrandevcic2014wikidata}, BigCodeBench~\citep{zhuo2024bigcodebench}, or PRM800K~\citep{lightman2023lets}, and produces five key components: the multi-step reasoning question $q$, final answer $a$, atomic facts $K_q$, probing questions $q_i$, and probing answers $a_i$. The generation method for each component is denoted by an icon: \pythonicon~for deterministic scripts, \openaiicon~for LLM-based generation, and both for a hybrid approach. Components with no icon are directly adapted from existing benchmarks.}
\label{fig:data}
\end{figure*}

The general pipeline for each scenario is visualized in \Figure~\ref{fig:data}. Here we provide additional details on the data generation pipeline for each of our three scenarios, supplementing the descriptions in \secref{sec:main:dataset}.

\subsection{Scenario \WIKI}
\label{sec:appendix:dataset:grow}

For this scenario, we generate multi-hop QA by sampling relational chains from recent Wikidata~\citep{vrandevcic2014wikidata}.\footnote{We sample chains from Wikidata in Sep. 2025.} Our pipeline first samples a chain of varied depths (2 to 6 hops) that satisfies multiple constraints, such as ensuring entities are unique and have valid labels. From this chain, we automatically construct the five key data components. The complex reasoning question ($q$) is generated by composing templated natural-language questions for each relation in the chain. The final entity in the chain serves as the final answer ($a$). Each triple in the chain is converted into a declarative sentence and an interrogative sentence to form an atomic fact ($k_i$) and a probing question ($q_i$), respectively. The object entity of the triple serves as a probing answer ($a_i$). Note that questions involving Wikidata may contain information of entities, but such information is public available and it does not include offensive content.

% The data for the \WIKI~scenario is generated by first programmatically sampling multi-hop relational chains from the Wikidata knowledge graph. Each sampled chain serves as the backbone for a single data instance. We then apply a series of deterministic, template-based transformations to this chain to automatically construct the five key data components: the complex reasoning question ($q$), the final answer ($a$), the set of atomic facts ($K_q$), and the corresponding probing questions ($q_i$) and answers ($a_i$).

\subsubsection*{Chain Sampling Constraints}
To ensure the quality, complexity, and realism of the generated data, we enforce several constraints during the chain sampling process:
\begin{itemize}[leftmargin=*]
    \item Varied Depth: We sample 120, 100, 100, 100, 80 chains with length of 2, 3, 4, 5, 6 hops (i.e., contain 2, 3, 4, 5, 6 triples), respectively.
    \item No Cycles: An entity cannot appear more than once in a given chain, preventing trivial loops.
    \item Property Whitelist: All relations (properties) in the chain must belong to a predefined set of allowed properties (e.g., `P39': position held, `P26': spouse), focusing the data on common and meaningful relationships.
    \item Single Claim Constraint: We only consider properties for which an entity has a single, unambiguous claim, avoiding ambiguity from multiple values.
    \item Label Validity: All entities and properties in the chain must have non-empty English labels. We also filter out entities with excessively long labels (more than 5 words or 50 characters) to maintain readability.
    \item Type Constraints: For certain properties, we programmatically verify that the subject or object entity is of the correct type using SPARQL queries (e.g., for `P36' `capital of', the subject must be an instance of a country `wd:Q6256').
\end{itemize}

\subsubsection*{Component Generation from Templates}
Once a valid chain is sampled, its components are generated using the templates detailed in \Table~\ref{tab:appendix:grow_templates}.
\begin{itemize}[leftmargin=*]
    \item Complex Reasoning Question ($q$): We recursively compose a natural language question. Starting with the head entity of the chain, we use the ``attributive'' phrase from our templates (e.g., ``the spouse of [S]'') for each hop, nesting them to form a complex question.
    \item Final Answer ($a$): This is simply the English label of the final entity in the 4-hop chain.
    \item Atomic Fact ($k_i$): For each triple `(Subject, Property, Object)' in the chain, we generate a declarative sentence using a cloze-style template (e.g., ``[S] is married to [O].'').
    \item Probing Question ($q_i$): For each triple, we use a question template corresponding to the property (e.g., ``Who is [S] married to?'').
    \item Probing Answer ($a_i$): This is the English label of the object entity in the triple.
\end{itemize}

\begin{table*}[t!]
    \centering
    \resizebox{\textwidth}{!}{
    \begin{tabular}{llll}
    \toprule
    \textbf{Property ID} & \textbf{Question Template} & \textbf{Attributive Phrase Template} & \textbf{Cloze Template} \\
    \midrule
    P20 & Which city did [S] die in? & the city where [S] died & [S] died in the city of [O]. \\
    P26 & Who is [S] married to? & the person who is married to [S] & [S] is married to [O]. \\
    P27 & What is the country of citizenship of [S]? & the country of citizenship of [S] & [S] is a citizen of [O]. \\
    P35 & Who is the current head of state in [S]? & the person who is the current head of state in [S] & The current head of state in [S] is [O]. \\
    P36 & What is the capital of [S]? & the capital of [S] & The capital of [S] is [O]. \\
    P37 & What is the official language of [S]? & the official language of [S] & The official language of [S] is [O]. \\
    P39 & What position is held by [S]? & the position held by [S] & The position held by [S] is [O]. \\
    P40 & Who is [S]’s child? & the child of [S] & [S]’s child is [O]. \\
    P50 & Who is the author of [S]? & the author of [S] & The author of [S] is [O]. \\
    P69 & Which university was [S] educated at? & the university where [S] was educated & The univerisity where [S] was educated is [O]. \\
    P106 & What kind of work does [S] do? & the field of work of [S] & [S] works in the field of [O]. \\
    P108 & Which organization is the employer of [S]? & the organization that employs [S] & [S] is employed by the organization [O]. \\
    P112 & Who founded [S]? & the founder of [S] & [S] was founded by [O]. \\
    P136 & What type of music does [S] play? & the type of music that [S] plays & The type of music that [S] plays is [O]. \\
    P140 & Which religion is [S] affiliated with? & the religion that [S] is affiliated with & [S] is affiliated with the religion of [O]. \\
    P159 & Which city is the headquarter of [S] located in? & the city where the headquarters of [S] is located & The headquarters of [S] is located in the city of [O]. \\
    P169 & Who is the chief executive officer of [S]? & the chief executive officer of [S] & The chief executive officer of [S] is [O]. \\
    P170 & Who was [S] created by? & the creator of [S] & [S] was created by [O]. \\
    P175 & Who performed [S]? & the performer of [S] & [S] was performed by [O]. \\
    P176 & Which company is [S] produced by? & the company that produced [S] & The company that produced [S] is [O]. \\
    P178 & Who is the developer of [S]? & the developer of [S] & [S] was developed by [O]. \\
    P286 & Who is the head coach of [S]? & the head coach of [S] & The head coach of [S] is [O]. \\
    P364 & What is the original language of [S]? & the original language of [S] & The original language of [S] is [O]. \\
    P407 & Which language was [S] written in? & the language that [S] was written in & The language that [S] was written in [O]. \\
    P413 & What position does [S] play? & the position that [S] plays & [S] plays the position of [O]. \\
    P449 & Who is the original broadcaster of [S]? & the original broadcaster of [S] & The original broadcaster of [S] is [O]. \\
    P488 & Who is the chairperson of [S]? & the chairperson of [S] & The chairperson of [S] is [O]. \\
    P495 & Which country was [S] created in? & the country where [S] was created & [S] was created in the country of [O]. \\
    P641 & Which sport is [S] associated with? & the sport that [S] is associated with & [S] is associated with the sport of [O]. \\
    P740 & Which city was [S] founded? & the city where [S] was founded & [S] was founded in the city of [O]. \\
    P800 & What is [S] famous for? & the thing that [S] is famous for & [S] is famous for [O]. \\
    P937 & Which city did [S] work in? & the city where [S] worked & [S] worked in the city of [O]. \\
    P1037 & Who is the director of [S]? & the director of [S] & The director of [S] is [O]. \\
    P1412 & What language does [S] speak? & the language that [S] speaks & [S] speaks the language of [O]. \\
    \bottomrule
\end{tabular}
    }
    \caption{Templates used for generating questions and facts in the \WIKI~scenario. `[S]' is a placeholder for the subject entity's label, and `[O]' is for the object entity.}
    \label{tab:appendix:grow_templates}
\end{table*}

\subsection{Scenario \CODE}
\label{sec:appendix:dataset:code}

For this scenario, we adapt code problems from the BigCodeBench benchmark~\citep{zhuo2024bigcodebench}. The process involves programmatically identifying the necessary API calls for a given coding problem and then using \texttt{gpt-5-mini} to generate the final data components based on this extracted knowledge. Given a coding problem $q$ as a complex reasoning question, like ``Given two dataframes, find which blue car model has the most sales?'', the model must generate a complete, executable function as its reasoning process. The execution result of the function serves as the final answer $a$. The knowledge conflict occurs when the model has an outdated or incorrect knowledge of a required API call, which serves as an atomic fact $k_i$ (e.g., the new \texttt{pandas.concat} function replaces the old API \texttt{pandas.DataFrame.append}); this is probed by asking for its purpose ($q_i$) and expecting the function name and required arguments as the probing answer ($a_i$). We set temperature to 0.7, top P to 0.7, and max tokens to 4096. Note that the coding questions do not contain information that identifies individual people and do not include offensive content.

\subsubsection*{Component Generation Pipeline}
\begin{itemize}[leftmargin=*]
    \item Complex Reasoning Question ($q$) and Final Answer ($a$): We directly adapt these from BigCodeBench. The problem's instruction prompt serves as the question $q$. The final answer $a$ is not a static string but is determined by the result of executing the model's generated code against the problem's predefined unit tests.

    \item Atomic Fact ($k_i$) Extraction and Generalization: This is a multi-step process.
    \begin{enumerate}
        \item We first use a custom Python script with an Abstract Syntax Tree (AST) visitor to parse the canonical solution for each problem. This visitor identifies all calls to external libraries (e.g., \texttt{pandas}, \texttt{numpy}) and resolves their canonical names, handling import aliases.
        \item For each unique canonical function name identified, we programmatically inspect the installed library to retrieve its official function signature and docstring. This information is cached to ensure consistency.
        \item To create a generalized representation of an API call, we substitute the specific variables used in the source code with their formal parameter names from the retrieved signature. For example, two distinct calls in the source code, \texttt{pd.concat([df1, df2])} and \texttt{pd.concat([data\_a, data\_b])}, are both normalized to the same underlying knowledge representation corresponding to \texttt{pd.concat(objs=...)}. This ensures that the atomic fact represents the general usage of the API, not the specific variables in one instance.
        \item The summarized docstring for the API call is then provided to \texttt{o4-mini} with a structured prompt. The LLM's task is to generate a final, declarative sentence that describes the function's purpose. This sentence becomes the atomic fact $k_i$.
    \end{enumerate}

    \item Probing Question ($q_i$) and Answer ($a_i$): Using the same LLM call described above, we also prompt the model to generate a natural language question about the API's purpose ($q_i$). The probing answer ($a_i$) is the generalized, canonical function call itself (e.g., \texttt{pandas.concat(objs)}). This tests if a model can recall the correct API call for a given task.
\end{itemize}

\subsubsection*{Detailed Prompts}
\systemp{
You are a helpful assistant that analyzes code and generates questions about library calls.

Your task is to generate probe questions based on the provided code and library calls.
}

For the user prompt, we replace ``[ANSWER]'' with the generalized representation of the API, ``[DOCSTRING]'' with its complete docstring, and ``[LIB]'' with the library imported in the python script.

\userp{
You are asked to generate two items based on the function [ANSWER]:

1.  A **probe question** about the function's basic usage.

2.  A declarative **answer sentence** that resolves the question.
\\[1ex]
You can refer to the following docstring for context on the function's purpose:

[DOCSTRING]
\\[1ex]
---
**INSTRUCTIONS**

**1. For the "question":**
\quad\quad - It MUST start with "Given the library (libraries) [LIB], how can we ...?".

\quad\quad - It MUST be a single sentence.

\quad\quad - It MUST NOT reveal the function name `[ANSWER]` or its specific arguments.

\quad\quad - It should describe a goal that leads to the simplest, most basic call of the function.

\quad\quad - If `[ANSWER]` includes specific keyword arguments (e.g., `func(arg1=val)`), the question must be phrased to necessitate those exact arguments.

\quad\quad - If `[ANSWER]` is a simple call with no keyword arguments (e.g., `func()`), the question should ask for the standard way to achieve the action.

**2. For the "answer":**

\quad\quad - It MUST be a single, complete, context-independent sentence.

\quad\quad - It MUST combine the premise of the question you just generated with the code snippet `[ANSWER]` to form a factual statement.

\quad\quad - The sentence should state that the action in the question can be accomplished using the provided code.

---

**OUTPUT FORMAT**

You MUST output your response as a valid JSON object and nothing else. Do not add any explanatory text before or after the JSON.

Use the following structure:

\{

\quad\quad "question": "The question you generated.",

\quad\quad "answer": "The answer sentence you generated."

\}
}

\subsection{Scenario \MATH}
\label{sec:appendix:dataset:math}

For this scenario, we adapt problems from the PRM800K benchmark~\citep{lightman2023lets}, which contains math problems with annotated, step-by-step ground-truth solutions. The original problem is the complex reasoning question ($q$), and its final numerical or symbolic result is the final answer ($a$). We leverage the annotated reasoning steps to define the required atomic facts. Specifically, we prompt \texttt{gpt-5-mini} in a Socratic manner, where each crucial step---with a conceptual fact, a formula, or an intermediate calculation—is extracted as an atomic fact ($k_i$) and a context-independent probing question ($q_i$). We use the atomic fact $k_i$ as the probing answer ($a_i$) to verify a model's understanding of that specific reasoning step. Note that the math questions do not contain information that identifies individual people and do not include offensive content.
% The data for the \MATH~scenario is adapted from the ReasonEval benchmark~\citep{xia2025evaluating}, which provides math problems with human-annotated, step-by-step solutions. Our pipeline leverages these detailed solutions to generate the necessary data components by prompting \texttt{gpt-4.1-mini}. 
We set temperature to 0.7, top P to 0.7, and max tokens to 4096.

\subsubsection*{Component Generation Pipeline}
\begin{itemize}[leftmargin=*]
    \item Complex Reasoning Question ($q$) and Final Answer ($a$): These are taken directly from the ReasonEval dataset. The original math problem serves as the question $q$, and its final numerical or symbolic result is the answer $a$.

    \item Atomic Fact ($k_i$) and Probing Question ($q_i$) Generation: The core of our process is to deconstruct the ground-truth reasoning steps into a series of atomic, self-contained facts. We achieve this by prompting \texttt{gpt-4.1-mini} in a Socratic manner.
    \begin{enumerate}
        \item The LLM is provided with the math problem and its complete, step-by-step solution. It is instructed to act as an expert tutor and break down the solution into a chronological sequence of question-knowledge pairs.
        \item The prompt guides the LLM to distinguish between problem-specific steps and general mathematical principles. For a step involving specific numbers from the problem, the generated question must be concrete (e.g., ``Given the equations..., what operation eliminates `e'?''). For a step involving a general theorem, the question must be abstract (e.g., ``How can we convert a value from dollars to cents?'').
        \item The generated "knowledge" sentence, which is a complete, declarative statement, becomes the atomic fact ($k_i$). The corresponding generated "question" becomes the probing question ($q_i$).
    \end{enumerate}
    
    \item Probing Answer ($a_i$): For the \MATH~scenario, the probing answer is identical to the atomic fact ($a_i = k_i$). This tests if the model can recall the specific conceptual or procedural knowledge required for a given step when prompted with a context-independent question.
\end{itemize}

\subsubsection*{Detailed Prompts}
\systemp{
You are an expert AI assistant specializing in educational content creation and the Socratic method. Your primary function is to deconstruct a given mathematical problem's solution into a series of atomic, abstract, and chronological probing questions.
\\[1ex]
**Your Goal:**
Given a math question and its complete, step-by-step reasoning, you will generate a sequence of question-knowledge pairs that test a user's understanding of the solution process.
\\[1ex]
**Key Principles to Follow:**
\\[1ex]
1.  **Formulate Probing Questions:** You must adjust the level of abstraction based on the reasoning step for your questions, and it should be complete, self-contained, declarative sentence.

\quad\quad * **Be CONCRETE for problem-specific steps:** If a step involves translating the problem's specific text or operating on its specific numbers/equations, the question MUST refer to that concrete context.

\quad\quad * **Be ABSTRACT for general knowledge steps:** If a step relies on a general mathematical definition, theorem, or conversion formula that exists outside the specific problem, the question should ask about that general principle.

\quad\quad * **Be COMPLETE, SELF-CONTAINED, and DECLARATIVE:** The question must be independent to the provided math question, meaning it can be answered without the math question.

\quad\quad * **DO NOT** ask something like "According to the problem...", which is NOT SELF-CONTAINED.

2.  **Formulate Complete Knowledge:** The `knowledge` field MUST contain a complete, self-contained, declarative sentence.

\quad\quad * For **concrete questions**, the knowledge should state the specific result of the operation.

\quad\quad * For **abstract questions**, the knowledge should state the general rule or definition.

\quad\quad * The sentence must be context-independent, meaning it can be understood on its own as a piece of knowledge.

\quad\quad * **DO NOT** use short phrases as knowledge (e.g., "By finding the prime factorization").

\quad\quad * **DO NOT** use simple "Yes/No" or short phrases as knowledge.

3.  **Chronological Order:** Your questions must follow the logical sequence of the provided reasoning steps.

4.  **Strict Formatting:** For each generated item, you MUST provide:

\quad\quad * `Question:` [The probing question. It should be complete, self-contained, declarative, and independent to the context math question.]

\quad\quad * `Knowledge:` [The complete, declarative knowledge sentence. It should be complete, self-contained, declarative, and independent to the context math question.]

5.  **Only Necessary Question-Knowledge Pairs:** Provide only question-knowledge pairs that is necessary for solving the math question. [IMPORTANT] It should be as fewer as possible.
\\[1ex]
**Examples to Learn From:**
\\[1ex]
**Example 1:**
\\[1ex]
Math Question: Three pencils and a jumbo eraser cost \$1.24. Five pencils and a jumbo eraser cost \$1.82. No prices include tax. In cents, what is the cost of a pencil?

Reasoning Steps:

1.  Let's call the price of a pencil p and the price of a jumbo eraser e. Then we can write two equations.

2.  We have $3p+e=1.24$ and $5p+e=1.82$.

3.  To solve this system, let's subtract the first equation from the second equation. This will eliminate e.

4.  $5p+e-3p-e=1.82-1.24$.

5.  This simplifies to $2p=0.58$. So $p=0.29$.

6.  That means a pencil costs 29 cents.

Generated Output:

Question: Three pencils and a jumbo eraser cost \$1.24. Five pencils and a jumbo eraser cost \$1.82. No prices include tax. If 'p' is the price of a pencil and 'e' is the price of an eraser, what two equations do we have?

Knowledge: Three pencils and a jumbo eraser cost \$1.24. Five pencils and a jumbo eraser cost \$1.82. No prices include tax. If 'p' is the price of a pencil and 'e' is the price of an eraser, the two equations we have are $3p+e=1.24$ and $5p+e=1.82$.

Question: Given the equations $3p+e=1.24$ and $5p+e=1.82$, what specific operation will eliminate the variable 'e'?

Knowledge: Given the equations $3p+e=1.24$ and $5p+e=1.82$, subtracting the first equation from the second will eliminate the variable 'e'.

Question: After subtracting $3p+e=1.24$ from $5p+e=1.82$, what is the resulting value for p?

Knowledge: After subtracting $3p+e=1.24$ from $5p+e=1.82$, we will have $2p = 0.58$, which solves to $p = 0.29$.

Question: How can we convert a monetary value from dollars to cents?

Knowledge: To convert a value from dollars to cents, you multiply the dollar amount by 100.
\\[1ex]
**Example 2:**
\\[1ex]
Math Question: Compute $58_9 - 18_9.$ Express your answer in base $9.$
Reasoning Steps:

1.  Subtraction works the same in base $9$ as in base $10$.

2.  So we just find the difference of the numbers in the right column, which is $8-8=0$.

3.  Now we need to find the difference of the numbers in the left column. This is $5-1=4$.

4.  The answer is $40_9$.

Generated Output:

Question: Does the standard algorithm for column-based subtraction apply to number systems other than base 10?

Knowledge: The standard algorithm for column-based subtraction is a general method that applies to numbers in any integer base, not just base 10.

Question: Applying column-based subtraction to $58_9 - 18_9$, what are the results for the right and left columns respectively?

Knowledge: Applying column-based subtraction to $58_9 - 18_9$, the result for the right column is $8_9-8_9=0_9$, and the result for the left column is $5_9-1_9=4_9$.

Question: If the result is $0_9$ for the right column and $4_9$ for the left column after applying column-based subtraction in base $9$, what is the final answer in base 9?

Knowledge: If the result is $0_9$ for the right column and $4_9$ for the left column after applying column-based subtraction in base $9$, the final answer in base 9 will be $40_9$.
\\[1ex]
Now, await the user's input.
}

\userp{
Math Question: \{the complex math reasoning question\}

Steps:

\{annotated reasoning steps\}

Generated Output:
}

% Experiment Setup
\section{Experimental Setup}
\label{sec:appendix:setup}

Our experiments are deployed on the RunAI platform. We use AMD EPYC 7543 32-Core Processor as the CPU model and multiple NVIDIA A100-SXM4-80GB as the GPU.

\subsection{Knowledge Probing}
\label{sec:appendix:setup:probing}

\subsubsection*{Hyperparameters}

We set the temperature to 0.7 and top P to 0.7. For each probing question, we sample $M=10$ responses. The maximum output tokens is set to 4096.

\subsubsection*{Detailed Prompts}

The prompts are tailored to each complex reasoning scenario as follows:

\SmallHeading{Scenario \GROW}

\systemp{
Answer the question with the name of an entity. Provide only the name of the entity as your answer. Please make an educated guess and always return an entity.
\\[1ex]
[Here is one demonstration]
\\[1ex]
User:

Who is the developer of Telegram?

Assistant:

Telegram FZ-LLC
}

\userp{
User:

\{probing question\}

Assistant:
}

\SmallHeading{Scenario \CODE}

\systemp{
Answer the question with a Python code snippet, which requires ONLY ONE direct function or class constructor call from ONLY ONE library. Provide ONLY ONE function or constructor call itself with correct positional arguments.

- Do NOT include import statements.

- Do NOT include example data, variable assignments, or any other code.

- For each keyword argument of the function, if the question implies specific keyword arguments, include them in the function call. If the question does not require the keyword argument explicitly or only require it with its default value, the function can be called without this keyword argument.

- Please make an educated guess and always return a function call.
\\[1ex]
[Here is one demonstration]
\\[1ex]
User:

Given the library pandas, how can we create a DataFrame by explicitly passing the input data (such as an ndarray, Iterable, dict, or DataFrame) using the `data` parameter?

Assistant:

```python

pandas.DataFrame(data)

```
}

\userp{
User:

\{probing question\}

Assistant:
}

\SmallHeading{Scenario \MATH}

\systemp{
Answer the math question with a concise sentence. Provide only the direct answer to the math question and no more additional reasoning.
\\[1ex]
[Here is one demonstration]
\\[1ex]
User:

Given the equations $3p+e=1.24$ and $5p+e=1.82$, what specific operation will eliminate the variable 'e'?

Assistant:

Subtracting the first equation from the second will eliminate the variable 'e'.
}

\userp{
User:

\{probing question\}

Assistant:
}

\subsection{Knowledge Injection}
\label{sec:appendix:setup:injection}

\subsubsection*{Hyperparameters}

Across all methods, the final reasoning response is generated with a temperature of 0.7, top-p of 0.7, and a maximum of 8192 new tokens. Method-specific hyperparameters are detailed in \Table~\ref{tab:appendix:hyperparams}.

\begin{table*}[t!]
    \centering
    \resizebox{0.9\textwidth}{!}{
    \begin{tabular}{lll}
        \toprule
        \textbf{Method} & \textbf{Hyperparameter} & \textbf{Value} \\
        \midrule
        \multirow{4}{*}{\textbf{FT-CK}} & LoRA Rank ($r$) & 16 \\
        & LoRA Alpha ($\alpha$) & 16 \\
        & Target Modules & \texttt{q\_proj}, \texttt{k\_proj}, \texttt{v\_proj}, \texttt{o\_proj} \\
        & Learning Rate & 2e-4 \\
        & Num. Train Epochs & 4 \\
        % \midrule
        % \multirow{5}{*}{\textbf{}} & Layers to Edit (Llama-3.2-1B/3B) & [3] / [9] \\
        % & Layers to Edit (Qwen-2.5-1.5B/3B) & [5] / [10] \\
        % & Gradient Steps ($v_{\text{num\_grad\_steps}}$) & 20 \\
        % & Learning Rate ($v_{lr}$) & 0.5 \\
        % & KL Factor ($\text{kl}_{\text{factor}}$) & 0.625 \\
        \midrule
        \multirow{3}{*}{\textbf{MeLLo}} & Retriever Model & \texttt{facebook/contriever-msmarco} \\
        & Decomposer Temperature & 0.01 \\
        & Sub-Answer Generation Temperature & 0.01 \\
        \bottomrule
    \end{tabular}
    }
    \caption{Method-specific (FT-CK and MeLLo) hyperparameters used in our knowledge-injection experiments. The Append and Append-T methods do not have specific hyperparameters.}
    \label{tab:appendix:hyperparams}
\end{table*}

\subsubsection*{Detailed Prompts}

We use one-shot Chain-of-Thought (CoT) prompting for all methods. The core difference lies in how the context $C$ is presented to the model. The Base Model (closed-book) and Append (open-book) methods share a similar prompt structure, differing only in whether the conflicting knowledge is injected into the user prompt. The Append-T method uses the same inference prompt as Append, and FT-CK uses the same prompt as Base Model after fine-tuning.

Below are the prompt templates for the \GROW~scenario. The templates for \CODE~and \MATH~follow a similar structure with domain-specific examples.

\SmallHeading{Closed-book Prompt Template (for Base Model and FT-CK)}

\systemp{
You are given a question. To answer the question, you should think step by step. 
Use line breaks between steps, but do not use line breaks within each step. 
The final answer to the question should start with "The answer is ...", and 
should be placed at the final step. Please make an educated guess and always 
return an entity.
\\[1ex]
[Here is one demonstration]
\\[1ex]
User:

What is the capital of the country where Plainfield Town Hall was created?
\\[1ex]
Assistant:

1. Given my knowledge, Plainfield Town Hall was created in the United States of America.

2. Given my knowledge, the capital of United States is Washington, D.C.

3. Therefore, the capital of the country where Plainfield Town Hall was created 
is Washington, D.C.

4. The answer is Washington, D.C.
}

\userp{
User:

\{complex reasoning question\}

Assistant:
}

\SmallHeading{Open-book Prompt Template (for Append and Append-T):}
\systemp{
... (the system message is the same as closed-book) ...
Users may provide a set of facts or not. If they provide facts that conflict 
with your knowledge, you should update your knowledge and use the facts to 
answer the question.
\\[1ex]
[Here is one demonstration]
\\[1ex]
User:

Who is the person who is the current head of government of British married to?

Please update your knowledge with the following facts:

The name of the current head of the British government is Keir Starmer.
\\[1ex]
Assistant:

1. The user provided that Keir Starmer is the current head of government of the British government.

2. I will update my knowledge with the provided fact...

3. Given my knowledge, Keir Starmer is married to Victoria Starmer.

4. Therefore, the person who is the current head of government of British married to is Victoria Starmer.

5. The answer is Victoria Starmer.
}

\userp{
User:

\{complex reasoning question\}

Please update your knowledge with the following facts:

\{context C (atomic facts to inject)\}

Assistant:
}

\SmallHeading{MeLLo}
The MeLLo method uses a multi-step process involving question decomposition and retrieval. It first generates a series of subquestions, retrieves relevant facts for each, and generates sub-answers:

\systemp{
You are a machine that expertly breaks down complex reasoning questions. Your task is to produce only the VERY NEXT single-line subquestion needed to solve the main question. You MUST use the information based on the answers to the subquestions in the history provided (e.g., substitute entities, use previous answers, etc.). Do not add any extra text or explanation.
\\[1ex]
\#\#\# EXAMPLES \#\#\#
\\[1ex]
[Input]

Main Question: 

What is the capital city of the country of citizenship of Ivanka Trump's spouse?

History:

None

[Next Subquestion]

Who is Ivanka Trump's spouse?
\\[1ex]
[Input]

Main Question: 

What is the capital city of the country of citizenship of Ivanka Trump's spouse?

History:

Subquestion: 

Who is Ivanka Trump's spouse?

Generated Answer: 

Ivanka Trump's spouse is Jared Kushner.

[Next Subquestion]

What is the country of citizenship of Jared Kushner?
\\[1ex]
[Input]

Main Question: 

What is the capital city of the country of citizenship of Ivanka Trump's spouse?

History:

Subquestion: 

Who is Ivanka Trump's spouse?

Generated Answer: 

Ivanka Trump's spouse is Jared Kushner.

Subquestion: 

What is the country of citizenship of Jared Kushner?

Generated Answer: 

Jared Kushner is a citizen of the United States.

[Next Subquestion]

What is the capital city of the United States?
}

\userp{
[Input]

Main Question: 

\{complex reasoning question\}

History:

\{history subquestions and answers\}

[Next Subquestion]
}

The final prompt is then constructed from these generated subquestion-answer pairs as follows:

\systemp{
... (system message is the same as closed-book) ... Users have decomposed the question to multiple subquestions, and answer them one by one. Based on the provided subquestion-answer pairs, answer the question.
\\[1ex]
[Here is one demonstration]
\\[1ex]
User:

Who is the person who is the current head of government of British married to?
Please perform reasoning with following subquestion-answer pairs:

Subquestion: Who is the current head of government of British?

Generated Answer: The name of the current head of the British government is Keir Starmer.

Subquestion: Who is the current Prime Minister of the United Kingdom married to?

Generated Answer: Keir Starmer is married to Victoria Starmer.

Given these subquestion-answer pairs, please answer user's question by reasoning 
step by step.
\\[1ex]
Assistant:

The subquestion-answer pairs have provided all of the required knowledge:

1. Keir Starmer is the current head of government of the British government.

2. We also know that Keir Starmer is married to Victoria Starmer.

3. Therefore, the person who is the current head of government of British married to is Victoria Starmer.

4. The answer is Victoria Starmer.
}

\userp{
User:

\{complex reasoning question\}

Please perform reasoning with following subquestion-answer pairs:

\{subquestion-answer pairs\}

Given these subquestion-answer pairs, please answer user's question by reasoning 
step by step.

Assistant:
}

% Additional Results
\section{Discard Samples Without Knowledge Conflicts}
\label{sec:appendix:results}

\begin{table*}[t!]
    \centering
    \resizebox{0.93\textwidth}{!}{
    \begin{tabular}
{p{2cm}p{2.5cm}p{2.5cm}p{2.5cm}p{2.5cm}p{2.5cm}p{2.5cm}p{2.5cm}p{4cm}}
\toprule
& Llama-3.2 \textsubscript{(1B)} & Llama-3.2 \textsubscript{(3B)} & Llama-3.2 \textsubscript{(11B)} & Qwen-3 \textsubscript{(1.7B)} & Qwen-3 \textsubscript{(4B)} & Qwen-3 \textsubscript{(8B)} & GPT-4.1-mini \& o4-mini \\ \midrule
WIKI & 479 & 458 & 431 & 498 & 475 & 458 & 363\\
CODE & 317 & 246 & 225 & 262 & 452 & 327 & 119\\
MATH & 361 & 363 & 359 & 390 & 344 & 284 & 221 \\
\bottomrule
\end{tabular}
    }
    \caption{Conflicting example counts are reported for the \GROW, \CODE, and \MATH domains. As samples without knowledge gaps are removed, the final numbers vary according to each model's specific knowledge deficiencies.}
    \label{tab:appendix:discard}
\end{table*}

\begin{table*}[t!]
    \centering
    \resizebox{0.93\textwidth}{!}{
    \begin{tabular}{p{2.5cm}p{2cm}p{1.5cm}p{1.5cm}p{1.5cm}p{1.5cm}p{1.5cm}p{1.5cm}p{1.5cm}p{1.5cm}p{1.5cm}}
\toprule
\multirow{2}{*}{Backbone Model} & \multirow{2}{*}{Method} & \multicolumn{3}{c}{Scenario WIKI} & \multicolumn{3}{c}{Scenario CODE} & \multicolumn{3}{c}{Scenario MATH} \\ 
\cmidrule(r){3-5} \cmidrule(l){6-8} \cmidrule(l){9-11}
 &  & \makecell[l]{\% HP} & \makecell[l]{\% AP} & \makecell[l]{\% FKE} & \makecell[l]{\% HP} & \makecell[l]{\% AP} & \makecell[l]{\% FKE} & \makecell[l]{\% HP} & \makecell[l]{\% AP} & \makecell[l]{\% FKE} \\ \midrule
\multirow{4}{*}{Llama-3.2 \textsubscript{(1B)}} & Base Model & 0.7 \textsubscript{$\pm$ 0.7} & 5.9 \textsubscript{$\pm$ 2.0} & 0.7 \textsubscript{$\pm$ 0.7} & 3.9 \textsubscript{$\pm$ 2.1} & 6.8 \textsubscript{$\pm$ 2.7} & 34.5 \textsubscript{$\pm$ 5.2} & 13.4 \textsubscript{$\pm$ 3.5} & 19.7 \textsubscript{$\pm$ 3.9} & 19.4 \textsubscript{$\pm$ 3.9} \\
\cmidrule(l){2-11}
& Append & 0.9 \textsubscript{$\pm$ 0.7} & 7.5 \textsubscript{$\pm$ 2.3} & 0.9 \textsubscript{$\pm$ 0.7} & 3.9 \textsubscript{$\pm$ 2.1} & 5.7 \textsubscript{$\pm$ 2.5} & 31.7 \textsubscript{$\pm$ 5.2} & 16.1 \textsubscript{$\pm$ 3.6} & 22.0 \textsubscript{$\pm$ 4.0} & 21.2 \textsubscript{$\pm$ 4.0} \\
& FT-CK & 0.5 \textsubscript{$\pm$ 0.5} & 6.6 \textsubscript{$\pm$ 2.2} & 0.5 \textsubscript{$\pm$ 0.5} & 3.9 \textsubscript{$\pm$ 2.1} & 6.8 \textsubscript{$\pm$ 2.7} & 30.3 \textsubscript{$\pm$ 5.4} & 14.3 \textsubscript{$\pm$ 3.5} & 20.9 \textsubscript{$\pm$ 4.3} & 19.8 \textsubscript{$\pm$ 4.0} \\
& MeLLo & 1.3 \textsubscript{$\pm$ 1.0} & 10.6 \textsubscript{$\pm$ 2.7} & 1.4 \textsubscript{$\pm$ 0.9} & 3.9 \textsubscript{$\pm$ 2.1} & 5.8 \textsubscript{$\pm$ 2.7} & 30.1 \textsubscript{$\pm$ 5.2} & 12.2 \textsubscript{$\pm$ 3.3} & 17.6 \textsubscript{$\pm$ 3.7} & 18.0 \textsubscript{$\pm$ 3.9} \\
\midrule
\multirow{4}{*}{Llama-3.2 \textsubscript{(3B)}} & Base Model & 1.9 \textsubscript{$\pm$ 1.2} & 20.5 \textsubscript{$\pm$ 3.5} & 2.1 \textsubscript{$\pm$ 1.2} & 13.6 \textsubscript{$\pm$ 4.3} & 20.7 \textsubscript{$\pm$ 5.3} & 48.0 \textsubscript{$\pm$ 6.1} & 32.0 \textsubscript{$\pm$ 5.0} & 46.7 \textsubscript{$\pm$ 4.8} & 35.8 \textsubscript{$\pm$ 5.2} \\
\cmidrule(l){2-11}
& Append & 3.3 \textsubscript{$\pm$ 1.5} & 19.4 \textsubscript{$\pm$ 3.5} & 3.6 \textsubscript{$\pm$ 1.6} & 14.6 \textsubscript{$\pm$ 4.5} & 20.9 \textsubscript{$\pm$ 5.5} & 49.8 \textsubscript{$\pm$ 6.3} & 29.8 \textsubscript{$\pm$ 4.7} & 43.5 \textsubscript{$\pm$ 5.0} & 33.1 \textsubscript{$\pm$ 5.2} \\
& FT-CK & 2.5 \textsubscript{$\pm$ 1.4} & 18.6 \textsubscript{$\pm$ 3.5} & 3.2 \textsubscript{$\pm$ 1.6} & 11.4 \textsubscript{$\pm$ 3.7} & 19.7 \textsubscript{$\pm$ 5.1} & 49.0 \textsubscript{$\pm$ 5.9} & 30.3 \textsubscript{$\pm$ 4.7} & 44.2 \textsubscript{$\pm$ 5.1} & 33.9 \textsubscript{$\pm$ 5.0} \\
& MeLLo & 2.8 \textsubscript{$\pm$ 1.5} & 15.1 \textsubscript{$\pm$ 3.5} & 3.4 \textsubscript{$\pm$ 1.6} & 13.0 \textsubscript{$\pm$ 4.1} & 17.1 \textsubscript{$\pm$ 4.9} & 40.9 \textsubscript{$\pm$ 5.9} & 28.5 \textsubscript{$\pm$ 4.8} & 39.4 \textsubscript{$\pm$ 4.7} & 30.9 \textsubscript{$\pm$ 4.7} \\
\midrule
\multirow{4}{*}{Llama-3.2 \textsubscript{(11B)}} & Base Model & 4.5 \textsubscript{$\pm$ 2.0} & 25.2 \textsubscript{$\pm$ 4.1} & 4.5 \textsubscript{$\pm$ 2.0} & 19.6 \textsubscript{$\pm$ 5.3} & 28.2 \textsubscript{$\pm$ 6.0} & 45.3 \textsubscript{$\pm$ 6.7} & 27.2 \textsubscript{$\pm$ 4.6} & 33.4 \textsubscript{$\pm$ 4.7} & 40.8 \textsubscript{$\pm$ 5.2} \\
\cmidrule(l){2-11}
& Append & 4.8 \textsubscript{$\pm$ 2.0} & 26.6 \textsubscript{$\pm$ 4.1} & 4.9 \textsubscript{$\pm$ 2.1} & 15.1 \textsubscript{$\pm$ 4.5} & 23.8 \textsubscript{$\pm$ 5.6} & 47.1 \textsubscript{$\pm$ 6.7} & 25.9 \textsubscript{$\pm$ 4.5} & 34.1 \textsubscript{$\pm$ 4.9} & 38.9 \textsubscript{$\pm$ 4.9} \\
& FT-CK & 4.3 \textsubscript{$\pm$ 1.7} & 28.2 \textsubscript{$\pm$ 4.1} & 4.5 \textsubscript{$\pm$ 2.0} & 16.4 \textsubscript{$\pm$ 4.9} & 24.9 \textsubscript{$\pm$ 5.8} & 44.9 \textsubscript{$\pm$ 6.2} & 25.5 \textsubscript{$\pm$ 4.6} & 31.3 \textsubscript{$\pm$ 4.6} & 41.5 \textsubscript{$\pm$ 5.3} \\
& MeLLo & 3.8 \textsubscript{$\pm$ 1.7} & 14.4 \textsubscript{$\pm$ 3.2} & 5.0 \textsubscript{$\pm$ 2.0} & 7.3 \textsubscript{$\pm$ 3.3} & 13.8 \textsubscript{$\pm$ 4.4} & 22.9 \textsubscript{$\pm$ 5.6} & 15.6 \textsubscript{$\pm$ 3.6} & 19.2 \textsubscript{$\pm$ 3.9} & 29.2 \textsubscript{$\pm$ 4.7} \\
\midrule
\multirow{5}{*}{Qwen-3 \textsubscript{(1.7B)}} & Base Model & 4.1 \textsubscript{$\pm$ 1.7} & 16.7 \textsubscript{$\pm$ 3.4} & 4.1 \textsubscript{$\pm$ 1.7} & 15.5 \textsubscript{$\pm$ 4.4} & 23.5 \textsubscript{$\pm$ 5.2} & 50.6 \textsubscript{$\pm$ 5.9} & 43.5 \textsubscript{$\pm$ 5.0} & 56.2 \textsubscript{$\pm$ 4.9} & 57.4 \textsubscript{$\pm$ 4.9} \\
\cmidrule(l){2-11}
& Append & \textbf{83.7} \textsubscript{$\pm$ 3.2} & 90.7 \textsubscript{$\pm$ 2.5} & \textbf{87.2} \textsubscript{$\pm$ 2.8} & 17.2 \textsubscript{$\pm$ 4.6} & 25.2 \textsubscript{$\pm$ 5.3} & 58.6 \textsubscript{$\pm$ 5.9} & 43.3 \textsubscript{$\pm$ 4.9} & 53.5 \textsubscript{$\pm$ 5.0} & 60.3 \textsubscript{$\pm$ 4.6} \\
& Append-T & 6.1 \textsubscript{$\pm$ 2.1} & 50.9 \textsubscript{$\pm$ 4.3} & 6.1 \textsubscript{$\pm$ 2.1} & 13.6 \textsubscript{$\pm$ 4.0} & 21.9 \textsubscript{$\pm$ 5.2} & 32.4 \textsubscript{$\pm$ 5.7} & 6.2 \textsubscript{$\pm$ 2.3} & 72.3 \textsubscript{$\pm$ 4.6} & 19.6 \textsubscript{$\pm$ 3.7} \\
& FT-CK & 4.3 \textsubscript{$\pm$ 1.7} & 16.6 \textsubscript{$\pm$ 3.3} & 4.3 \textsubscript{$\pm$ 1.7} & 17.4 \textsubscript{$\pm$ 4.4} & 22.7 \textsubscript{$\pm$ 5.2} & 49.0 \textsubscript{$\pm$ 5.9} & 40.0 \textsubscript{$\pm$ 4.9} & 51.8 \textsubscript{$\pm$ 4.9} & 54.1 \textsubscript{$\pm$ 4.6} \\
& MeLLo & 5.1 \textsubscript{$\pm$ 1.9} & 16.2 \textsubscript{$\pm$ 3.3} & 5.3 \textsubscript{$\pm$ 1.9} & 10.1 \textsubscript{$\pm$ 3.6} & 13.9 \textsubscript{$\pm$ 4.0} & 46.6 \textsubscript{$\pm$ 5.7} & 24.9 \textsubscript{$\pm$ 4.4} & 36.8 \textsubscript{$\pm$ 4.5} & 29.1 \textsubscript{$\pm$ 4.5} \\
\midrule
\multirow{5}{*}{Qwen-3 \textsubscript{(4B)}} & Base Model & 3.9 \textsubscript{$\pm$ 1.8} & 22.9 \textsubscript{$\pm$ 3.8} & 3.8 \textsubscript{$\pm$ 1.7} & 22.2 \textsubscript{$\pm$ 3.7} & 34.2 \textsubscript{$\pm$ 4.3} & 58.4 \textsubscript{$\pm$ 4.6} & 53.5 \textsubscript{$\pm$ 5.2} & 67.3 \textsubscript{$\pm$ 5.1} & 65.3 \textsubscript{$\pm$ 5.1} \\
\cmidrule(l){2-11}
& Append & 80.0 \textsubscript{$\pm$ 3.6} & 94.2 \textsubscript{$\pm$ 2.0} & 82.2 \textsubscript{$\pm$ 3.5} & 26.4 \textsubscript{$\pm$ 3.9} & 33.8 \textsubscript{$\pm$ 4.4} & 69.1 \textsubscript{$\pm$ 4.3} & 51.3 \textsubscript{$\pm$ 5.4} & 66.6 \textsubscript{$\pm$ 4.9} & 66.0 \textsubscript{$\pm$ 4.9} \\
& Append-T & 44.8 \textsubscript{$\pm$ 4.4} & 62.1 \textsubscript{$\pm$ 4.6} & 45.5 \textsubscript{$\pm$ 4.8} & 21.5 \textsubscript{$\pm$ 3.8} & 28.9 \textsubscript{$\pm$ 4.3} & 40.2 \textsubscript{$\pm$ 4.5} & 28.5 \textsubscript{$\pm$ 4.7} & 83.7 \textsubscript{$\pm$ 3.8} & 37.6 \textsubscript{$\pm$ 5.1} \\
& FT-CK & 5.7 \textsubscript{$\pm$ 2.1} & 27.3 \textsubscript{$\pm$ 4.1} & 5.5 \textsubscript{$\pm$ 2.1} & 23.8 \textsubscript{$\pm$ 3.9} & 35.8 \textsubscript{$\pm$ 4.6} & 59.2 \textsubscript{$\pm$ 4.8} & 53.5 \textsubscript{$\pm$ 5.2} & 68.6 \textsubscript{$\pm$ 4.9} & 66.1 \textsubscript{$\pm$ 5.4} \\
& MeLLo & 4.2 \textsubscript{$\pm$ 1.7} & 26.7 \textsubscript{$\pm$ 4.0} & 4.2 \textsubscript{$\pm$ 1.7} & 20.7 \textsubscript{$\pm$ 3.7} & 31.6 \textsubscript{$\pm$ 4.4} & 54.0 \textsubscript{$\pm$ 4.6} & 50.0 \textsubscript{$\pm$ 5.2} & 69.6 \textsubscript{$\pm$ 4.8} & 50.6 \textsubscript{$\pm$ 5.2} \\
\midrule
\multirow{5}{*}{Qwen-3 \textsubscript{(8B)}} & Base Model & 3.6 \textsubscript{$\pm$ 1.6} & 21.1 \textsubscript{$\pm$ 3.6} & 3.7 \textsubscript{$\pm$ 1.7} & 25.1 \textsubscript{$\pm$ 4.6} & 38.7 \textsubscript{$\pm$ 5.4} & 58.0 \textsubscript{$\pm$ 5.0} & 59.0 \textsubscript{$\pm$ 5.8} & 78.3 \textsubscript{$\pm$ 4.8} & 60.4 \textsubscript{$\pm$ 5.8} \\
\cmidrule(l){2-11}
& Append & 77.3 \textsubscript{$\pm$ 3.7} & 91.6 \textsubscript{$\pm$ 2.5} & 78.5 \textsubscript{$\pm$ 3.6} & 25.1 \textsubscript{$\pm$ 4.6} & 39.4 \textsubscript{$\pm$ 5.2} & 57.5 \textsubscript{$\pm$ 5.2} & 60.2 \textsubscript{$\pm$ 5.6} & 82.0 \textsubscript{$\pm$ 4.6} & 61.3 \textsubscript{$\pm$ 5.6} \\
& Append-T & 72.7 \textsubscript{$\pm$ 4.1} & 93.9 \textsubscript{$\pm$ 2.2} & 73.5 \textsubscript{$\pm$ 3.8} & 23.1 \textsubscript{$\pm$ 4.7} & 30.6 \textsubscript{$\pm$ 4.9} & 38.4 \textsubscript{$\pm$ 5.4} & 60.4 \textsubscript{$\pm$ 5.8} & 73.8 \textsubscript{$\pm$ 5.1} & 76.1 \textsubscript{$\pm$ 4.9} \\
& FT-CK & 3.4 \textsubscript{$\pm$ 1.6} & 22.2 \textsubscript{$\pm$ 3.6} & 3.4 \textsubscript{$\pm$ 1.6} & 25.7 \textsubscript{$\pm$ 4.6} & 37.3 \textsubscript{$\pm$ 5.2} & 58.6 \textsubscript{$\pm$ 5.0} & 57.2 \textsubscript{$\pm$ 5.8} & 78.0 \textsubscript{$\pm$ 4.8} & 58.3 \textsubscript{$\pm$ 5.8} \\
& MeLLo & 5.0 \textsubscript{$\pm$ 2.0} & 22.9 \textsubscript{$\pm$ 3.9} & 5.3 \textsubscript{$\pm$ 2.1} & 24.9 \textsubscript{$\pm$ 4.7} & 35.0 \textsubscript{$\pm$ 5.0} & 56.1 \textsubscript{$\pm$ 5.7} & 35.0 \textsubscript{$\pm$ 5.8} & 52.5 \textsubscript{$\pm$ 5.6} & 35.9 \textsubscript{$\pm$ 5.3} \\
\midrule
\multirow{2}{*}{GPT-4.1-mini \&} & Base Model & 7.3 \textsubscript{$\pm$ 2.6} & 35.4 \textsubscript{$\pm$ 4.8} & 7.3 \textsubscript{$\pm$ 2.6} & 34.5 \textsubscript{$\pm$ 8.4} & 51.3 \textsubscript{$\pm$ 9.2} & 62.6 \textsubscript{$\pm$ 8.8} & 69.3 \textsubscript{$\pm$ 6.1} & 89.8 \textsubscript{$\pm$ 3.9} & 72.0 \textsubscript{$\pm$ 6.1} \\
\cmidrule(l){2-11}
& Append & 74.8 \textsubscript{$\pm$ 4.5} & \textbf{95.3} \textsubscript{$\pm$ 2.2} & 75.5 \textsubscript{$\pm$ 4.4} & 39.5 \textsubscript{$\pm$ 8.4} & \textbf{51.7} \textsubscript{$\pm$ 8.8} & 67.6 \textsubscript{$\pm$ 8.0} & 86.4 \textsubscript{$\pm$ 4.5} & \textbf{99.3} \textsubscript{$\pm$ 0.7} & \textbf{87.1} \textsubscript{$\pm$ 4.3} \\
o4-mini & Append-T & 82.6 \textsubscript{$\pm$ 3.9} & \textbf{95.3} \textsubscript{$\pm$ 2.2} & 82.6 \textsubscript{$\pm$ 3.9} & \textbf{45.0} \textsubscript{$\pm$ 8.8} & 50.4 \textsubscript{$\pm$ 8.4} & \textbf{73.5} \textsubscript{$\pm$ 8.0} & \textbf{86.6} \textsubscript{$\pm$ 4.8} & 99.3 \textsubscript{$\pm$ 0.7} & 86.8 \textsubscript{$\pm$ 4.5} \\
\bottomrule
\end{tabular}
    }
    \caption{Percentage of Holistic Pass (\% HP), Answer Pass (\% AP), and Full Knowledge Entailment (\% FKE) on the \GROW (multi-hop QA on Wikidata), \CODE (code generation with external APIs), and \MATH (multi-step mathematical reasoning) scenarios. \textbf{We discard all examples without knowledge conflicts} and compare the Base Model (closed-book) against several open-book knowledge injection methods (Append, FT-CK, MeLLo) across Llama-3.2, Qwen-3, and GPT series. We set KAS to 1 in the open-book setting so each question receives only its missing facts. We report 95\% confidence intervals (CIs) in the $\pm$ sign and \textbf{bold} the best scores per column.}
    \label{tab:appendix:extramain}
\end{table*}

We do not discard samples without knowledge conflicts. Keeping them ensures a fixed test set for fair comparison across different models and avoids introducing inductive bias \cite{si-etal-2023-measuring} from pretraining models. In \Table~\ref{tab:appendix:discard}, we report the number of remaining conflicting examples, and in \Table~\ref{tab:appendix:extramain}, we further report the performance of each model if we were to remove non-conflict samples. Despite the removal of examples, the relative performance ranking between models and methods remains almost consistent. For instance, even when we compare the performance of GPT-4.1-mini and o4-mini on CODE, which removes the most number of examples, we show that the exclusion of non-conflict samples does not alter our core conclusions.

% Checklists

\section{Responsible NLP Research} \label{sec:appendix:checklist}

\subsection{Artifacts}
\label{sec:appendix:checklist:artifacts}
To foster reproducibility and open science, we will make our complete codebase and all reconstructed datasets publicly available under the MIT License. The essential artifacts used in this project, including datasets, backbone models, and major software libraries, are detailed in \Table~\ref{tab:tools}. It is important to note that all data is in English. We have adhered to the intended-use licenses for all artifacts, which permit non-commercial research applications.

\begin{table*} 
    \centering
    \resizebox{\textwidth}{!}{
    \begin{tabular} 
    {llll} 
       \toprule % <-- Toprule here
        \textbf{Artifacts/Models/Packages} & \textbf{Citation} & \textbf{Link} & \textbf{License}\\ 
        \rowcolor{\grayColor} \multicolumn{4}{c}{\textit{Data Artifacts}}\\
        Wikidata & \cite{vrandevcic2014wikidata} & \url{https://www.wikidata.org/} & Creative Commons BY-SA 4.0 License \\
        BigCodeBench & \cite{zhuo2024bigcodebench} & \url{https://bigcode-bench.github.io/} & Apache-2.0 License \\
        PRM800K & \cite{lightman2023lets} & \url{https://github.com/openai/prm800k/} & MIT License \\
        \rowcolor{\grayColor} \multicolumn{4}{c}{\textit{Backbone Models}} \\
        LlaMA-3.2 (1B, 3B, 11B) & \cite{grattafiori2024llama} & \url{https://ai.meta.com/blog/llama-3-2-connect-2024-vision-edge-mobile-devices} & Llama 3.2 Community License Agreement \\
        Qwen-3 (1.7B, 4B, 8B) & \cite{yang2025qwen3technicalreport} & \url{https://qwen3.org} & Apache-2.0 License \\
        GPT-4.1-mini & \cite{openai2025gpt41} & \url{https://openai.com/index/gpt-4-1/} & Missing \\
        o4-mini & \cite{openai2025o4} & \url{https://openai.com/index/introducing-o3-and-o4-mini/} & Missing \\
        Gemini-2.5-Pro & \cite{comanici2025gemini} & \url{https://deepmind.google/models/gemini/pro/} & Missing \\
        \rowcolor{\grayColor} \multicolumn{4}{c}{\textit{Packages}}  \\
        PyTorch & \cite{paszke2019pytorch} & \url{https://pytorch.org/} & BSD 3-Clause License\\
        transformers & \cite{wolf2020transformers} & \url{https://huggingface.co/docs/transformers/index} & Apache-2.0 License\\
        numpy & \cite{harris2020array} & \url{https://numpy.org/} & BSD License \\
        pandas & \cite{mckinney2011pandas} & \url{https://pandas.pydata.org/} & BSD 3-Clause License \\
        matplotlib & \cite{hunter2007matplotlib} & \url{https://matplotlib.org/} & Python Software Foundation License\\
        seaborn & \cite{waskom2021seaborn} & \url{https://seaborn.pydata.org/} & BSD 3-Clause License\\
        openai-python & \cite{achiam2023gpt} & \url{https://pypi.org/project/openai/} & Apache-2.0 License \\
        togetherai & \cite{togetherai} & \url{https://www.together.ai/} & Apache-2.0 License \\
        Flash Attention 2 & \cite{dao2023flashattention2} & \url{https://github.com/Dao-AILab/flash-attention} & BSD-3-Clause License \\
        amCharts 5 & \cite{amcharts5} & \url{https://www.amcharts.com/docs/v5/} & Basic License \\
      \bottomrule % <-- Bottomrule here
    \end{tabular}
    }
    \caption{Data artifacts, backbone models, and major packages utilized in our study. All the reconstructed datasets and the provided code of our project are released under the MIT License to support open science and reproducibility.}
    \label{tab:tools}
\end{table*}

\subsection{Usage of Artificial Intelligence Assistants} \label{sec:appendix:checklist:ai}
We use Artificial Intelligence (AI) assistants only to help us with data annotation, basic code completions and grammar checking.

\SmallHeading{AI Annotators}
To validate the quality of our \taskalias benchmark, we employ two AI assistants (\texttt{gpt-5-mini} and \texttt{gemini-2.5-pro}\footnote{Released on June 17, 2025.}) to annotate a sample of 50 items per domain for two distinct tasks. The first task evaluates the factual correctness of probing questions, while the second assesses the necessity of each knowledge step in solving the reasoning questions. Across all domains, \texttt{gpt-5-mini} identified 96.0\% of items as factual and 86.7\% as necessary. Similarly, \texttt{gemini-2.5-pro} annotated 92.7\% as factual and 89.3\% as necessary. The average ratio is 94.4\% for factuality and 88.0\% for necessity, respectively. The high scores demonstrate a high degree of consistency in the validation process.
To ensure the AI annotation is reliable, we further sample 50 annotations from both GPT-5-mini and Gemini-2.5-Pro. One PhD student who majors in Computer Science from university in the US rigorously annotated these samples using the same prompt sent to LLMs. The annotations of LLMs have high average F1 scores with the human annotator (98.0\% for factuality and 93.5\% for necessity).
% The joint agreement where both models concurred on a positive label is 90.00\% for factuality annotations and 79.33\% for necessity annotations, demonstrating a high degree of consistency in the validation process. 
Detailed prompts are shown below:

\begin{boxuser}
    {\rmfamily\textbf{User (WIKI, Factuality):}}\\[1ex]
    You are given a question and an answer. Is the answer to the question factually correct? Please provide a short sentence as explanation and then answer Yes if the answer is factually correct or No if it is not.
\end{boxuser}

\begin{boxuser}
    {\rmfamily\textbf{User (WIKI, Necessity):}}\\[1ex]
    You are given a multi-hop reasoning question. You are also given a list of knowledge steps. Your task is to determine if every piece of knowledge is essential for solving the problem. A step is essential if the final answer cannot be reached without it.
\end{boxuser}

\begin{boxuser}
    {\rmfamily\textbf{User (CODE, Factuality):}}\\[1ex]
    You are given a coding question which requires a function call from Python external libraries as the ground truth answer. Is the answer factually correct? Please provide a short sentence as explanation and then answer Yes if the answer is factually correct or No if it is not.

    Question: \{probing question\}

    Answer: \{probing answer\}
\end{boxuser}

\begin{boxuser}
    {\rmfamily\textbf{User (CODE, Necessity):}}\\[1ex]
    You are given a coding question which requires implementing a Python code with multiple required functions from external libraries. Does the problem description explicitly require the use of every function listed below to be considered a valid solution? Answer Yes if all functions are mandated by the prompt's text, or No otherwise.
\end{boxuser}

\begin{boxuser}
    {\rmfamily\textbf{User (MATH, Factuality):}}\\[1ex]
    You are given a math question. Is the answer to the question factually correct? Please provide a short sentence as explanation and then answer Yes if the answer is factually correct or No if it is not.
\end{boxuser}

\begin{boxuser}
    {\rmfamily\textbf{User (MATH, Necessity):}}\\[1ex]
    You are given a math question and a required solution path. You are also given a list of knowledge steps. Your task is to determine if every piece of knowledge is essential for solving the problem **by following the required path exactly**.

    Are all the listed knowledge steps required to construct the solution **as demanded by the problem's constraints**? An alternative mathematical method is irrelevant if it deviates from the specified path. Provide a short sentence as explanation. Then answer **Yes** if all knowledge is essential to the required method, or **No** if some knowledge is irrelevant or deviates from the method.
\end{boxuser}

\SmallHeading{AI Code Completions}
To streamline the development process, we leveraged \href{https://github.com/features/copilot}{GitHub Copilot} for assistance. The tool was primarily used to generate routine code, such as inline comments, function header documentation, and boilerplate statements like \texttt{if \_\_name\_\_ == "\_\_main\_\_":}. The high-level software architecture and the core logic of all functions were manually designed and implemented by the authors.

\SmallHeading{Grammar Checking}
The initial draft of this paper was composed manually. For refinement, we employed a suite of AI-powered writing assistants, including \href{https://www.deepl.com/translator}{DeepL} for translation, and \href{https://chatgpt.com/chat}{ChatGPT}, \href{https://gemini.google.com/}{Gemini} for grammatical correctness.

\end{document}